%% file: main.tex
\documentclass{article}

\PassOptionsToPackage{numbers,compress}{natbib}

 \usepackage[main, final]{neurips_2026}

\usepackage[utf8]{inputenc} 
\usepackage[T1]{fontenc}    
\usepackage{hyperref}       
\usepackage{url}            
\usepackage{booktabs}       
\usepackage{amsfonts}       
\usepackage{nicefrac}       
\usepackage{microtype}      
\usepackage{xcolor}         

\usepackage{graphicx}       

\usepackage{amsmath}

\usepackage{multirow}
\usepackage{tabularx}
\usepackage{adjustbox}

\usepackage[table]{xcolor}
\definecolor{gold}{HTML}{FFF2CC} 
\definecolor{goldtext}{HTML}{D4AF37} 
\definecolor{silver}{HTML}{959595} 
\definecolor{bronze}{HTML}{A97142} 

\newcommand{\bronze}[1]{\textcolor{bronze}{\textbf{#1}}}

\usepackage{amsthm}

\newtheorem{lemma}{Lemma}

\theoremstyle{definition}

\title{Forest-Guided Semantic Transport for Label-Supervised Manifold Alignment}

\author{%
Adrien Aumon\textsuperscript{\textnormal{1}}\\
\texttt{adrien.aumon@umontreal.ca}
\And
Myriam Lizotte\textsuperscript{\textnormal{1}}\\
\texttt{myriam.lizotte@mila.quebec}
\And
Guy Wolf\textsuperscript{\textnormal{1}}\\
\texttt{guy.wolf@umontreal.ca}
\And
Kevin R.~Moon\textsuperscript{\textnormal{2}}\\
\texttt{kevin.moon@usu.edu}
\And
Jake S.~Rhodes\textsuperscript{\textnormal{3}\ddag}\\
\texttt{rhodes@stat.byu.edu}
\AND
\normalfont
\centerline{\textsuperscript{1}Université de Montréal; Mila -- Quebec AI Institute}\\[3pt]
\centerline{
\textsuperscript{2}Utah State University\quad\textsuperscript{3}Brigham Young University}\\[3pt]
\centerline{\textsuperscript{\ddag}Corresponding author}
}

\begin{document}

\maketitle

\begin{abstract}

Label-supervised manifold alignment bridges the gap between unsupervised and correspondence-based paradigms by leveraging shared label information to align multimodal datasets. However, existing methods either rely on label-independent intra-domain geometry or incorporate supervision primarily through the alignment objective, rather than directly constructing task-aware intra-domain affinities. To address this limitation, we introduce FoSTA (Forest-guided Semantic Transport Alignment), which learns supervised forest geometries independently within each domain and uses their shared class-semantic structure to infer cross-domain correspondences without predefined anchors. FoSTA extends RF-GAP affinities to partially labeled data and aligns the resulting semantic representations through scalable hierarchical transport. Extensive comparisons with established baselines demonstrate strong correspondence recovery and semantic structure preservation on synthetic and real-world benchmarks, including multimodal data integration and single-cell batch correction.

\end{abstract}

\section{Introduction}\label{sec:intro}

Manifold alignment (MA) seeks a shared representation for datasets collected in different modalities or feature spaces, and is a key primitive in multimodal learning and single-cell integration~\cite{ham2003learning, ham2005semisupervised, hieEfficientIntegrationHeterogeneous2019}. Most approaches fall into two broad families: \emph{semi-supervised} methods that rely on a set of known cross-domain correspondences (anchors)~\cite{wangManifoldAlignmentUsing2008, duqueDiffusionTransportAlignment2023}, and \emph{unsupervised} methods that assume no shared information and instead match domains using geometric or distributional structure~\cite{memoli2011gromov, demetci2022scot}. In many practical settings, however, pointwise correspondences are impossible or expensive to obtain, while weaker side information---like shared class labels---is readily available (e.g., curated cell types across batches~\cite{caoManifoldAlignmentHeterogeneous2021}). This motivates \emph{label-supervised} manifold alignment, which uses labels to guide cross-domain matching without using anchors~\cite{wang2011heterogeneous, tuiaKernelManifoldAlignment2016, duqueManifoldAlignmentLabel2023}.

Most label-supervised alignment methods rely on Euclidean geometry to model intra-domain structure. This fails when only a subset of features is relevant to the task, as Euclidean distances weigh all coordinates equally, mixing signal with nuisance variation. In high-dimensional settings, this creates misleading neighborhoods that degrade the alignment. Tree ensemble affinities offer a principled alternative by adaptively emphasizing label-relevant structure. In particular, RF-GAP proximities~\cite{rhodesGeometryAccuracyPreservingRandom2023} have been shown to outperform standard forest-based proximities and class-conditional Euclidean kernels in guided representation learning~\cite{rhodesGeometryAccuracyPreservingRandom2023, rhodes2026gaining, aumonRandomForestAutoencoders2025}. However, existing forest-supervised alignment methods~\cite{rhodesRandomForestSupervisedManifold2024} use forest affinities for within-domain geometry while still requiring known cross-domain correspondences. Whether independently learned supervised forest geometries can instead provide sufficient shared structure to \emph{infer} these correspondences remains unexplored.


To address this gap, we introduce \textbf{FoSTA} (\underline{Fo}rest-guided \underline{S}emantic \underline{T}ransport \underline{A}lignment), a label-supervised framework that uses independently learned forest geometries to infer correspondences between heterogeneous domains without predefined anchors. We extend RF-GAP~\cite{rhodesGeometryAccuracyPreservingRandom2023} to partial label assignment and use the resulting affinities to represent samples through their relationships to shared semantic classes. Cross-domain correspondences are then inferred by scalable semantic transport and propagated through the forest graphs to construct a joint geometry. The formulation further accommodates unequal domain sizes and partially overlapping class sets.

Across real-world multimodal benchmarks, FoSTA achieves strong correspondence recovery and label transfer compared to established baselines~\cite{wang2011heterogeneous, tuiaKernelManifoldAlignment2016, duqueManifoldAlignmentLabel2023, caoManifoldAlignmentHeterogeneous2021}. On a single-cell batch alignment task, it achieves competitive batch correction while preserving biological variation. Controlled synthetic experiments further show its robustness to different domain transformations, levels of label availability, and design choices, confirming that the alignment gains mainly come from the upstream forest geometry.


\section{Background}

\subsection{Guided manifold learning with tree ensembles}\label{subsec:treeprox}


Manifold learning methods traditionally rely on unsupervised, Euclidean-based kernels (e.g., radial basis functions) to capture local geometric structure. When label information is available, however, meaningful representations should reflect not only feature–feature relationships but also feature–label dependencies. In such settings, purely unsupervised geometry can be misleading, as it treats all features equally regardless of their relevance to the label signal.

Pairwise proximities derived from leaf co-occurrence in trained Random Forests (RFs) or other tree ensembles~\cite{breimanRandomForests2001, breiman2003manual, tanTreeSpacePrototypes2020, rhodesGeometryAccuracyPreservingRandom2023} address this limitation by jointly capturing similarity in the feature space and the task-relevant structure, implicitly downweighting features that are irrelevant for prediction. In particular, RF-GAP proximities~\cite{rhodesGeometryAccuracyPreservingRandom2023} provide a robust label-informed proximity measure through a leaf weighting scheme based on in-bag/out-of-bag status, which reflects the predictive behavior of the underlying forest. This property has led to substantial improvements in guided representation learning compared to standard forest-based proximity definitions and Euclidean class-conditional approaches~\cite{rhodes2026gaining, aumonRandomForestAutoencoders2025}. In addition, tree-induced proximities naturally accommodate both continuous and categorical features and their exact calculation scales near-linearly in time and memory~\cite{aumon2026revisitingforestproximitiessparse}, whereas Euclidean kernels often require approximate nearest-neighbor methods to remain tractable~\cite{indyk1998approximate}.

\subsection{Manifold alignment approaches}\label{subsec:manifold_alignment}
Given multiple domains containing observations from different modalities that describe related phenomena, manifold alignment aims to learn a shared representation that preserves both (i) the intrinsic geometric structure within each domain and (ii) information shared across domains. While intra-domain geometry can typically be inferred directly from the respective feature spaces, capturing shared cross-domain information depends critically on the assumptions made about how the domains are related.

\textbf{Semi-supervised MA} assumes access to a subset of one-to-one correspondences across domains, which serve as anchor points to guide alignment. These anchors provide explicit cross-domain constraints that allow the remaining unmatched samples to be embedded into a common space.

Traditional approaches include canonical correlation analysis~\cite{thompson1984canonical}, which learns linear projections that maximize the correlation between paired observations. While efficient, such global linear models are often inadequate for capturing the nonlinear structure of high-dimensional data. This limitation motivated subsequent graph-based methods~\cite{ham2003learning, ham2005semisupervised}, which preserve local manifold geometry within each domain while enforcing alignment through anchor constraints. Alternatively, \citet{wangManifoldAlignmentUsing2008} projects both domains into a shared space using dimensionality reduction (DR), followed by Procrustes alignment on anchor points, and generally outperforms earlier approaches.

More recent semi-supervised MA methods leverage anchor correspondences while better preserving intra-domain geometry. Generative approaches~\cite{amodioMAGANAligningBiological2018} learn mappings between domains in a CycleGAN-style framework~\cite{zhu2017unpaired}, augmented with correspondence losses to enforce alignment consistency and reduce superimposition. \citet{duqueDiffusionTransportAlignment2023} incorporates diffusion geometry by using anchor correspondences to derive diffusion-informed inter-domain distances, which define the cost of an entropic OT problem~\cite{peyre2019computational}; the resulting coupling is then used to construct inter-domain affinities prior to joint embedding. In contrast, \citet{rhodesGraphIntegrationDiffusionBased2024} aligns domains by directly linking their diffusion operators through a union graph. Building on this idea,~\citet{rhodesRandomForestSupervisedManifold2024} moves beyond Euclidean-based alignment by using RF-derived affinities within each domain. However, their method assumes fully labeled source and target domains, which could be restrictive; we relax this assumption in Section~\ref{subsec:intradomain} by allowing partially labeled settings. 

Other multimodal integration methods rely on different forms of cross-domain prior information. For instance, GLUE~\cite{caoMultiomicsSinglecellData2022} uses a guidance graph encoding known relationships between features across modalities. While effective when such prior relationships are available, this assumption limits its applicability to general heterogeneous domains with unrelated feature spaces.

Anchor-based approaches rely on known correspondences, which can be expensive or impossible to obtain. This limitation is particularly pronounced in biological applications, where measurement processes are often destructive and prevent observing multiple modalities from the same cells.

\textbf{Unsupervised MA} is the most challenging setting, as it assumes no explicit shared information across domains. However, domains describing the same phenomenon are expected to exhibit similar underlying structure. This assumption allows correspondences to be inferred indirectly by relating points across domains based solely on their locations within their respective manifolds.

Early unsupervised approaches sought to recover shared structure by constructing joint representations from intra-domain geometry. \citet{wang2009manifold} introduced a cross-domain heat kernel based on permutations of local geometric structure, combined with intra-domain kernels, followed by DR to obtain a shared embedding. Subsequent work framed unsupervised MA as a problem of implicit global distribution matching, e.g., by minimizing discrepancies between kernel mean embeddings~\cite{liuJointlyEmbeddingMultiple2019} or by matching intrinsic geometric structure using geodesic distances rather than kernel matrices~\cite{caoUnsupervisedTopologicalAlignment2020}.

Rather than enforcing global matching, a complementary line of work infers correspondences through local neighborhood consistency. \citet{haghverdiBatchEffectsSinglecell2018} and~\citet{hieEfficientIntegrationHeterogeneous2019} identify mutual nearest neighbors (NN) across domains and use these pairs as anchors to apply local, translation-based corrections. Similarly,~\citet{korsunskyFastSensitiveAccurate2019} and~\citet{welchSinglecellMultiomicIntegration2019} rely on local alignment heuristics, using cluster-wise linear corrections or shared (dataset-invariant) latent factors, respectively.

To move beyond heuristic local matching and explicitly enforcing global geometric consistency, another line of work formulates unsupervised MA as an OT problem. In this setting, correspondences are inferred globally using the Gromov--Wasserstein OT family (GWOT~\cite{memoli2011gromov}), with discrepancies in geodesic distances serving as the transport cost, followed by DR~\cite{demetci2022scot, caoManifoldAlignmentHeterogeneous2021}. Alternatively,~\citet{chenUnsupervisedManifoldAlignment2022} extends Procrustes-based alignment by first projecting each domain into a shared latent space and then refining the alignment via Wasserstein Procrustes~\cite{graveUnsupervisedAlignmentEmbeddings2019}.

Even in the absence of explicit correspondences, one does not need to resort to fully unsupervised manifold alignment when side information is available in both domains.

\textbf{Label-supervised MA} occupies a middle ground between fully unsupervised and semi-supervised MA by leveraging an auxiliary form of shared information. A common source of such side information is a set of discrete class labels~\cite{wang2011heterogeneous, tuiaKernelManifoldAlignment2016, wangLabelSpaceEmbedding2019}, for example, known cell types in single-cell analysis~\cite{caoManifoldAlignmentHeterogeneous2021}.

Early work in this setting directly learned a latent joint representation by incorporating label information into the alignment objective, enforcing strict same-class matching alongside intra-domain geometric regularization (KEMA~\cite{wang2011heterogeneous, tuiaKernelManifoldAlignment2016}). However, this strategy is highly sensitive to label availability, as samples with missing labels are left unmatched. In contrast, Manifold Alignment with Label Information (MALI~\cite{duqueManifoldAlignmentLabel2023}) adopts a two-stage approach in which cross-domain correspondences are first inferred through an OT formulation, and a joint similarity matrix is then constructed from the resulting coupling. This matrix can subsequently be used to obtain a shared latent representation if desired. MALI proceeds similarly to~\citet{duqueDiffusionTransportAlignment2023}, but replaces cross-domain diffusion-based distances through anchors with cosine distances between semantic representations, defined as projections onto a label space obtained by aggregating class-wise affinities within each domain.

In single-cell analysis, recent methods extend unsupervised GWOT alignment by incorporating labels into the transport cost, thus favoring matches between cells of the same type~\cite{caoManifoldAlignmentHeterogeneous2021}. Others leverage label information through probabilistic deep generative models for batch correction~\cite{xuProbabilisticHarmonizationAnnotation2021}. However, these generative models typically assume that all domains share the same feature space, limiting them to homogeneous data integration.

These methods fall under a \textit{weakly} supervised MA setting, as correspondences are inferred rather than provided explicitly, while still exploiting shared semantic information across domains. Existing label-supervised alignment methods differ in how supervision enters the alignment: MALI uses labels primarily to infer cross-domain correspondences, while KEMA incorporates class information directly into its alignment objective. In contrast, FoSTA uses labels to construct the intra-domain geometry itself through supervised forest affinities, which subsequently drive correspondence inference.

\section{Methods}\label{sec:methods}

\begin{figure*}[!ht]
    \centering
    \includegraphics[width = 1\textwidth]{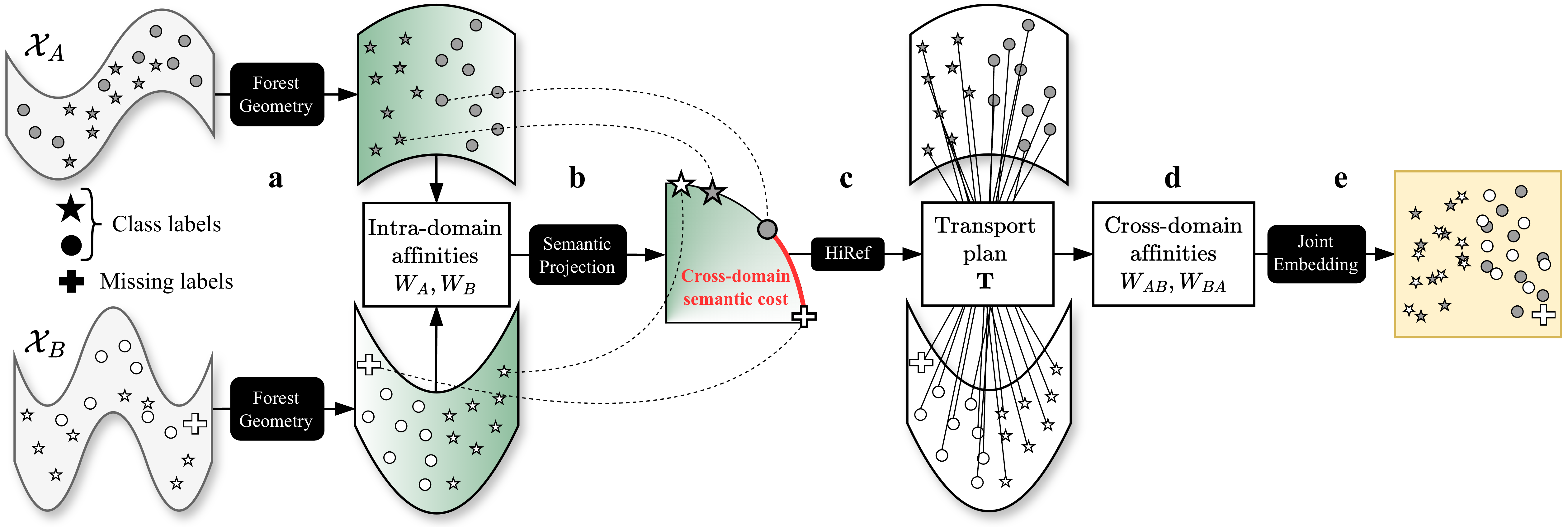}
    \caption{Overview of FoSTA. \textbf{(a)} Given two (partially labeled) domains $\mathcal{X}_A$ and $\mathcal{X}_B$, we compute semi-supervised, forest-derived affinities $W_A$ and $W_B$ to capture task-relevant neighborhood structure (Section~\ref{subsec:intradomain}). \textbf{(b)} We form class-wise semantic profiles and map samples from both domains to a shared, $\ell_2$-normalized semantic space, which defines a cross-domain semantic cost (Section~\ref{subsec:transport}). \textbf{(c)} Using this cost, HiRef estimates an efficient bijective transport plan $\mathbf{T}$ between domains (Section~\ref{subsec:transport}). \textbf{(d)} We propagate correspondences through $\mathbf{T}$ to build cross-domain affinities $W_{AB}$ and $W_{BA}$ consistent with the intra-domain graphs (Section~\ref{subsec:cross-domain}). \textbf{(e)} Finally, we compute a joint low-dimensional embedding from the resulting block affinity matrix (Section~\ref{subsec:joint_embedding}).}
\label{fig:fosta}
\end{figure*}

\subsection{Problem formulation}\label{subsec:formulation}

Given a source domain $\mathcal{X}_{A} = \{\mathbf{x}_1^{A}, \ldots, \mathbf{x}_{n}^{A}\}$ and a target domain $\mathcal{X}_{B} = \{\mathbf{x}_1^{B}, \ldots, \mathbf{x}_{m}^{B}\}$, manifold alignment aims to learn a shared representation that preserves both intra-domain geometric structure and cross-domain relationships. 

We consider a weakly supervised setting in which the two domains share a common label space, but explicit correspondences are unavailable. We further relax the fully labeled assumption by allowing both domains to be only partially labeled. Let $\mathcal{Y}_A = \{\mathbf{y}_1^A, \ldots, \mathbf{y}_r^A\}$ and
$\mathcal{Y}_B = \{\mathbf{y}_1^B, \ldots, \mathbf{y}_s^B\}$, with $r \le n$ and $s \leq m$,
denote the observed labels in the source and target domains, respectively.
We assume that domain labels take values in finite class sets
$\mathcal{C}_A$ and $\mathcal{C}_B$, respectively, and that $\mathcal{C}_A\cap \mathcal{C}_B\neq \emptyset$. The goal is therefore to leverage this limited shared supervision to infer meaningful relationships across domains.


Our method focuses on learning cross-domain similarities that are comparable in scale to intra-domain affinities, which naturally supports both correspondence estimation and the construction of shared embeddings~\cite{duqueDiffusionTransportAlignment2023, duqueManifoldAlignmentLabel2023}. Thus, the core objective is to build a joint affinity matrix of the form
\[
\mathbf{W} =
\begin{bmatrix}
W_A & W_{AB} \\
W_{BA} & W_{B}
\end{bmatrix} \in \mathbb{R}_{\geq 0}^{(n+m) \times (n+m)},
\]
where the diagonal blocks represent intra-domain affinities and the off-diagonal blocks represent cross-domain affinities. The construction of each affinity block is detailed in the sections that follow, and Fig.~\ref{fig:fosta} presents a schematic representation of our FoSTA framework.

\subsection{Label-informed intra-domain affinities}
\label{subsec:intradomain}


Let $\mathcal X = \{\mathbf{x}_1, \ldots, \mathbf{x}_n\}$ be a partially labeled dataset, and denote by $\mathcal I^l$ and $\mathcal I^u$ the sets of labeled and unlabeled indices, respectively, with $\mathcal I^l \cup \mathcal I^u = \{1,\ldots,n\}$. Our goal is to construct class-aware intra-domain affinities that better reflect the label structure than standard distance-based kernels.

We train a bootstrapped forest classifier (typically an RF) on the labeled subset $\{\mathbf{x}_i : i \in \mathcal I^l\}$ and use it to define affinities between all pairs of points. Let $\mathcal T$ be the set of trees, with $T = |\mathcal T|$. For a query point $\mathbf{x}_i$, denote by $S_i \subset \mathcal T$ the set of trees in which $\mathbf{x}_i$ is out-of-bag. For each tree $t \in \mathcal T$, let $\ell_i(t)$ be the leaf reached by $\mathbf{x}_i$, let $M_i(t)$ be the multiset of in-bag training samples that fall in that leaf, and let $c_j(t)$ denote the bootstrap multiplicity of any labeled point $\mathbf{x}_j$. We define the intra-domain affinity matrix $W$ by
\begin{equation}
\label{eq:intradomain_affinity}
W_{ij}
=
\frac{1}{|S_i|}
\sum_{t \in S_i}
\frac{c_j(t)\, \mathbb{I}\!\left(\ell_i(t)=\ell_j(t)\right)}{|M_i(t)|}\in \mathbb{R}_{\ge 0}^{n \times n}.
\end{equation}

When the target index $j \in \mathcal I^l$, this expression coincides with the standard RF-GAP proximity~\cite{rhodesGeometryAccuracyPreservingRandom2023}. For unlabeled points, however, the bootstrap multiplicity $c_j(t)$ is not available, since these samples are not used during training. To extend the definition, we treat each unlabeled point as a \emph{phantom training sample}: it is routed through the forest exactly as a labeled sample, but its contribution on the target side is replaced by a simple deterministic surrogate derived from labeled bootstrap statistics~\cite{breiman1996bagging}. Concretely, unlabeled points act as out-of-bag queries while contributing as unit-weight in-bag targets, i.e., $c_j(t)=1$ for $j \in \mathcal I^u$. This choice preserves the scale of RF-GAP contributions while extending the weighted leaf co-occurrence mechanism to unlabeled targets. We further justify this surrogate by showing that it provides an optimal approximation, in a least-squares sense, of the labeled multiplicities under standard RF assumptions (Lemma~\ref{lem:rf_gap_surrogate}).

\subsection{Learning correspondences via semantic transport}\label{subsec:transport}

For clarity, we focus on the balanced setting with $n = m$ and assume full semantic overlap between the source and target domains, i.e. $\mathcal{C}_A=\mathcal{C}_B=\{1,\ldots,C\}$. Extensions to imbalanced settings and partially overlapping domains are discussed in Appendix~\ref{app:fosta_extensions}.
To establish correspondences across domains, we seek a shared representation that abstracts away domain-specific variability while preserving class-level structure. Starting from the label-informed RF-GAP affinity matrix $W_*$ in domain $*\in\{A,B\}$ (Eq.~\ref{eq:intradomain_affinity}), we construct a \emph{semantic profile} for each sample that summarizes its relationships to labeled data across classes. Specifically, for any point $\mathbf{x}_i^* \in \mathcal{X}_*$, we define a $C$-dimensional semantic vector
\begin{equation}\label{eq:semantic_vector}
\tilde{S}_*(i,:) \;=\; \frac{S_*(i,:)}{\|S_*(i,:)\|_2},
\qquad
S_*(i,c) \;=\; \frac{1}{p_c^*}
\sum_{j \in \mathcal{I}_*^c} (W_*)_{ij}.
\end{equation}
Here, $\mathcal{I}_*^c$ denotes the set of samples in domain $*$ labeled with class $c$, and $p_c^*=|\mathcal{I}_*^c|/|\mathcal{I}_*^l|$ is the estimated class prior. The factor $1/p_c^*$ mitigates class imbalance, while the $\ell_2$ normalization makes the representation scale-invariant, so alignment depends on the relative shape of the class-affinity profile rather than its total magnitude. Each sample is therefore embedded into a shared semantic space, regardless of its original domain. Because we aggregate affinities only to labeled samples (Eq.~\ref{eq:semantic_vector}), the unlabeled--unlabeled portion of $W_*$ does not define the label-supervised semantic signal. Instead, it helps preserve a coherent affinity structure for the downstream joint embedding once labeled and unlabeled samples are embedded together. Intuitively, samples with similar semantic roles are mapped close to each other, while samples associated with different classes are separated. Unlike MALI, which applies DPT to an unsupervised geometry before semantic matching, FoSTA derives semantic representations directly from its supervised forest geometry.

To align samples across domains, one could naively rely on a cross-domain $k$-NN graph. However, formulating alignment as an OT assignment has been shown to provide greater robustness by enforcing global structural consistency across domains~\cite{duqueDiffusionTransportAlignment2023, duqueManifoldAlignmentLabel2023}. Given normalized semantic representations $\tilde{S}_A, \tilde{S}_B \in \mathbb{R}^{n \times C}$, we define the OT problem
\[
\mathbf{T}^\star
=
\arg\min_{\mathbf{T} \in \mathcal{U}}
\;\langle \mathbf{T}, \mathbf{C} \rangle,
\qquad
\mathcal{U}
=
\left\{
\mathbf{T} \in \{0,1\}^{n \times n}
\;\middle|\;
\mathbf{T}\mathbf{1} = \mathbf{1},\;
\mathbf{T}^\top \mathbf{1} = \mathbf{1}
\right\},
\]
where $\mathbf{C}_{ij} = \| \tilde{S}_A(i,:) - \tilde{S}_B(j,:) \|_2^2$ is the pairwise cost matrix, and $\langle \cdot, \cdot \rangle$ denotes the Frobenius inner product. The optimal coupling $\mathbf{T}^\star$ defines a one-to-one correspondence between samples in the two domains. Solving this OT problem exactly is computationally prohibitive for large-scale datasets, as it requires explicit construction of the dense cost matrix $\mathbf{C}$, leading to at least quadratic time and memory complexity. We therefore approximate $\mathbf{T}^\star$ using HiRef~\cite{halmosHierarchicalRefinementOptimal2025}, which recursively partitions the two domains and solves a sequence of smaller, localized transport problems. This divide-and-conquer strategy yields quasilinear complexity while producing a sparse, bijective Monge map.

Finally, note that on the unit hypersphere induced by $\ell_2$ normalization in Eq.~\ref{eq:semantic_vector}, minimizing squared Euclidean distance is equivalent to maximizing cosine similarity (Lemma~\ref{lemma:l2_norm}).
This equivalence allows HiRef to leverage fast low-rank factorizations of the Euclidean cost~\cite{scetbonLowRankSinkhornFactorization2021} without explicitly forming $\mathbf{C}$, effectively performing cosine-based semantic matching while retaining the scalability of the Euclidean-based HiRef solver. The resulting coupling $\mathbf{T} \approx \mathbf{T}^\star$ thus provides an efficient and robust cross-domain correspondence.

\subsection{Cross-domain affinities through correspondence propagation}\label{subsec:cross-domain}

While one could directly populate the off-diagonal blocks of the joint affinity matrix using the binary assignment encoded by $\mathbf{T}$ (i.e., setting entries to either 0 or 1), such a construction fails to capture meaningful similarities between pairs that are not directly matched. That is, it does not provide a notion of graded cross-domain similarity beyond the hard assignment.

To obtain smoother and more informative cross-domain affinities, we propagate intra-domain affinities in Eq.~\ref{eq:intradomain_affinity} across the estimated correspondences by transferring local geometric structure from each domain to the other:
\[
W_{AB} = W_A \mathbf{T},
\qquad
W_{BA} = W_B\mathbf{T}^\top .
\]
As a result, two samples from different domains are assigned a high affinity if they are either directly matched by $\mathbf{T}$ or if they are close to matched samples within their respective domains. This construction yields cross-domain similarities that are consistent in scale with intra-domain affinities and better reflect the underlying manifold structure across domains.

\subsection{Joint embedding}\label{subsec:joint_embedding}

The final step of our framework projects both domains into a shared latent space. Given the label-informed cross-domain affinities, a natural approach would be to perform a barycentric projection, i.e., to map samples from one domain onto the other using weighted averages of neighboring coordinates. However, such projections are highly sensitive to the curse of dimensionality and operate directly in the original feature space. This is undesirable in our setting, as the raw feature space may be noisy and may not faithfully reflect the underlying phenomenon of interest. Instead, we rely on nonlinear, diffusion-based DR to embed both domains into a shared low-dimensional representation. Specifically, we use Landmark PHATE~\cite{moonVisualizingStructureTransitions2019}, a scalable variant of PHATE, which has been shown to work effectively with forest-derived affinities in guided representation learning~\cite{rhodes2026gaining, aumonRandomForestAutoencoders2025}.

\subsection{Complexity analysis}\label{subsec:complexity}
FoSTA is designed to scale efficiently with the number of samples. Its sparse leaf-based RF-GAP affinities, hierarchical OT, and Landmark PHATE avoid dense pairwise computations, resulting in near-linear time and memory complexity. A detailed complexity analysis is provided in Appendix~\ref{app:complexity}, with empirical scaling results reported in Appendix~\ref{app:empirical_scaling}.

\section{Empirical results}
\label{sec:results}

Code to reproduce all experiments is publicly available at \url{https://github.com/aumona/FoSTA-experiments}.

\subsection{Multimodal alignment for data integration}
\label{subsec:real_multimodal}

We compare FoSTA with four label-supervised manifold alignment baselines: kernel manifold alignment with radial basis function (KEMArbf~\cite{tuiaKernelManifoldAlignment2016}) and its linear variant (KEMAlin~\cite{wang2011heterogeneous}), MALI~\cite{duqueManifoldAlignmentLabel2023}, and Pamona~\cite{caoManifoldAlignmentHeterogeneous2021}. As our evaluation focuses on joint visualization, we use two-dimensional embeddings for all methods. Other methods discussed in Section~\ref{subsec:manifold_alignment} are not included because they are either fully unsupervised or assume additional cross-domain information, such as known anchors or shared feature spaces. Implementation and hyperparameter settings are provided in Appendix~\ref{app:implementation}.

We consider four real-world multimodal alignment tasks in which paired domains provide distinct views of the same underlying observations: audio and video from AVE~\cite{tianAudioVisualEventLocalization2018}, RGB and depth images from RGB-D~\cite{laiLargescaleHierarchicalMultiview2011}, photos and human sketches from Sketchy~\cite{sangkloySketchyDatabaseLearning2016}, and accelerometer and gyroscope measurements from HAR~\cite{anguitaPublicDomainDataset2013}. We evaluate the methods under partial label supervision by masking a subset of labels during alignment and evaluating label transfer on the held-out labels. Table~\ref{tab:real_multimodal} reports three complementary criteria: label-transfer accuracy~\cite{caoUnsupervisedTopologicalAlignment2020,demetci2022scot}, domain mixing through the Alignment Score (AS~\cite{butler2018integrating}), and correspondence recovery through the Fraction of Samples Closer Than the True Match (FOSCTTM~\cite{liuJointlyEmbeddingMultiple2019}). Full experimental details are provided in Appendix~\ref{app:multimodal_extended}.

\input{tables/real_multimodal}

Across the four real-world tasks, FoSTA shows consistently strong label transfer and correspondence recovery across modalities and feature representations. It achieves the best accuracy on AVE, both RGB-D feature representations, and HAR, as well as the best top-$k$ accuracy on AVE, RGB-D with ResNet18, and Sketchy with DINOv2-B. FoSTA also achieves the best FOSCTTM on both RGB-D representations and HAR, while remaining competitive on Sketchy. Other methods perform better on individual criteria, notably KEMArbf for correspondence recovery on AVE and Sketchy and MALI for AS on several datasets. FoSTA nevertheless remains robust across alignment settings, consistently ranking among the top two methods in both label-transfer accuracy and FOSCTTM, and among the top three in AS in all settings except AVE. As further discussed in Appendix~\ref{app:uci_full_section}, we place greater emphasis on label transfer and FOSCTTM, which directly measure semantic preservation and ground-truth correspondence recovery, respectively, rather than domain mixing alone. The UCI experiments (Appendices~\ref{app:uci_datasets} and~\ref{app:uci_general}) complement these real-world results by isolating different forms of cross-domain perturbation, where FoSTA achieves consistently strong accuracy and FOSCTTM across most settings. Finally, the ablations in Appendices~\ref{app:uci_general_ablation} and~\ref{app:uci_ablation_labels} show that these gains are robust to secondary design choices and partial label availability, and are primarily driven by the upstream forest geometry.

\subsection{Batch effect removal in single-cell genomics}
\label{subsec:singlecell}

\begin{figure} 
    \centering
     \includegraphics[width = \textwidth]{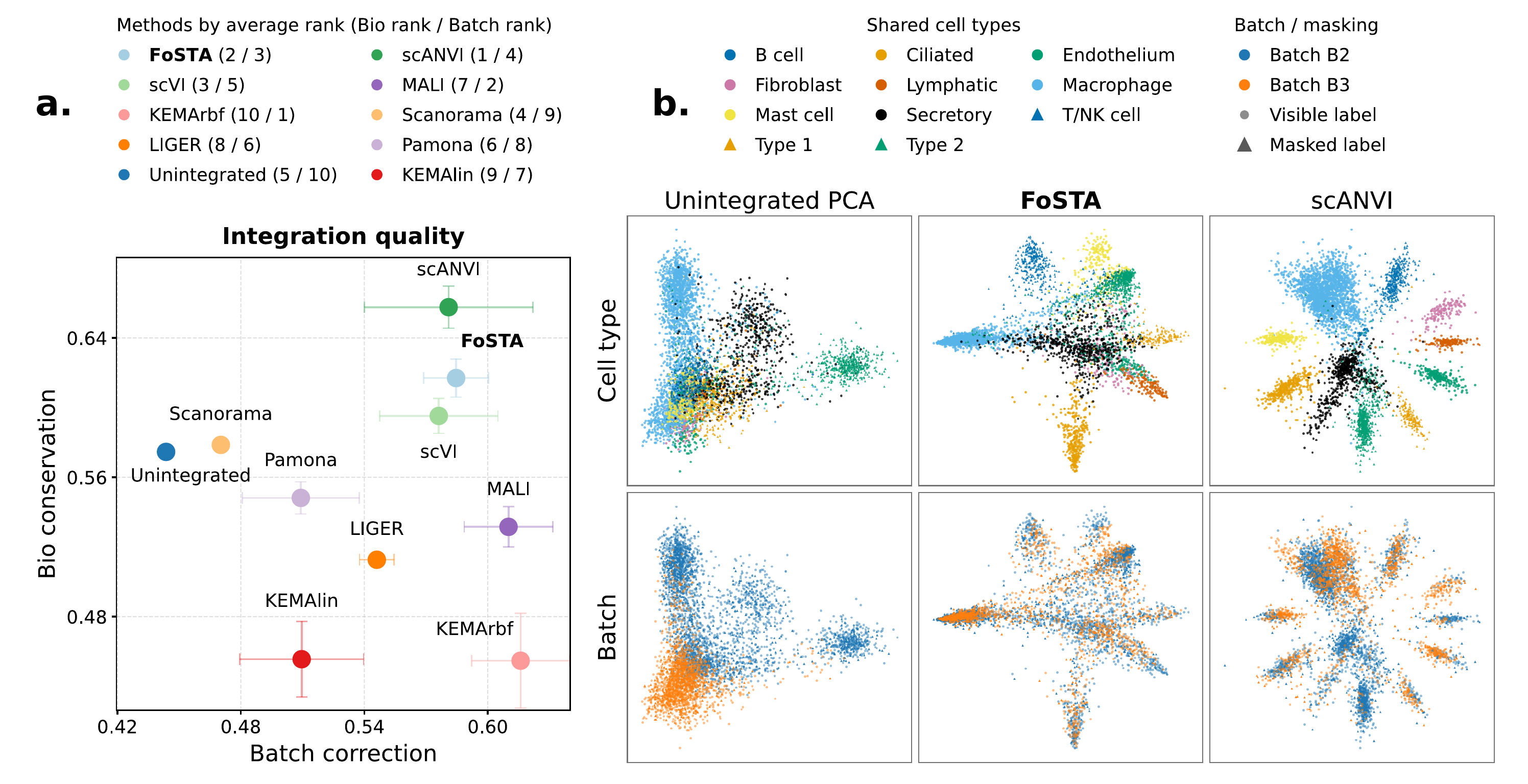}
     \caption{\textbf{Results of batch effect removal on lung single-cell data with 20\% label masking~\cite{vieirabragaCellularCensusHuman2019}.}\\
\textbf{a.} Biological preservation (Bio) vs.\ batch correction (Batch) scores~\cite{lueckenBenchmarkingAtlaslevelData2022}, averaged across 36 batch pairs and five seeds. Error bars represent seed variability, computed as the average within-pair standard deviation across the 36 batch pairs.
\textbf{b.}
Embeddings of the B2/B3 batch integration task. Left column: data before correcting for batch effects, visualized with PCA. Center column: resulting embedding after applying FoSTA. Right column: resulting embedding after applying scANVI. The plots are colored by cell type (top) and batch (bottom). FoSTA mixes the batches well while preserving the biological structure of the data.
}
    \label{fig:singlecell_main}
\end{figure}

We now consider the task of batch effect removal in single-cell genomics, where the goal is to combine data batches of a single modality (i.e., sharing the same feature space). Different batches may contain data collection artifacts called batch effects, which are due to differences in environment or experimental setup (e.g., protocol, time, instrument, or donors). Batch effects are widespread in single-cell data and must be removed so that downstream analyses are not impacted by them~\cite{haghverdiBatchEffectsSinglecell2018, Stuart2019Seurat, Tran2020Benchmark, lueckenBenchmarkingAtlaslevelData2022, Leek2010BatchEffects}.

We use the lung atlas dataset from~\citet{vieirabragaCellularCensusHuman2019}, following the benchmark setting of~\citet{lueckenBenchmarkingAtlaslevelData2022}. We evaluate 36 real within-family batch pairs, which may contain only partially overlapping cell types. To evaluate the methods under partial label supervision, we independently mask 20\% of the cell-type labels in each batch during alignment. Full dataset, preprocessing, and masking details are provided in Appendix~\ref{app:singlecell_protocol}.

We compare FoSTA with eight prominent batch correction and manifold alignment methods. These include Scanorama~\cite{hieEfficientIntegrationHeterogeneous2019} and LIGER~\cite{welchSinglecellMultiomicIntegration2019}, which are unsupervised methods relying on local anchoring or matrix-factorization strategies; scVI and its label-supervised variant scANVI~\cite{xuProbabilisticHarmonizationAnnotation2021}, which are probabilistic deep generative models; MALI~\cite{duqueManifoldAlignmentLabel2023} and Pamona~\cite{caoManifoldAlignmentHeterogeneous2021}, which are label-supervised manifold alignment methods based on optimal transport; and the linear and RBF variants of KEMA~\cite{wang2011heterogeneous, tuiaKernelManifoldAlignment2016}, which are label-supervised kernel-based approaches. Following~\citet{lueckenBenchmarkingAtlaslevelData2022}, we evaluate biological conservation and batch correction. We note that these metrics differ from those used in Section~\ref{subsec:real_multimodal}, since there are no known cell-level correspondences between batches. Full evaluation details are provided in Appendix~\ref{app:singlecell_metrics}, implementation settings in Appendix~\ref{app:implementation}, and complete quantitative results in Appendix~\ref{app:singlecell_full_results_table}.

As discussed in~\citet{lueckenBenchmarkingAtlaslevelData2022}, batch correction can come at the expense of biological conservation by removing meaningful biological variation associated with batch differences. Our results follow this pattern (Fig.~\ref{fig:singlecell_main}a). With the exception of FoSTA, scVI, scANVI, and, to a lesser extent, Scanorama, methods that improve batch correction over the unintegrated representation generally achieve lower biological conservation. FoSTA achieves a strong balance between the two objectives, ranking second in biological conservation and third in batch correction, and tying with scANVI for the best overall average rank. scANVI achieves the highest biological conservation, while FoSTA provides stronger batch correction. Conversely, MALI and KEMArbf achieve high batch correction but substantially lower biological conservation. This balance remains robust under reduced label availability, with FoSTA maintaining strong biological conservation and batch correction when 50\% of the labels are masked (Appendix~\ref{app:singlecell_mask_ablation}). Overall, FoSTA remains competitive with methods specifically designed for single-cell batch correction such as scVI/scANVI while also supporting the heterogeneous feature spaces considered in our multimodal alignment experiments.

To complement the aggregate metrics, Fig.~\ref{fig:singlecell_main}b shows a representative B2/B3 batch pair. Beyond reducing batch separation, FoSTA reveals continuous structure between cell populations that is consistent with known lung organization. In particular, the endothelial populations (Endothelium/BEC and Lymphatic/LEC) and the stromal population (Fibroblast) form a connected neighborhood, with LECs at its edge. This captures both the relationship between blood and lymphatic endothelium and their association with fibroblasts~\cite{travaglini2020molecular}. In scANVI, these populations form more isolated clusters, with LECs separating BECs from fibroblasts.

FoSTA also places the epithelial Secretory population at the center of a broader structure, connecting the BEC–LEC–Fibroblast neighborhood to other epithelial populations (Ciliated, Type 1, and Type 2). This agrees with known interactions between epithelial, endothelial, and stromal populations in the lung~\cite{raredonSinglecellConnectomicAnalysis2019}, as well as established relationships between secretory, ciliated, and alveolar cells~\cite{goldfarbmurenDissectingCellularSpecificity2020,basilHumanDistalAirways2022}. Under scANVI, Secretory instead forms a more isolated branch, making these relationships less apparent. Thus, FoSTA does more than separate cell types and mix batches: it preserves continuous biological organization between cell populations that is less apparent in the more specialized scANVI representation. Additional visualizations including scVI, MALI, and KEMArbf are provided in Appendix~\ref{app:singlecell_plot_extended}.

Overall, FoSTA achieves a strong balance between biological conservation and batch correction while recovering meaningful class-aware biological structure. Although methods such as scVI and scANVI are specifically designed for single-cell integration, FoSTA remains competitive while also applying to heterogeneous modalities with unrelated feature spaces.

\section{Conclusion and discussion}\label{sec:discussion}

We proposed FoSTA, a forest-guided framework for label-supervised manifold alignment that learns task-aware geometry independently within each domain and uses this shared semantic structure to infer cross-domain correspondences without predefined anchors. By extending RF-GAP affinities to partially labeled data and combining them with scalable semantic transport, FoSTA provides a general framework for aligning domains with potentially unrelated feature spaces.

Across real-world multimodal benchmarks spanning audio–video, RGB–depth, photo–sketch, and sensor modalities, FoSTA achieves strong label transfer and correspondence recovery compared with established weakly supervised alignment methods. On real single-cell batch alignment tasks, FoSTA also achieves a strong balance between biological conservation and batch correction, remaining competitive with specialized methods such as scVI and scANVI. The resulting embeddings preserve continuous relationships between biologically related cell populations that are less apparent in more strongly clustered representations. Together with our controlled perturbation experiments and ablations, these results show that FoSTA’s alignment gains are primarily driven by the upstream forest geometry.

Beyond its empirical performance, FoSTA accommodates partial label availability, unequal domain sizes, and partially overlapping class sets, while its sparse forest affinities and hierarchical transport provide favorable scaling with the number of samples. Its main limitations remain its dependence on the quality of the underlying forest geometry and the current restriction to pairwise alignment. Future work could improve robustness to challenging transformations through modern oblique forests and extend the framework to multiple domains. Overall, our results establish task-aware forest geometry as an effective foundation for weakly supervised alignment across heterogeneous multimodal data and specialized applications such as single-cell batch correction.

\section*{AI use statement}
In this work, we used generative AI tools to assist with method implementation and the development of experimental and result-compilation pipelines. We additionally used generative AI tools to assist with drafting and polishing parts of the manuscript. All AI-assisted code was reviewed and tested by the authors, and all reported results were independently verified. We reviewed all AI-assisted work and take responsibility for the final content of this work, including text, code, and artifacts produced with the aid of generative AI.

\section*{Broader impacts}

FoSTA provides a general framework for aligning heterogeneous datasets using shared semantic annotations without requiring known sample correspondences or shared feature spaces. This may facilitate the joint analysis and visualization of complementary data sources that are otherwise difficult to compare, with potential applications in multimodal and biomedical data analysis. In particular, its ability to operate with partial label information and partially overlapping classes may be useful in settings where annotations are costly or incomplete, such as single-cell genomics and other biological applications.

Like many machine learning pipelines operating on sensitive biological or clinical measurements, FoSTA may inherit privacy and fairness concerns from the underlying data and annotations. In particular, biomedical datasets may contain patient-derived information whose use requires appropriate consent, anonymization, and secure handling. Moreover, because FoSTA uses class labels to guide the learned geometry, biases or inaccuracies in these annotations may be reflected in the resulting alignment and representation. These concerns are not unique to FoSTA but should be considered when applying supervised representation learning to sensitive data. Future work could explore privacy-preserving extensions and robustness to demographic, annotation, or sampling biases in partially labeled biomedical datasets.

\begin{ack}
This research was supported in part by the Ministry of Health and Social Services (Quebec) in collaboration with the Centre intégré de santé et de services sociaux Centre-Sud-de-l'Île-de-Montréal [A.A.], the Outstanding PhD Candidate Scholarship from the Institut des sciences mathématiques [A.A.], Canada CIFAR AI Chair [G.W.], NSERC Discovery grant 03267 [G.W.], NIH grant R01GM135929 [G.W.], NSF grant DMS-2327211 [G.W.], NSF grant 221232 [K.M], NIH grant 1R15HL168697 [K.M.], FRQNT Doctoral research scholaship [M.L.], and the IVADO Visiting Scholar program [K.M]. The authors are solely responsible for the content of this work, and the views expressed in it do not necessarily reflect those of the funding agencies. We sincerely thank the anonymous reviewers and Area Chairs for their constructive feedback and efforts throughout the review process, which helped us improve the paper.

\end{ack}

{
\small

\bibliographystyle{unsrtnat}
\bibliography{references_zotero, Other_references}

}

\clearpage
\appendix

\section{Justification of the semi-supervised RF-GAP target surrogates}
\label{app:semisup_rfgap}

We justify the surrogate target-side multiplicities used in the semi-supervised RF-GAP construction. Since unlabeled target points are not included in the bootstrap samples used to train the forest, their in-bag multiplicities are unobserved and must be replaced by deterministic surrogates.

We work under standard assumptions for Random Forest training: each tree is grown independently using bootstrap sampling with replacement from the labeled subset, and each bootstrap sample has the same size as the labeled dataset.

\begin{lemma}[Optimal constant surrogate for target-side multiplicities]
\label{lem:rf_gap_surrogate}
Fix a domain $\mathcal X$ and a tree $t \in \mathcal T$. Let $\mathcal I^l$ denote the labeled index set, and let $c_k(t)$ be the bootstrap multiplicity of $\mathbf x_k$ in tree $t$. Consider the problem of approximating the labeled multiplicities by a constant $c \in \mathbb{R}$ in the least-squares sense:
\[
c(t)
=
\arg\min_{c \in \mathbb R}
\sum_{k \in \mathcal I^l}
\bigl(c_k(t)-c\bigr)^2.
\]
Then the unique minimizer satisfies
$
c(t) = 1.
$
\end{lemma}

\begin{proof}
The objective is strictly convex in $c$, and its minimizer is given by the empirical mean:
\[
c(t)
=
\frac{1}{|\mathcal I^l|}
\sum_{k \in \mathcal I^l} c_k(t).
\]
Under bootstrap sampling with replacement, the total multiplicity equals the sample size, yielding
\[
\sum_{k \in \mathcal I^l} c_k(t) = |\mathcal I^l|\qquad\Longrightarrow \qquad c(t)=1.
\]
\end{proof}
This result shows that the unit weight is the exact empirical $L^2$-optimal constant surrogate for missing target-side multiplicities, which justifies its use in Eq.~\ref{eq:intradomain_affinity} for unlabeled targets.


\section{Extension to imbalanced domain sizes and partial matching}\label{app:fosta_extensions}

We focused on the balanced setting, assuming full overlap between the two domains. In practice, however, many applications involve imbalanced domain sizes ($n \neq m$). Several strategies can be employed to extend our framework to this more general setting.

Recall that in the balanced case, correspondence matching is formulated using HiRef, which seeks a Monge map between domains and therefore enforces a bijection (Section~\ref{subsec:transport}). In the imbalanced setting, this naturally generalizes to seeking a \emph{surjection} from the larger domain onto the smaller one. Without loss of generality, assume $n > m$, and that the semantic structures present in domain $B$ are contained within domain $A$. To satisfy the balanced assignment constraint required by HiRef, we first augment the smaller domain by duplicating samples from domain $B$ in the semantic space until both domains contain the same number of points. HiRef is then applied to these augmented domains, producing an intermediate balanced coupling corresponding to a bijective assignment between samples from domain $A$ and the augmented samples from domain $B$.

Since duplicated samples in the augmented domain correspond to the same original observation in $B$, the resulting coupling can subsequently be condensed back to the original domain sizes, yielding a sparse coupling matrix $T \in \mathbb{R}^{n \times m}$. By construction, each sample from domain $B$ remains matched to exactly one sample from domain $A$, while multiple samples from $A$ may now map to the same sample in $B$. The final transport therefore defines a surjective assignment from the larger domain onto the smaller one, while preserving the hard one-to-one structure imposed by HiRef during the intermediate balanced optimization. This is the strategy adopted in our single-cell alignment experiments involving imbalanced real batches.

In practice, the assumption of full overlap between domains may not hold. For example, the label spaces may only partially overlap, leading to domain-specific structures that should not be matched across domains. In this setting, we first subset each domain to retain only samples that are plausibly matchable to points in the other domain, and then perform the surjective matching procedure on these subsets while leaving domain-specific points unmatched in the resulting transport plan.

When labels are fully available, the shared subset can be directly inferred from the intersection of the label spaces across domains. When labels are only partially observed, and the label spaces do not fully overlap, stronger assumptions about the missing-label mechanism become necessary. In particular, one must decide whether unlabeled points are expected to belong to the shared label space or instead correspond to domain-specific structures. In the former case, unlabeled samples should be included in the OT problem; in the latter, they should be excluded.

To address this ambiguity, we implemented an automatic strategy based on the underlying forest classifier. Unlabeled samples are first assigned pseudo-labels using the forest out-of-bag (OOB) predictions. If the predicted label belongs to the shared label space, the sample is included in the OT matching; otherwise, it is excluded. While this heuristic provided satisfactory empirical behavior in our single-cell alignment tasks with label masking and potential batch-specific cell types (Section~\ref{subsec:singlecell}), we leave a more principled treatment of this problem for future work.






\section{Equivalence between Euclidean and cosine costs under quadratic normalization}
\label{app:l2_normalization_lemma}

We justify the $\ell_2$ normalization in Eq.~\ref{eq:semantic_vector} by showing that squared Euclidean distances between normalized semantic vectors are equivalent to cosine distances computed on the original (unnormalized) semantic representations. This ensures compatibility with Euclidean OT solvers such as HiRef, while preserving the cosine-based geometry commonly used in prior work~\cite{duqueDiffusionTransportAlignment2023, duqueManifoldAlignmentLabel2023}.

\begin{lemma}\label{lemma:l2_norm}
Let $S_*(i,:), S_*(j,:) \in \mathbb{R}^C$ be two nonzero semantic vectors, and define their $\ell_2$-normalized versions
\[
\tilde{S}_*(i,:) = \frac{S_*(i,:)}{\|S_*(i,:)\|_2},
\qquad
\tilde{S}_*(j,:) = \frac{S_*(j,:)}{\|S_*(j,:)\|_2}.
\]
Then,
\[
\|\tilde{S}_*(i,:) - \tilde{S}_*(j,:)\|_2^2
=
2\left(
1 - 
\frac{\langle S_*(i,:), S_*(j,:) \rangle}
{\|S_*(i,:)\|_2 \, \|S_*(j,:)\|_2}
\right).
\]
In particular, the squared Euclidean distance between normalized semantic vectors is proportional to the cosine distance between the original semantic vectors.
\end{lemma}

\begin{proof}
By construction,
\[
\|\tilde{S}_*(i,:)\|_2 = \|\tilde{S}_*(j,:)\|_2 = 1.
\]
Expanding the squared Euclidean distance yields
\[
\|\tilde{S}_*(i,:) - \tilde{S}_*(j,:)\|_2^2
=
\|\tilde{S}_*(i,:)\|_2^2 + \|\tilde{S}_*(j,:)\|_2^2
- 2 \langle \tilde{S}_*(i,:), \tilde{S}_*(j,:) \rangle
=
2 - 2 \langle \tilde{S}_*(i,:), \tilde{S}_*(j,:) \rangle.
\]
Substituting the definition of the normalized inner product gives
\[
\langle \tilde{S}_*(i,:), \tilde{S}_*(j,:) \rangle
=
\frac{\langle S_*(i,:), S_*(j,:) \rangle}
{\|S_*(i,:)\|_2 \, \|S_*(j,:)\|_2},
\]
which yields the result.
\end{proof}

\section{Extended complexity analysis}\label{app:complexity}

Without loss of generality, assume balanced domains with $N=n=m$ samples each. FoSTA consists of three main stages: (i) computing label-informed RF-GAP affinities within each domain, (ii) estimating a cross-domain transport plan with HiRef in semantic space, and (iii) producing a joint low-dimensional embedding with Landmark PHATE.

Computing RF-GAP within a domain has time complexity $\mathcal{O}\!\left(T\,N\log N \cdot p \;+\; T N (h+\ell)\right)$, where $T$ is the number of trees, $p$ is the number of candidate features evaluated per split, $h \approx \log N$ is the average tree height, and $\ell$ is the average number of same-leaf interactions per sample and per tree~\cite{louppe2014understanding, hassine2019important, aumon2026revisitingforestproximitiessparse}. The associated memory footprint scales as $\mathcal{O}\!\left(NT(1+\ell)\right)$ for storing the sparse leaf-based interactions used to assemble affinities~\cite{aumon2026revisitingforestproximitiessparse}. HiRef  computes an approximately bijective Monge map in log-linear time and linear space in $N$~\cite{halmosHierarchicalRefinementOptimal2025}. Finally, Landmark PHATE with $M$ landmarks builds landmark relationships at cost $\mathcal{O}(2NM)$ and stores them in $\mathcal{O}(2NM)$ memory, followed by diffusion and embedding on the landmark set, which costs $\mathcal{O}(M^3)$~\cite{moonVisualizingStructureTransitions2019, demaineMultidimensionalScalingApproximation2021}.

In general, $T$, $p$, and $\ell$ are treated as small constants relative to $N$, and $M \ll N$ is set to $M=2000$ by default~\cite{moonVisualizingStructureTransitions2019}. Under these assumptions, FoSTA has effectively near-linear runtime and memory scaling in the number of samples. In contrast, MALI, Pamona and related OT-based alignment methods rely on full DPT and/or explicit dense cross-domain cost matrices, leading to at least quadratic time and space complexity in $N$.

\section{Empirical scaling validation}\label{app:empirical_scaling}

To complement the theoretical analysis in Appendix~\ref{app:complexity}, we include an empirical scalability study on RGB-D (DINOv2-B features) and Sketchy (ResNet18 features). See Appendix~\ref{app:real_multimodal_datasets} for dataset details. Table~\ref{tab:empirical_scaling} reports runtime and peak memory for computing the joint embedding with 6k, 12k, and 24k samples.

FoSTA exhibits an approximately linear increase in both runtime and memory as the dataset size grows. MALI is faster at smaller sample sizes due to FoSTA’s additional computational overhead, but its runtime grows substantially faster as the number of samples increases. The difference is more pronounced in memory usage: at 24k samples, MALI requires approximately 29 GB on both datasets, compared with 5.6 GB and 4.1 GB for FoSTA on RGB-D and Sketchy, respectively. Overall, these empirical results are consistent with the theoretical complexity analysis in Appendix~\ref{app:complexity}, showing that FoSTA maintains more favorable scaling as the dataset size increases.

\begin{table}[ht]
{\small
\caption{\textbf{Runtime and peak-memory scaling across datasets and methods.} Values are means $\pm$ sample standard deviations across successful seed repetitions. Columns are grouped by method, with subcolumns giving the total number of samples across both domains. Runtime is in seconds and memory is in GB (1,024 MB per GB).}
}
\label{tab:empirical_scaling}
\vspace{6pt}
\centering
\small
\setlength{\tabcolsep}{2pt}
\renewcommand{\arraystretch}{1.0}

\begin{tabular}{ll c c c c c c}
\toprule
 &  & \multicolumn{3}{c}{FoSTA} & \multicolumn{3}{c}{MALI} \\
\cmidrule(lr){3-5} \cmidrule(lr){6-8}
Metric & Dataset & \makebox[1.5cm][c]{6,000} & \makebox[1.5cm][c]{12,000} & \makebox[1.5cm][c]{24,000} & \makebox[1.5cm][c]{6,000} & \makebox[1.5cm][c]{12,000} & \makebox[1.5cm][c]{24,000} \\
\midrule
\multirow{2}{*}{Runtime (s)} & RGB-D & \makebox[1.5cm][c]{26.24{\tiny $\pm$0.20}} & \makebox[1.5cm][c]{50.76{\tiny $\pm$1.24}} & \makebox[1.5cm][c]{100.65{\tiny $\pm$0.46}} & \makebox[1.5cm][c]{5.51{\tiny $\pm$0.13}} & \makebox[1.5cm][c]{14.99{\tiny $\pm$0.22}} & \makebox[1.5cm][c]{  79.99{\tiny $\pm$11.45}} \\
 & Sketchy & \makebox[1.5cm][c]{39.43{\tiny $\pm$1.03}} & \makebox[1.5cm][c]{74.10{\tiny $\pm$1.45}} & \makebox[1.5cm][c]{149.22{\tiny $\pm$2.74}} & \makebox[1.5cm][c]{5.77{\tiny $\pm$0.08}} & \makebox[1.5cm][c]{15.99{\tiny $\pm$0.26}} & \makebox[1.5cm][c]{77.36{\tiny $\pm$1.37}} \\
\midrule
\multirow{2}{*}{Memory (GB)} & RGB-D & \makebox[1.5cm][c]{2.03{\tiny $\pm$0.22}} & \makebox[1.5cm][c]{3.18{\tiny $\pm$0.17}} & \makebox[1.5cm][c]{5.55{\tiny $\pm$0.06}} & \makebox[1.5cm][c]{3.36{\tiny $\pm$0.21}} & \makebox[1.5cm][c]{7.91{\tiny $\pm$0.01}} & \makebox[1.5cm][c]{29.30{\tiny $\pm$0.20}} \\
 & Sketchy & \makebox[1.5cm][c]{1.62{\tiny $\pm$0.17}} & \makebox[1.5cm][c]{2.45{\tiny $\pm$0.21}} & \makebox[1.5cm][c]{4.07{\tiny $\pm$0.18}} & \makebox[1.5cm][c]{3.19{\tiny $\pm$0.07}} & \makebox[1.5cm][c]{7.85{\tiny $\pm$0.23}} & \makebox[1.5cm][c]{29.01{\tiny $\pm$0.18}} \\
\bottomrule
\end{tabular}
\end{table}

\section{Implementation and hyperparameter settings}
\label{app:implementation}

All experiments were conducted on a MacBook Pro equipped with an Apple M4 Pro processor (12 CPU cores: 8 performance and 4 efficiency cores) and 24 GB of unified memory, running macOS 26.6.2 (Darwin 25.6.0).

Unless otherwise specified, we use the default hyperparameters from the official implementations of all methods. For FoSTA and MALI, we use a PHATE diffusion scale of $t=2$ throughout all experiments. For Pamona, we use the label-informed OT correction with $\gamma=0.5$~\cite{caoManifoldAlignmentHeterogeneous2021} throughout all experiments.

For the real-world multimodal experiments in Section~\ref{subsec:real_multimodal} only, we additionally relax Pamona’s convergence parameters to $\epsilon=0.1$, \texttt{max\_iter}$=100$, and \texttt{tol}$=10^{-5}$ to obtain reasonable runtimes. All other Pamona experiments use its default convergence parameters.

FoSTA uses the standard implementations of random forests from \texttt{scikit-learn}~\cite{scikit-learn}, HiRef~\cite{halmosHierarchicalRefinementOptimal2025}, and Landmark PHATE~\cite{moonVisualizingStructureTransitions2019}.

\section{Real-world multimodal data integration: experimental details and extended results}\label{app:multimodal_extended}

\subsection{Datasets and experimental protocol}
\label{app:real_multimodal_datasets}

We evaluate FoSTA on four real-world multimodal datasets in which the two domains describe the same underlying objects or events through different modalities and share a common class-label space. Ground-truth sample correspondences are retained exclusively for evaluation and are not provided to the alignment methods.

\begin{itemize}

\item \textbf{Audio-Visual Event (AVE~\cite{tianAudioVisualEventLocalization2018}).}
The dataset contains 4,143 paired audio-video observations spanning 28 event classes, represented by 128 audio and 512 visual features. Correspondence is defined by the underlying event instance. We use the provided features and predefined train/test split: training labels are available during alignment, while test labels are masked and retained for label transfer evaluation. No additional feature normalization or dimensionality reduction is applied.

\item \textbf{RGB-D Object~\cite{laiLargescaleHierarchicalMultiview2011}.}
Paired RGB and depth observations represent physical objects from 51 classes. From the 207,920 available observations, we use at most 15,000 paired samples for each experiment. Following~\citet{duqueManifoldAlignmentLabel2023}, each modality is encoded using an ImageNet-pretrained ResNet18~\cite{heDeepResidualLearning2015}, with global average pooling of the final convolutional representation yielding 512 features per domain. We construct a class-stratified split with 50\% of the labels retained and 50\% masked, using the same split in both domains.

\item \textbf{Sketchy~\cite{sangkloySketchyDatabaseLearning2016}.}
Photographs and human-drawn sketches span 125 object classes, with multiple sketches potentially corresponding to each photograph. For each seed, we randomly select one corresponding sketch per photograph, yielding 12,500 paired observations. Following~\citet{duqueManifoldAlignmentLabel2023}, both modalities are encoded using an ImageNet-pretrained ResNet18~\cite{heDeepResidualLearning2015}, yielding 512 features per domain. Photographs are treated as the source domain, with all labels retained, while 50\% of the sketch target labels are masked within each class.

\item \textbf{Human Activity Recognition (HAR~\cite{anguitaPublicDomainDataset2013}).}
The dataset contains 10,299 paired observations spanning six activity classes. We form the two domains from accelerometer and gyroscope measurements, each represented by 384 features, with paired observations corresponding to the same activity instance. The predefined train/test split determines label availability: training labels are provided during alignment and test labels are masked. Features are standardized independently within each modality to zero mean and unit variance.

\item \textbf{Alternative image representations.}
For RGB-D and Sketchy, we additionally replace ResNet18 with DINOv2-Base~\cite{oquabDINOv2LearningRobust2024}, representing each image by its 768-dimensional CLS token. All sample selection, label masking, and evaluation procedures remain unchanged, isolating the effect of the input image representation.

\end{itemize}

To evaluate the methods under partial label supervision, HAR and AVE use their predefined train/test splits, while RGB-D uses a stratified 50\% labeled split shared across both domains. For Sketchy, all photo labels are retained and 50\% of the sketch labels are masked, following the source-to-target semi-supervised setting of~\citet{duqueManifoldAlignmentLabel2023}. Models are fit using both labeled and masked samples, with only the available labels used for supervision, and label transfer is evaluated on the held-out labels. Since Sketchy contains multiple sketches per photo, we randomly select one corresponding sketch per photo in each seed to obtain one-to-one ground-truth correspondences for evaluation. Experiments are repeated over five random seeds.

\subsection{Evaluation metrics with ground-truth correspondences}\label{app:multimodal_metrics}
Label transfer accuracy is measured by predicting masked labels using a $k$-NN classifier trained on the opposite domain embeddings~\cite{caoUnsupervisedTopologicalAlignment2020, demetci2022scot}. When labels are masked in both domains, label transfer is evaluated in both directions and the two accuracies are averaged; when a natural source and target are defined, as for Sketchy (Appendix~\ref{app:real_multimodal_datasets}) and our UCI experiments (Appendix~\ref{app:uci_full_section}), we evaluate source-to-target transfer only. Domain mixing is quantified by an alignment score (AS~\cite{butler2018integrating}) defined as $2(1-\bar{x}/k)$, where $\bar{x}$ is the average number of $k$-NN belonging to the same domain as the query sample. An AS near 2 corresponds to $\bar{x}\approx 0$, meaning that a point’s neighbors come almost entirely from the other domain; AS near 1 corresponds to $\bar{x}\approx k/2$, meaning balanced local mixing across domains; and AS near 0 corresponds to $\bar{x}\approx k$, meaning complete separation. Correspondence recovery is assessed using the Fraction of Samples Closer Than the True Match (FOSCTTM~\cite{liuJointlyEmbeddingMultiple2019}), defined as the fraction of cross-domain samples that are closer than the true correspondence (lower is better).




\section{Synthetic multimodal data integration benchmarks and ablation experiments}\label{app:uci_full_section}

\subsection{Datasets and experimental protocol}
\label{app:uci_datasets}
We use 15 supervised learning datasets from the \href{https://archive.ics.uci.edu/}{UCI Machine Learning Repository} to evaluate synthetic multimodal data integration performance in Sections~\ref{app:uci_general},~\ref{app:uci_general_ablation}, and~\ref{app:uci_ablation_labels}. Following the experimental protocol of~\citet{rhodesRandomForestSupervisedManifold2024}, we select the Balance Scale, Breast Cancer, CRX, Diabetes, E.coli (restricted to five majority classes), Flare1, Glass, Heart Disease, Heart Failure, Hepatitis, Ionosphere, Iris, Parkinson’s, Seeds, and Tic-Tac-Toe datasets. Observations with missing values are removed prior to analysis. Table~\ref{tab:datasets} summarizes each dataset, including the number of observations, features, and classes.

To construct two domains from each dataset, we consider six controlled perturbations~\cite{rhodesRandomForestSupervisedManifold2024}. The \textit{random} split randomly assigns features to either domain. The \textit{importance} split assigns the most task-relevant features, as determined by a random forest model, to the first domain, and the least important features to the second domain. The \textit{alternating importance} split assigns features to either domain by alternating by feature importance, creating balanced domains. In the \textit{add noise} split, both domains contain all of the features, but the second additionally contains Gaussian noise features ($\sigma=1$), with a signal-to-noise ratio of 1 to 10. The \textit{distort} split creates a second domain by adding Gaussian noise ($\sigma=0.5$) to the first, while the \textit{rotate} split creates a second domain by a random rotation of the first.

Unless otherwise specified, we mask 50\% of the target-domain labels to evaluate the methods in a semi-supervised setting. We use five different random seeds to generate the domains and report the averages. As in Section~\ref{subsec:real_multimodal}, we evaluate the methods using label-transfer accuracy, domain mixing (AS), and correspondence recovery (FOSCTTM). Section~\ref{app:uci_general} compares the methods across the six controlled perturbations, while Sections~\ref{app:uci_general_ablation} and~\ref{app:uci_ablation_labels} respectively evaluate the robustness of FoSTA to secondary design choices and varying levels of label availability.

\begin{table}[!ht]
\centering
\caption{\textbf{Summary of the UCI datasets used to evaluate synthetic multimodal data integration performance.} Observations with missing values were removed prior to analysis.}
\label{tab:datasets}

\small

\setlength{\tabcolsep}{1pt}

\begin{tabular}{l *{15}{>{\centering\arraybackslash}p{0.75cm}}}
& \rotatebox{75}{Balance Scale} & \rotatebox{75}{Breast Cancer} & \rotatebox{75}{CRX} & \rotatebox{75}{Diabetes}
& \rotatebox{75}{E.~coli} & \rotatebox{75}{Flare1} & \rotatebox{75}{Glass} & \rotatebox{75}{Heart Disease}
& \rotatebox{75}{Heart Failure} & \rotatebox{75}{Hepatitis} & \rotatebox{75}{Ionosphere}
& \rotatebox{75}{Iris} & \rotatebox{75}{Parkinson’s} & \rotatebox{75}{Seeds} & \rotatebox{75}{Tic-Tac-Toe} \\
\midrule
\# Obs.      & 625 & 699 & 664 & 678 & 336 & 323 & 214 & 303 & 299 & 138 & 351 & 150 & 195 & 199 & 958 \\
\# Features  & 4   & 16  & 14  & 8   & 8   & 10  & 9   & 13  & 10  & 15  & 34  & 4   & 22  & 7   & 9   \\
\# Classes   & 3   & 2   & 2   & 2   & 5   & 3   & 6   & 5   & 2   & 2   & 2   & 3   & 2   & 3   & 2   \\
\bottomrule
\end{tabular}
\end{table}

\subsection{Performance under controlled perturbations}\label{app:uci_general}

\input{tables/uci_general}

In Table~\ref{tab:uci}, FoSTA achieves the best or second-best accuracy in five of the six splits (4th in rotate) and the best FOSCTTM in all but two (3rd in distort and rotate). FoSTA also ranks either second or third in AS. The rotate split remains the most challenging regime for FoSTA. This is not surprising: standard tree ensembles rely on axis-aligned splits, so a global rotation can substantially alter the induced forest geometry even when the underlying class semantics remain unchanged. These results indicate strong correspondence recovery and preservation of task-relevant semantic structure across different controlled perturbations.

We emphasize label-transfer accuracy and correspondence recovery over AS, as these metrics are directly grounded in available ground-truth information. Label transfer evaluates whether semantic information is preserved across domains, while FOSCTTM explicitly measures the recovery of true cross-domain correspondences. In contrast, AS is a heuristic that assesses domain mixing under the assumption that local neighborhoods should contain uniformly distributed samples from each domain. As a result, a method could achieve a high AS by indiscriminately mixing domains—potentially even collapsing structure—without accurately aligning shared semantic manifolds. Thus, a moderate AS does not contradict strong performance in label transfer and correspondence recovery, which provide a more direct assessment of alignment quality in our setting.

\subsection{Robustness to secondary design choices}\label{app:uci_general_ablation}

\input{tables/uci_general_ablation}

This section presents an ablation study of FoSTA across secondary architectural choices in the pipeline, following the upstream-to-downstream construction described in the Methods. Table~\ref{tab:uci_ablation} summarizes the results under the same experimental protocol as in Section~\ref{app:uci_general} on the UCI benchmark. For each variant, we report rankings across splits and metrics relative to the baseline competitors in Table~\ref{tab:uci}.

\begin{itemize}
    \item \textbf{Upstream geometry (Random Forests vs ExtraTrees vs Rotation Forests).} Replacing Random Forests with Extremely Randomized Trees (ExtraTrees~\cite{geurts2006extremely}) consistently improves performance, particularly in AS and FOSCTTM. We hypothesize that, on relatively simple prediction tasks, the additional randomness introduced by ExtraTrees leads to coarser yet informative partitions of the feature space. This produces semantic representations that better localize samples in the supervised manifold, whereas Random Forests may be overly confident despite the presence of bootstrap variability. Rotation Forests (RotF~\cite{rodriguezRotationForestNew2006}) yield similar overall performance to the default Random Forest geometry, with slightly improved accuracy and FOSCTTM on average. The improvement is most apparent under the rotation distortion, supporting the interpretation that part of FoSTA’s sensitivity in this setting comes from the axis-aligned splits of standard Random Forests rather than from the alignment procedure itself.

    \item \textbf{Affinity construction (semi-supervised RF-GAP vs KeRF).} KeRF affinities~\cite{scornet2016random} can be interpreted as a symmetric variant of RF-GAP, where pairwise similarities are still defined through leaf collisions, but without distinguishing between in-bag and out-of-bag interactions, and with leaf mass normalization computed over the full training set. KeRF yields slightly lower accuracy and AS than the default setting, while achieving comparable FOSCTTM. Overall, these results support semi-supervised RF-GAP as the default affinity construction, while showing that FoSTA remains effective with alternative forest-based affinities.

    \item \textbf{Diffusion coarse-graining (DPT vs no DPT).} Applying Diffusion Pseudotime (DPT) to the RF-GAP affinities before semantic matching provides little improvement in AS while degrading accuracy and FOSCTTM. This additional diffusion step is therefore both computationally unnecessary and detrimental to correspondence recovery. More importantly, these results suggest that the forest affinities already provide the graded, local-to-global class-aware structure needed for semantic optimal transport, without requiring additional diffusion-based preprocessing.

    \item \textbf{Alignment stage (HiRef vs dense OT).} HiRef and exact balanced Wasserstein OT with uniform masses yield comparable performance. This indicates that HiRef offers a favorable trade-off between computational efficiency and alignment quality, integrating well within the FoSTA pipeline while preserving near-linear scalability.

    \item \textbf{Affinity propagation (with vs without propagation).} Removing affinity propagation substantially degrades performance, notably reducing AS from $0.618$ to $0.069$ on average, while also worsening accuracy and FOSCTTM. This confirms that the semantic correspondences identified by optimal transport alone are insufficient: propagating them through the forest affinities is important to extend the class-level matching to the full sample geometry and obtain a coherent cross-domain alignment.

    \item \textbf{Downstream embedding (default PHATE ($t=2$) vs automatic $t$ selection vs UMAP~\cite{mcinnes2018umap} vs Laplacian Eigenmaps (LE~\cite{belkinLaplacianEigenmapsDimensionality2003}).} Automatic entropy-based scale selection in PHATE~\cite{moonVisualizingStructureTransitions2019} improves AS and slightly improves FOSCTTM, at the expense of some label-transfer accuracy. This is consistent with stronger diffusion promoting mixing across the cross-domain affinity blocks, while also smoothing local class structure. UMAP substantially improves AS and ranks first overall on this metric, but yields worse FOSCTTM, highlighting a trade-off between global domain mixing and accurate correspondence recovery, consistent with Section~\ref{subsec:real_multimodal}. LE also improves AS over the default PHATE embedding, but decreases both accuracy and FOSCTTM. 
\end{itemize}

Overall, FoSTA remains competitive across forest, transport, and embedding variants, although individual choices such as UMAP and LE shift the relative performance across metrics. Affinity propagation is a notable exception, with its removal substantially degrading alignment and confirming its importance in transferring semantic correspondences through the forest geometry. Together, these results indicate that FoSTA is generally robust to secondary design choices, while supporting that its main advantage over Euclidean-based approaches such as MALI stems from the upstream forest-guided geometry and its propagation across domains.

\subsection{Robustness to partial label availability}\label{app:uci_ablation_labels}

We evaluate the robustness of FoSTA to varying levels of label availability in the target (``weak'') domain. Following the protocol of Section~\ref{app:uci_general}, we vary the proportion of masked target labels from 20\% to 80\% in increments of 20\%, rather than fixing it at 50\% as in our previous UCI experiments.

\begin{figure} 
    \centering
     \includegraphics[width = \textwidth]{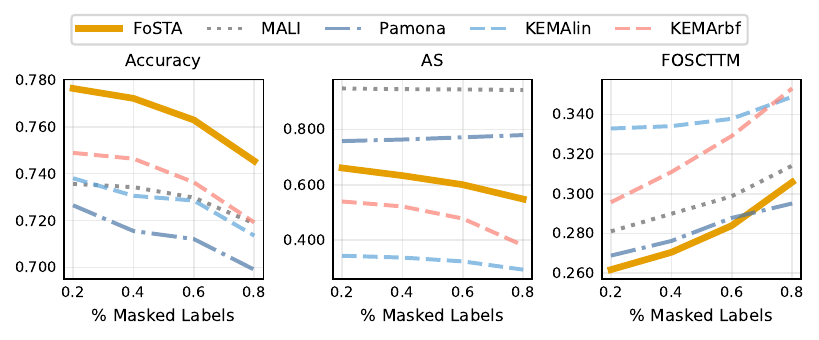}
    \caption{Aggregated performance across 15 UCI datasets as a function of the percentage of available labels in the target domain.}
    \label{fig:uci_ablation_labels}
\end{figure}

The results are reported in Fig.~\ref{fig:uci_ablation_labels}. FoSTA consistently ranks among the top methods in Accuracy and FOSCTTM across masking levels, falling to second in FOSCTTM only under the lowest label-availability regimes, while maintaining the third-best AS throughout. Performance degrades gradually as fewer labels become available, without a marked drop relative to competing methods. Overall, these results show that FoSTA remains competitive even under limited label supervision.

\section{Batch effect removal in single-cell genomics: experimental details and extended results}\label{app:singlecell}

\subsection{Datasets and experimental protocol}
\label{app:singlecell_protocol}

We use the lung atlas dataset from~\citet{vieirabragaCellularCensusHuman2019}, following the benchmark setting of~\citet{lueckenBenchmarkingAtlaslevelData2022}. The 16 batches are organized into three families: A (A1–A6), B (B1–B4), and 1–6. We consider all pairwise combinations within each family, resulting in 36 real batch-alignment tasks. Each pair is treated as two separate domains and aligned independently. Paired batches do not necessarily contain the same cell types, allowing evaluation under partial overlap of the shared semantic space.

The data are scran pool-normalized, and we retain 2,000 highly variable genes selected across each pair using the \texttt{cell\_ranger} flavor. To evaluate the methods under partial label supervision, we independently mask 20\% of the cell-type labels within each batch during alignment, while ensuring that at least 10 cells of every masked cell type remain unmasked. Masking is performed independently for each batch, pair, and random seed.

Experiments are repeated over five random seeds. As our evaluation focuses on joint visualization, all methods produce two-dimensional embeddings. Biological conservation and batch correction are computed independently for each batch pair and seed before aggregation across the 36 batch pairs and five seeds.

\subsection{Evaluation metrics}\label{app:singlecell_metrics}

The implementation of all metrics for batch correction and biological conservation is taken from the scib-metrics package~\cite{lueckenBenchmarkingAtlaslevelData2022}. All metrics are normalized and transformed appropriately so that the result lies between 0 and 1, with higher scores corresponding to better performance.

For all definitions, we assume the following notation. 
Denote the dataset by $\mathcal{D} = \{x_1, ..., x_n\}$, with $x_i \in \mathbb{R}^d$ for all $i$.
Let $X \in \mathbb{R}^{n \times d}$ denote the data matrix, with samples in the rows and $n$ denoting the number of samples.
Let $\mathcal{L} = \{1, ..., n_l\}$ denote a set of $n_l$ (encoded) labels, which could be the cell type labels or the batch label, depending on the case, and let $L \in \mathcal{L}^n$ denote the vector of labels for each sample.

\subsubsection{Biological conservation metrics}\label{app:single_bio_metrics}
Normalized mutual information (NMI~\cite{Danon2005}) and
Adjusted Rand index (ARI~\cite{RandIndex}) measure the preservation of clustering, i.e., how well a clustering of the integrated data overlaps with the cell type labels.

\textbf{NMI} is
computed as follows.  
Let $\mathcal{C} = \{1, ..., n_c\}$ denote a set of $n_c$ clusters from $k$-means in the embedding under evaluation, and let $C \in \mathcal{C}^n$ denote the vector of cluster assignments for each sample.
Let $N$ be the confusion matrix with rows corresponding to cell type labels, and columns corresponding to clusters in the embedding under evaluation, with entries $n_{lc}$ equal to the number of samples with cell type label $l$ that appear in cluster $c$ of the embedding.
Now let 
$n_{l \cdot } = \sum_{i} n_{lj}$ denote the total number of samples in cell type $l$,
$n_{\cdot c} = \sum_{i} n_{ic}$ denote the total number of samples in cluster $c$, and $n =  \sum_{ij} n_{ij} $ denote the total number of cells.

Then, the mutual information of labels and clusters in the confusion matrix $N$ is given by
\[\text{MI}(L, C) 
= \sum_{l \in \mathcal{L}} \sum_{c \in \mathcal{C}} 
n_{lc} \log \frac{n \cdot n_{lc}}{n_{\cdot c} n_{l\cdot}}.\]

Define the entropy of the cell type labels as 
$H(L) = - \sum_{l \in \mathcal{L}} n_{l\cdot} \log (n_{l\cdot} /n )$
, and the entropy of the clusters as
$H(C) = - \sum_{c \in \mathcal{C}} n_{\cdot c} \log (n_{\cdot c}/n)$.
Then, 
\[
\text{NMI}(L, C)
=
\frac{
2 \cdot \text{MI} (N)
}
{H(L) + H(C)}.
\]

\textbf{ARI} is another metric to evaluate the alignment of two $k$-means clusterings, based on comparing the amount of agreement to that of a random clustering. 
Let
\[
n_{\text{agreements}} =
\sum_{i,j} \mathbb{I}( L_i = L_j \text{ and } C_i = C_j )
+\sum_{i,j} \mathbb{I}( L_i \neq L_j \text{ and } C_i \neq C_j ).
\]

Define the number of of
true positives $n_{\text{tp}}$,  false positives $n_{\text{fp}}$, true negatives
$n_{\text{tn}}$, and false negatives $n_{\text{fn}}$ as
$$n_{\text{tp}} =
\sum_{i,j} \mathbb{I}( L_i = L_j \text{ and } C_i = C_j ),
\hspace{2cm}
n_{\text{fp}} =
\sum_{i,j} \mathbb{I}( L_i \neq L_j \text{ and } C_i = C_j ),$$

$$n_{\text{tn}} =
\sum_{i,j} \mathbb{I}( L_i \neq L_j \text{ and } C_i \neq C_j ),
\hspace{2cm}
n_{\text{tn}} =
\sum_{i,j} \mathbb{I}( L_i = L_j \text{ and } C_i \neq C_j ).$$

The ARI is then given by \citet{scikit-learn}:
\[
ARI(L,C) 
= \frac{
2 (n_{\text{tp}} n_{\text{tn}} - n_{\text{fn}} n_{\text{fp}}) 
} 
{ (n_{\text{tp}} + n_{\text{fn}}) \cdot (n_{\text{fn}}+n_{\text{tn}}) 
+ (n_{\text{tp}} + n_{\text{fp}})  \cdot (n_{\text{fp}} + n_{\text{tn}})}.
\] 



\textbf{Silhouette label} evaluates cell type separation in the integrated embedding. It is computed with the Average Silhouette Width (ASW~\cite{Rousseeuw1987}) of cell types, i.e., the average over all samples $i$ of 
\[ ASW (L) 
= \frac1n \sum_{i \in \mathcal{D}} s(i) 
= \frac1n \sum_{i \in \mathcal{D}} \frac{b(i) - a(i)}{\max\{a(i), b(i)\}},\]
where\[
a(i) = \frac{1}{|\{j  \in \mathcal{D} | L_j = L_i\}|} \sum_{j  \in \mathcal{D} |L_j = L_i} d(i, j)\]
denotes the average distance to cells in the same cluster, and \[
b(i) = \min_{L_k \neq L_i} \frac{1}{|\{j  \in \mathcal{D} | L_j = L_k\}|} \sum_{j \in \mathcal{D} | L_j = L_k} d(i, j)
\]
denotes the minimum average distance to another cluster.
For silhouette label, $L$ is the vector of cell type labels.

\textbf{cLISI} is based on the
local Inverse Simpson's Index (LISI), a measure of diversity that captures how many samples must be drawn from a neighborhood of a sample before drawing a duplicate label. 
For each cell, a probability distribution $p_i$ over its neighborhood is defined based on a Gaussian kernel of the distances to the cell of interest. Then
\[\text{LISI} = \text{median}_{i\in \mathcal{D}} \frac{1}{\sum_{l \in \mathcal L } p_i(l) }, \]
where $p_i(l) = \sum_{j} p_i(j) \mathbb{I}(L_j = l) $.
The result is transformed so that high scores correspond to low cell type diversity (high purity).
cLISI~\cite{korsunskyFastSensitiveAccurate2019} is the LISI where $L$ is the cell types, measuring the level of cell type mixing that occurs in the neighborhoods of all cells.

The \textbf{isolated labels} metric evaluates the separation of isolated labels, defined as labels that are present in the fewest batches, from other labels. It is computed as the ASW of the isolated labels versus the other labels.

\subsubsection{Batch correction metrics}

Batch removal adapted silhouette (\textbf{BRAS}~\cite{Rautenstrauch2025})
is an adaptation of ASW where $L$ is the batch identity labels that is better suited to evaluate batch mixing and nested batch effects. 
BRAS is defined like ASW, except that $b(i)$ is defined as the mean distance to cells in other clusters:
$b(i) = \frac{1}{|\{j  \in \mathcal{D} | L_j \neq L_k\}|} \sum_{j \in \mathcal{D} | L_j \neq L_k} d(i, j)
$.

Then, \[
\text{BRAS} = \frac1{n_l} \sum_{l \in \mathcal{L}} 
\left[
\frac{1}{|\{i \in \mathcal{D}|L_j = l\}|} \sum_{\{i \in \mathcal{D} |L_j = l\}} (1- |s(i)|)
\right]
\]

\textbf{iLISI}~\cite{korsunskyFastSensitiveAccurate2019} is the median of LISI applied to the batch labels (see section \ref{app:single_bio_metrics}).

\textbf{kBET}~\cite{Bttner2019} performs statistical tests to determine whether local composition of neighborhoods in terms of batch labels corresponds to the global batch distribution. 
The output is the average rejection rate of the null hypothesis (complete batch mixing) computed over the neighborhoods of samples selected at random. 
Here, kBET is computed separately for each cell type, and then averaged over all cell types:
\[ \text{kBET}(\mathcal{D}) 
=  \frac1 {n_l} \sum_{l \in \mathcal{L}} \text{kBET}(l) 
= 
\frac1 {n_l} \sum_{l \in \mathcal{L}} 
\left(
\frac1{|\mathcal{Y}|} \sum_{i \in \mathcal{Y}}
S_i^l
\right),
\]
where $S_i^l$ is the rejection rate of the statistical test on the batch labels of a neighborhood centered at $i$ restricted to cells with label $l$, and $\mathcal{Y} \subset \mathcal{D}$ is a random subset of samples.

\textbf{Graph connectivity} measures, for each label $l$, the connectivity of a $k$-NN graph $G_l$ restricted to samples with label $l$, i.e.,
\[
GC = 
\frac1 {|\mathcal{L}|} \sum_{l \in \mathcal{L}} \frac{ |LCC(G_l)|}{n_l},
\]
where $LCC$ denotes the largest connected component of a graph
and $n_l$ denotes the number of samples with label $l$.

Principal Component Regression (\textbf{PCR}~\cite{Bttner2019}) measures the variance explained by the batch. It is based on Principal component analysis (PCA), which returns $n$ principal components (PC).
By fitting a linear regression of the batches on the $i$th PC, we obtain an $R^2$ value, denoted $R^2_i$. 
The total variance explained by the batch is the weighted average over all PCs of the variance explained by the PC with the $R^2$ of the batch on this PC:
\[
\text{Var} = \sum_{i} \text{Var}_i \cdot R^2_i,
\]
where $\text{Var}_i$ denotes the variance of the data matrix $X$ explained by the $i$th principal component.

\subsection{Complete quantitative results}\label{app:singlecell_full_results_table}

\input{tables/singlecell}

In Table~\ref{tab:singlecell_full}, we provide the full results for all methods and metrics in the batch effect correction task described in Section~\ref{subsec:singlecell}.

\subsection{Robustness to partial label availability}\label{app:singlecell_mask_ablation}

\begin{figure} 
    \centering
     \includegraphics[width = \textwidth]{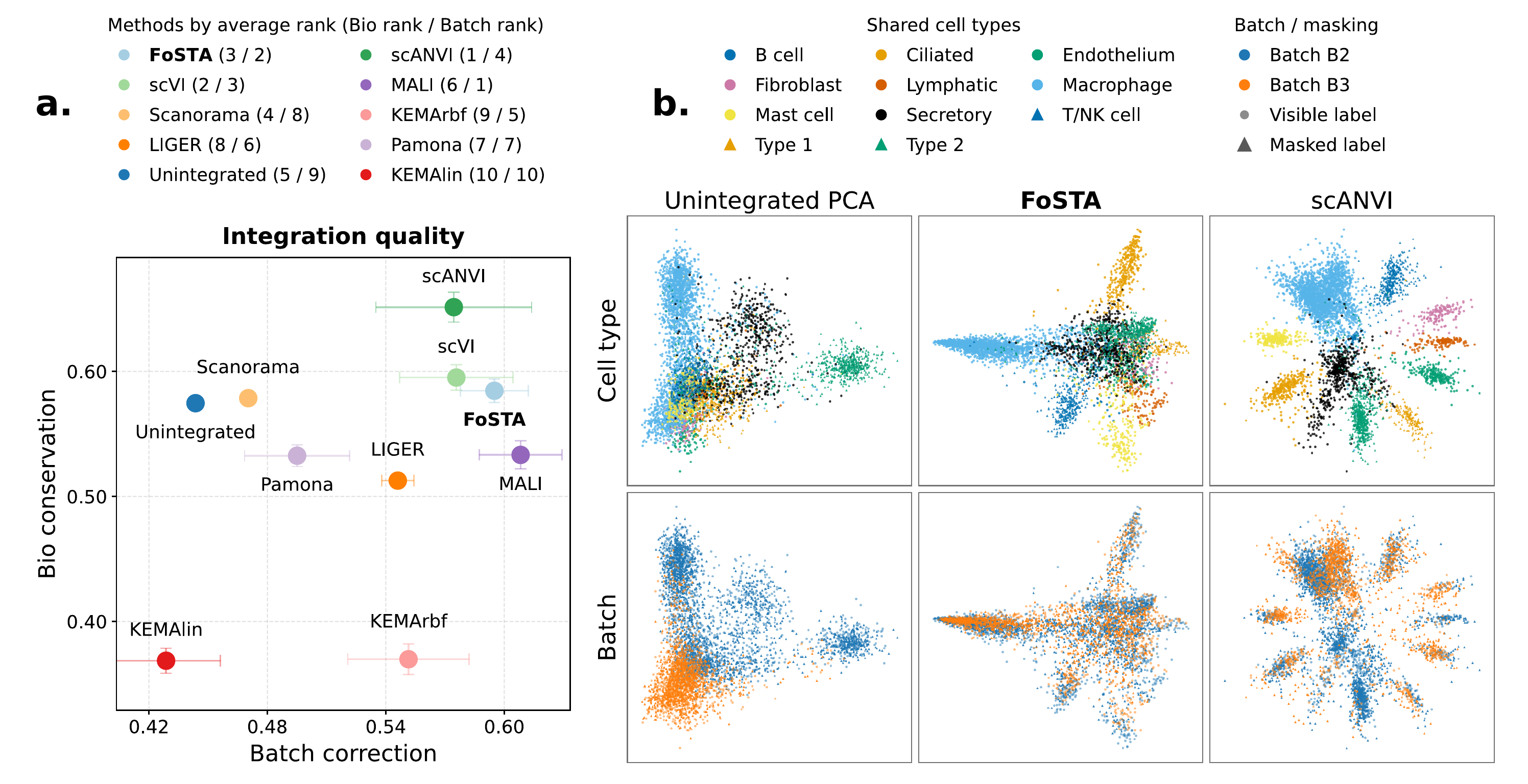}
     \caption{\textbf{Results of batch effect removal on lung single-cell data~\cite{vieirabragaCellularCensusHuman2019} under reduced label availability, with 50\% of labels masked instead of 20\% as in Fig.~\ref{fig:singlecell_main}.}
     \textbf{a.} Biological preservation (Bio) vs.\ batch correction (Batch) scores~\cite{lueckenBenchmarkingAtlaslevelData2022}, averaged across 36 batch pairs and five seeds. Error bars represent seed variability, computed as the average within-pair standard deviation across the 36 batch pairs.
    \textbf{b.} 
    Embeddings of the simulated batch integration task. Left column: data before correcting for batch effects, visualized with PCA.
    Center column: resulting embedding after applying FoSTA to correct the batch effects. Right column: resulting embedding after applying scANVI to correct the batch effects. The plots are colored by cell type (top) and batch (bottom). FoSTA mixes the batches well, while preserving the biological structure of the data. 
    }
    \label{fig:single_mask_ablation}
\end{figure}

In Fig.~\ref{fig:single_mask_ablation}, we repeat the batch alignment experiments from Section~\ref{subsec:singlecell} under reduced label availability, masking 50\% of the labels instead of 20\%.

\subsection{Extended visualizations}\label{app:singlecell_plot_extended}

\begin{figure}[ht] 
    \centering
    \includegraphics[width=\linewidth]{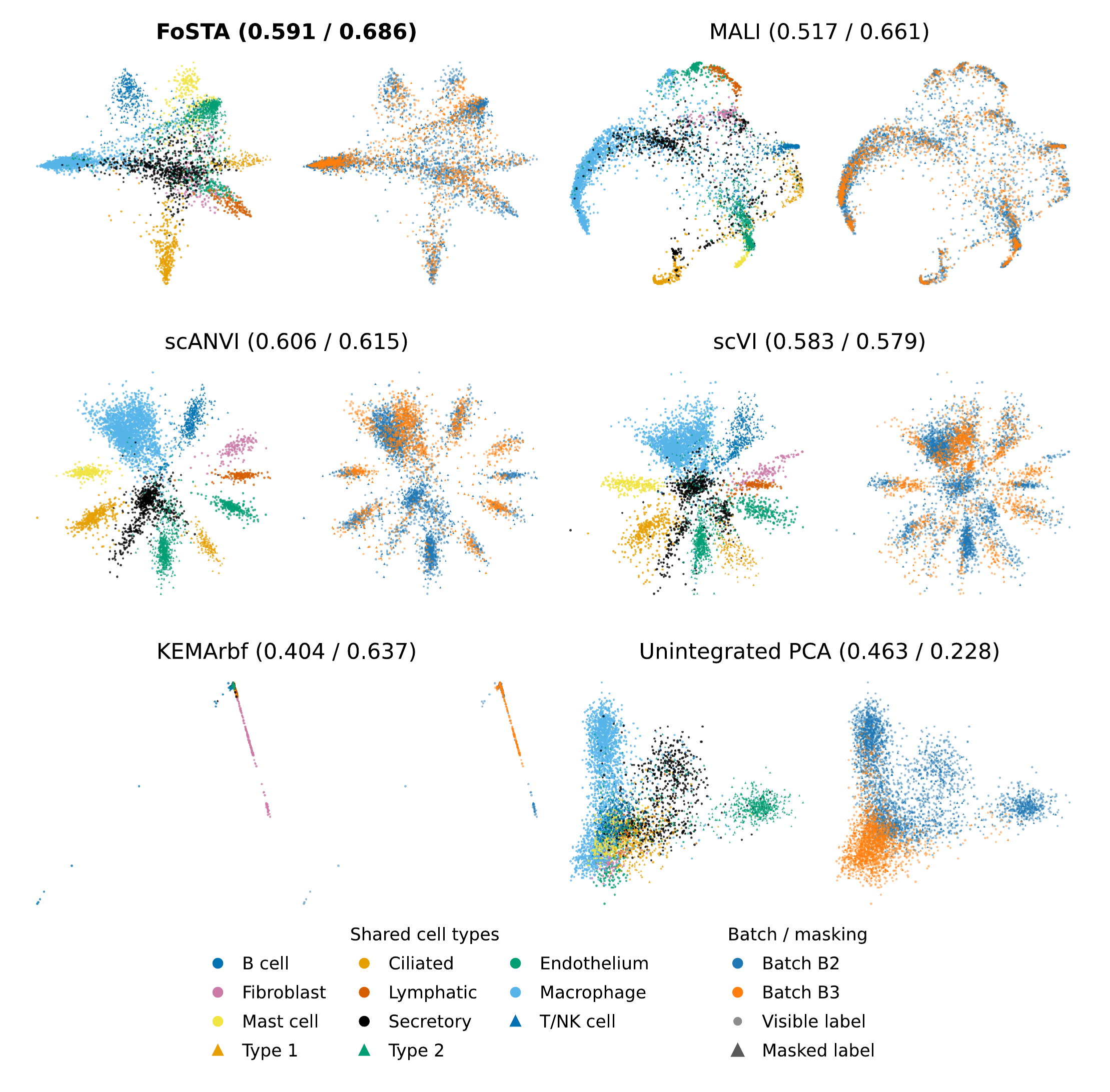}
    \caption{\textbf{Pairwise integration visualizations for the B2/B3 real batch pair from the lung dataset~\cite{lueckenBenchmarkingAtlaslevelData2022} using seed 56089.} For each method, the left panel is labeled by cell type and the right panel by batch; values in parentheses report biological conservation (Bio) and batch correction (Batch) scores, respectively. The selected example is representative of the aggregated quantitative results across the B-family batches and seeds. FoSTA produces a structured embedding with meaningful local groupings and good domain mixing while clearly distinguishing cell types. MALI also achieves good batch mixing and preserves global structure, but cell types are less clearly separated. scANVI and scVI produce more isolated cluster configurations, while KEMArbf strongly compresses the embedding and loses biological structure. The unintegrated PCA embedding illustrates the original batch effects present in the data.
    }
    \label{fig:singlecell_plot_extended}
\end{figure}

To complement the qualitative assessment in Section~\ref{subsec:singlecell}, Fig.~\ref{fig:singlecell_plot_extended} presents an extended comparison of two-dimensional embeddings for the B2/B3 batch pair (seed 56089). In addition to FoSTA, scANVI, and the unintegrated PCA representation shown in Fig.~\ref{fig:singlecell_main}, we include scVI, MALI, and KEMArbf. MALI and KEMArbf are included as the next highest-ranking methods after scANVI and scVI in the aggregated quantitative comparison, while the unintegrated representation is retained as a control.

Consistent with the quantitative results, FoSTA achieves good batch mixing while preserving a structured embedding with clearly distinguishable cell types. MALI also achieves strong batch correction and retains meaningful global structure, although cell types are less clearly separated. scANVI and scVI preserve biological groupings but produce more isolated cluster configurations, while KEMArbf strongly mixes the batches at the expense of biological structure. Overall, the visualization illustrates the trade-off between batch correction and biological conservation captured by the quantitative metrics.

\clearpage

\section*{NeurIPS Paper Checklist}


\begin{enumerate}

\item {\bf Claims}
    \item[] Question: Do the main claims made in the abstract and introduction accurately reflect the paper's contributions and scope?
    \item[] Answer: \answerYes{} 
    \item[] Justification: Our abstract and introduction accurately reflect the main contributions and scope of the paper. These claims are supported in the main text and appendix through methodological descriptions and empirical results. 

\item {\bf Limitations}
    \item[] Question: Does the paper discuss the limitations of the work performed by the authors?
    \item[] Answer: \answerYes{} 
    \item[] Justification: We explicitly discuss the limitations of our method in the discussion section (Section~\ref{sec:discussion}).

\item {\bf Theory assumptions and proofs}
    \item[] Question: For each theoretical result, does the paper provide the full set of assumptions and a complete (and correct) proof?
    \item[] Answer: \answerYes{} 
    \item[] Justification: The assumptions and proofs of Lemmas~\ref{lem:rf_gap_surrogate} and~\ref{lemma:l2_norm} are provided in Appendices~\ref{app:semisup_rfgap} and~\ref{app:l2_normalization_lemma}, respectively.

    \item {\bf Experimental result reproducibility}
    \item[] Question: Does the paper fully disclose all the information needed to reproduce the main experimental results of the paper to the extent that it affects the main claims and/or conclusions of the paper (regardless of whether the code and data are provided or not)?
    \item[] Answer: \answerYes{} 
    \item[] Justification: The paper provides sufficient details to reproduce the main experimental results, including a description of the model architecture (Section~\ref{sec:methods} and Fig.~\ref{fig:fosta}), implementation and hyperparameter settings (Appendix~\ref{app:implementation}), and dataset, preprocessing, experimental protocol, and evaluation details for the real-world multimodal, synthetic, and single-cell experiments (Appendices~\ref{app:multimodal_extended},~\ref{app:uci_full_section}, and~\ref{app:singlecell}).

\item {\bf Open access to data and code}
    \item[] Question: Does the paper provide open access to the data and code, with sufficient instructions to faithfully reproduce the main experimental results, as described in supplemental material?
    \item[] Answer: \answerYes{} 
    \item[] Justification: Code to reproduce all experiments is publicly available at \url{https://github.com/aumona/FoSTA-experiments} and is linked directly in Section~\ref{sec:results} of the main paper.

\item {\bf Experimental setting/details}
    \item[] Question: Does the paper specify all the training and test details (e.g., data splits, hyperparameters, how they were chosen, type of optimizer) necessary to understand the results?
    \item[] Answer: \answerYes{} 
    \item[] Justification: The paper specifies the relevant experimental details, including datasets, preprocessing, experimental protocols, implementation and hyperparameter settings, and evaluation metrics. These details are provided in the main text (Section~\ref{sec:results}) and expanded for the real-world multimodal, synthetic, and single-cell experiments in Appendices~\ref{app:implementation},~\ref{app:multimodal_extended},~\ref{app:uci_full_section}, and~\ref{app:singlecell}.

\item {\bf Experiment statistical significance}
    \item[] Question: Does the paper report error bars suitably and correctly defined or other appropriate information about the statistical significance of the experiments?
    \item[] Answer: \answerYes{} 
    \item[] Justification: The paper reports standard deviations for the main experimental results across multiple random seeds, including the real-world multimodal experiments in Section~\ref{subsec:real_multimodal} and the single-cell batch alignment experiments in Section~\ref{subsec:singlecell}, as well as the extended experiments reported in Appendices~\ref{app:multimodal_extended},~\ref{app:uci_full_section}, and~\ref{app:singlecell}.

\item {\bf Experiments compute resources}
    \item[] Question: For each experiment, does the paper provide sufficient information on the computer resources (type of compute workers, memory, time of execution) needed to reproduce the experiments?
    \item[] Answer: \answerYes{} 
    \item[] Justification: The paper provides sufficient information on the compute resources required to reproduce the experiments. These details are documented in Appendix~\ref{app:implementation}.
    
\item {\bf Code of ethics}
    \item[] Question: Does the research conducted in the paper conform, in every respect, with the NeurIPS Code of Ethics \url{https://neurips.cc/public/EthicsGuidelines}?
    \item[] Answer: \answerYes{} 
    \item[] Justification: This research presented in the paper fully complies with the NeurIPS Code of Ethics. 

\item {\bf Broader impacts}
    \item[] Question: Does the paper discuss both potential positive societal impacts and negative societal impacts of the work performed?
    \item[] Answer: \answerYes{} 
    \item[] Justification: The paper discusses both the potential positive and negative societal impacts of the proposed work in the Broader Impacts section.
    
\item {\bf Safeguards}
    \item[] Question: Does the paper describe safeguards that have been put in place for responsible release of data or models that have a high risk for misuse (e.g., pre-trained language models, image generators, or scraped datasets)?
    \item[] Answer: \answerNA{} 
    \item[] Justification: The paper does not involve models or datasets that pose a high risk of misuse or dual use. The proposed method is intended for scientific research and does not generate or process sensitive or potentially harmful content.

\item {\bf Licenses for existing assets}
    \item[] Question: Are the creators or original owners of assets (e.g., code, data, models), used in the paper, properly credited and are the license and terms of use explicitly mentioned and properly respected?
    \item[] Answer: \answerYes{} 
    \item[] Justification: All external datasets and models used in the paper are properly cited with references to their original sources in Section~\ref{sec:results} and Appendices~\ref{app:real_multimodal_datasets},~\ref{app:uci_datasets} and~\ref{app:singlecell_protocol}.

\item {\bf New assets}
\item[] Question: Are new assets introduced in the paper well documented and is the documentation provided alongside the assets?
\item[] Answer: \answerYes{}
\item[] Justification: The FoSTA source code and experimental pipelines are publicly available, together with documentation, installation instructions, and scripts to reproduce the experiments. The repository is linked in Section~\ref{sec:results}.

\item {\bf Crowdsourcing and research with human subjects}
    \item[] Question: For crowdsourcing experiments and research with human subjects, does the paper include the full text of instructions given to participants and screenshots, if applicable, as well as details about compensation (if any)? 
    \item[] Answer: \answerNA{} 
    \item[] Justification: The paper uses publicly a available single-cell dataset that was collected and shared by third parties. We do not conduct any new research involving human subjects or crowdsourcing, and the data used is ethically sourced.

\item {\bf Institutional review board (IRB) approvals or equivalent for research with human subjects}
    \item[] Question: Does the paper describe potential risks incurred by study participants, whether such risks were disclosed to the subjects, and whether Institutional Review Board (IRB) approvals (or an equivalent approval/review based on the requirements of your country or institution) were obtained?
    \item[] Answer: \answerNA{} 
    \item[] Justification: The paper does not involve any direct research with human subjects or participant interaction. All data used are publicly available and were collected under proper ethical oversight by the original data providers.

\item {\bf Declaration of LLM usage}
    \item[] Question: Does the paper describe the usage of LLMs if it is an important, original, or non-standard component of the core methods in this research? Note that if the LLM is used only for writing, editing, or formatting purposes and does \emph{not} impact the core methodology, scientific rigor, or originality of the research, declaration is not required.
    \item[] Answer: \answerNA{} 
    \item[] Justification: No LLMs were used in the development of the core methods presented in this research. See the AI Use Disclosure section for details.

\end{enumerate}

\end{document}

%% file: tables/real_multimodal.tex
\begin{table}[t]
\caption{\textbf{Model performance on real-world multimodal alignment benchmarks.} Results are averaged over five seeds for label transfer accuracy (Acc), alignment score (AS), and FOSCTTM (FOS), with $\uparrow$/$\downarrow$ indicating better performance. Gold, silver, and bronze denote the top three methods. RGB-D and Sketchy include both ResNet18~\cite{heDeepResidualLearning2015} and DINOv2-Base~\cite{oquabDINOv2LearningRobust2024} features. Top-k accuracy (Acc@5 or Acc@10) is also reported where applicable. RGB-D scores are not reported for Pamona due to excessive runtimes ($>10$ hours).}
\label{tab:real_multimodal}
\centering
\small
\setlength{\tabcolsep}{2pt}
\setlength{\fboxsep}{1pt}
\renewcommand{\arraystretch}{1.0}

\begin{tabular}{lr c c c c c}
\toprule
Alignment task & Metric & \makebox[1.5cm][c]{\textbf{FoSTA}} & \makebox[1.5cm][c]{KEMAlin} & \makebox[1.5cm][c]{KEMArbf} & \makebox[1.5cm][c]{MALI} & \makebox[1.5cm][c]{Pamona} \\
\midrule
\multirow[c]{4}{*}[-\dimexpr 3\fboxsep\relax]{\begin{tabular}[c]{@{}l@{}}Audio $\leftrightarrow$ Video \\ (AVE, 28 classes)\end{tabular}} & Acc$\uparrow$ & \makebox[1.5cm][c]{\colorbox[rgb]{1,0.95,0.8}{\makebox[\dimexpr 1.5cm-2\fboxsep\relax][c]{\raisebox{0pt}[\ht\strutbox][\dp\strutbox]{\textcolor{black}{\textbf{0.404{\tiny $\pm$0.01}}}}}}} & \makebox[1.5cm][c]{0.226{\tiny $\pm$0.01}} & \makebox[1.5cm][c]{\bronze{0.362{\tiny $\pm$0.02}}} & \makebox[1.5cm][c]{\colorbox[gray]{0.85}{\makebox[\dimexpr 1.5cm-2\fboxsep\relax][c]{\raisebox{0pt}[\ht\strutbox][\dp\strutbox]{\textcolor{black}{\textbf{0.374{\tiny $\pm$0.01}}}}}}} & \makebox[1.5cm][c]{0.202{\tiny $\pm$0.01}} \\
 & Acc@5$\uparrow$ & \makebox[1.5cm][c]{\colorbox[rgb]{1,0.95,0.8}{\makebox[\dimexpr 1.5cm-2\fboxsep\relax][c]{\raisebox{0pt}[\ht\strutbox][\dp\strutbox]{\textcolor{black}{\textbf{0.786{\tiny $\pm$0.01}}}}}}} & \makebox[1.5cm][c]{0.687{\tiny $\pm$0.01}} & \makebox[1.5cm][c]{\bronze{0.747{\tiny $\pm$0.00}}} & \makebox[1.5cm][c]{\colorbox[gray]{0.85}{\makebox[\dimexpr 1.5cm-2\fboxsep\relax][c]{\raisebox{0pt}[\ht\strutbox][\dp\strutbox]{\textcolor{black}{\textbf{0.765{\tiny $\pm$0.01}}}}}}} & \makebox[1.5cm][c]{0.442{\tiny $\pm$0.02}} \\
 & AS$\uparrow$ & \makebox[1.5cm][c]{0.914{\tiny $\pm$0.00}} & \makebox[1.5cm][c]{\bronze{0.921{\tiny $\pm$0.00}}} & \makebox[1.5cm][c]{\colorbox[gray]{0.85}{\makebox[\dimexpr 1.5cm-2\fboxsep\relax][c]{\raisebox{0pt}[\ht\strutbox][\dp\strutbox]{\textcolor{black}{\textbf{0.930{\tiny $\pm$0.00}}}}}}} & \makebox[1.5cm][c]{\colorbox[rgb]{1,0.95,0.8}{\makebox[\dimexpr 1.5cm-2\fboxsep\relax][c]{\raisebox{0pt}[\ht\strutbox][\dp\strutbox]{\textcolor{black}{\textbf{1.002{\tiny $\pm$0.00}}}}}}} & \makebox[1.5cm][c]{0.333{\tiny $\pm$0.01}} \\
 & FOS$\downarrow$ & \makebox[1.5cm][c]{\colorbox[gray]{0.85}{\makebox[\dimexpr 1.5cm-2\fboxsep\relax][c]{\raisebox{0pt}[\ht\strutbox][\dp\strutbox]{\textcolor{black}{\textbf{0.093{\tiny $\pm$0.00}}}}}}} & \makebox[1.5cm][c]{\bronze{0.167{\tiny $\pm$0.00}}} & \makebox[1.5cm][c]{\colorbox[rgb]{1,0.95,0.8}{\makebox[\dimexpr 1.5cm-2\fboxsep\relax][c]{\raisebox{0pt}[\ht\strutbox][\dp\strutbox]{\textcolor{black}{\textbf{0.075{\tiny $\pm$0.00}}}}}}} & \makebox[1.5cm][c]{0.170{\tiny $\pm$0.00}} & \makebox[1.5cm][c]{0.196{\tiny $\pm$0.02}} \\
\midrule
\multirow[c]{4}{*}[-\dimexpr 3\fboxsep\relax]{\begin{tabular}[c]{@{}l@{}}Image $\leftrightarrow$ Depth crop \\ (RGB-D DINOv2-B, 51 classes)\end{tabular}} & Acc$\uparrow$ & \makebox[1.5cm][c]{\colorbox[rgb]{1,0.95,0.8}{\makebox[\dimexpr 1.5cm-2\fboxsep\relax][c]{\raisebox{0pt}[\ht\strutbox][\dp\strutbox]{\textcolor{black}{\textbf{0.324{\tiny $\pm$0.00}}}}}}} & \makebox[1.5cm][c]{\colorbox[gray]{0.85}{\makebox[\dimexpr 1.5cm-2\fboxsep\relax][c]{\raisebox{0pt}[\ht\strutbox][\dp\strutbox]{\textcolor{black}{\textbf{0.287{\tiny $\pm$0.00}}}}}}} & \makebox[1.5cm][c]{\bronze{0.191{\tiny $\pm$0.02}}} & \makebox[1.5cm][c]{0.097{\tiny $\pm$0.02}} & --- \\
 & Acc@10$\uparrow$ & \makebox[1.5cm][c]{\colorbox[gray]{0.85}{\makebox[\dimexpr 1.5cm-2\fboxsep\relax][c]{\raisebox{0pt}[\ht\strutbox][\dp\strutbox]{\textcolor{black}{\textbf{0.792{\tiny $\pm$0.00}}}}}}} & \makebox[1.5cm][c]{\colorbox[rgb]{1,0.95,0.8}{\makebox[\dimexpr 1.5cm-2\fboxsep\relax][c]{\raisebox{0pt}[\ht\strutbox][\dp\strutbox]{\textcolor{black}{\textbf{0.813{\tiny $\pm$0.00}}}}}}} & \makebox[1.5cm][c]{\bronze{0.578{\tiny $\pm$0.05}}} & \makebox[1.5cm][c]{0.469{\tiny $\pm$0.03}} & --- \\
 & AS$\uparrow$ & \makebox[1.5cm][c]{\bronze{0.934{\tiny $\pm$0.01}}} & \makebox[1.5cm][c]{\colorbox[gray]{0.85}{\makebox[\dimexpr 1.5cm-2\fboxsep\relax][c]{\raisebox{0pt}[\ht\strutbox][\dp\strutbox]{\textcolor{black}{\textbf{0.944{\tiny $\pm$0.00}}}}}}} & \makebox[1.5cm][c]{0.884{\tiny $\pm$0.03}} & \makebox[1.5cm][c]{\colorbox[rgb]{1,0.95,0.8}{\makebox[\dimexpr 1.5cm-2\fboxsep\relax][c]{\raisebox{0pt}[\ht\strutbox][\dp\strutbox]{\textcolor{black}{\textbf{0.984{\tiny $\pm$0.00}}}}}}} & --- \\
 & FOS$\downarrow$ & \makebox[1.5cm][c]{\colorbox[rgb]{1,0.95,0.8}{\makebox[\dimexpr 1.5cm-2\fboxsep\relax][c]{\raisebox{0pt}[\ht\strutbox][\dp\strutbox]{\textcolor{black}{\textbf{0.079{\tiny $\pm$0.00}}}}}}} & \makebox[1.5cm][c]{\colorbox[gray]{0.85}{\makebox[\dimexpr 1.5cm-2\fboxsep\relax][c]{\raisebox{0pt}[\ht\strutbox][\dp\strutbox]{\textcolor{black}{\textbf{0.082{\tiny $\pm$0.00}}}}}}} & \makebox[1.5cm][c]{\bronze{0.133{\tiny $\pm$0.02}}} & \makebox[1.5cm][c]{0.294{\tiny $\pm$0.02}} & --- \\
\midrule
\multirow[c]{4}{*}[-\dimexpr 3\fboxsep\relax]{\begin{tabular}[c]{@{}l@{}}Image $\leftrightarrow$ Depth crop \\ (RGB-D ResNet18, 51 classes)\end{tabular}} & Acc$\uparrow$ & \makebox[1.5cm][c]{\colorbox[rgb]{1,0.95,0.8}{\makebox[\dimexpr 1.5cm-2\fboxsep\relax][c]{\raisebox{0pt}[\ht\strutbox][\dp\strutbox]{\textcolor{black}{\textbf{0.284{\tiny $\pm$0.01}}}}}}} & \makebox[1.5cm][c]{\colorbox[gray]{0.85}{\makebox[\dimexpr 1.5cm-2\fboxsep\relax][c]{\raisebox{0pt}[\ht\strutbox][\dp\strutbox]{\textcolor{black}{\textbf{0.213{\tiny $\pm$0.00}}}}}}} & \makebox[1.5cm][c]{\bronze{0.208{\tiny $\pm$0.03}}} & \makebox[1.5cm][c]{0.085{\tiny $\pm$0.01}} & --- \\
 & Acc@10$\uparrow$ & \makebox[1.5cm][c]{\colorbox[rgb]{1,0.95,0.8}{\makebox[\dimexpr 1.5cm-2\fboxsep\relax][c]{\raisebox{0pt}[\ht\strutbox][\dp\strutbox]{\textcolor{black}{\textbf{0.813{\tiny $\pm$0.00}}}}}}} & \makebox[1.5cm][c]{\colorbox[gray]{0.85}{\makebox[\dimexpr 1.5cm-2\fboxsep\relax][c]{\raisebox{0pt}[\ht\strutbox][\dp\strutbox]{\textcolor{black}{\textbf{0.768{\tiny $\pm$0.01}}}}}}} & \makebox[1.5cm][c]{\bronze{0.585{\tiny $\pm$0.04}}} & \makebox[1.5cm][c]{0.431{\tiny $\pm$0.02}} & --- \\
 & AS$\uparrow$ & \makebox[1.5cm][c]{\bronze{0.936{\tiny $\pm$0.00}}} & \makebox[1.5cm][c]{\colorbox[gray]{0.85}{\makebox[\dimexpr 1.5cm-2\fboxsep\relax][c]{\raisebox{0pt}[\ht\strutbox][\dp\strutbox]{\textcolor{black}{\textbf{0.972{\tiny $\pm$0.00}}}}}}} & \makebox[1.5cm][c]{0.898{\tiny $\pm$0.02}} & \makebox[1.5cm][c]{\colorbox[rgb]{1,0.95,0.8}{\makebox[\dimexpr 1.5cm-2\fboxsep\relax][c]{\raisebox{0pt}[\ht\strutbox][\dp\strutbox]{\textcolor{black}{\textbf{0.986{\tiny $\pm$0.00}}}}}}} & --- \\
 & FOS$\downarrow$ & \makebox[1.5cm][c]{\colorbox[rgb]{1,0.95,0.8}{\makebox[\dimexpr 1.5cm-2\fboxsep\relax][c]{\raisebox{0pt}[\ht\strutbox][\dp\strutbox]{\textcolor{black}{\textbf{0.080{\tiny $\pm$0.00}}}}}}} & \makebox[1.5cm][c]{\colorbox[gray]{0.85}{\makebox[\dimexpr 1.5cm-2\fboxsep\relax][c]{\raisebox{0pt}[\ht\strutbox][\dp\strutbox]{\textcolor{black}{\textbf{0.112{\tiny $\pm$0.00}}}}}}} & \makebox[1.5cm][c]{\bronze{0.124{\tiny $\pm$0.01}}} & \makebox[1.5cm][c]{0.342{\tiny $\pm$0.01}} & --- \\
\midrule
\multirow[c]{4}{*}[-\dimexpr 3\fboxsep\relax]{\begin{tabular}[c]{@{}l@{}}Image $\leftrightarrow$ Human sketch \\ (Sketchy DINOv2-B, 125 classes)\end{tabular}} & Acc$\uparrow$ & \makebox[1.5cm][c]{\colorbox[gray]{0.85}{\makebox[\dimexpr 1.5cm-2\fboxsep\relax][c]{\raisebox{0pt}[\ht\strutbox][\dp\strutbox]{\textcolor{black}{\textbf{0.137{\tiny $\pm$0.01}}}}}}} & \makebox[1.5cm][c]{0.092{\tiny $\pm$0.01}} & \makebox[1.5cm][c]{\colorbox[rgb]{1,0.95,0.8}{\makebox[\dimexpr 1.5cm-2\fboxsep\relax][c]{\raisebox{0pt}[\ht\strutbox][\dp\strutbox]{\textcolor{black}{\textbf{0.209{\tiny $\pm$0.02}}}}}}} & \makebox[1.5cm][c]{\bronze{0.106{\tiny $\pm$0.00}}} & \makebox[1.5cm][c]{0.011{\tiny $\pm$0.00}} \\
 & Acc@10$\uparrow$ & \makebox[1.5cm][c]{\colorbox[rgb]{1,0.95,0.8}{\makebox[\dimexpr 1.5cm-2\fboxsep\relax][c]{\raisebox{0pt}[\ht\strutbox][\dp\strutbox]{\textcolor{black}{\textbf{0.613{\tiny $\pm$0.01}}}}}}} & \makebox[1.5cm][c]{\bronze{0.521{\tiny $\pm$0.03}}} & \makebox[1.5cm][c]{\colorbox[gray]{0.85}{\makebox[\dimexpr 1.5cm-2\fboxsep\relax][c]{\raisebox{0pt}[\ht\strutbox][\dp\strutbox]{\textcolor{black}{\textbf{0.596{\tiny $\pm$0.01}}}}}}} & \makebox[1.5cm][c]{0.410{\tiny $\pm$0.01}} & \makebox[1.5cm][c]{0.117{\tiny $\pm$0.01}} \\
 & AS$\uparrow$ & \makebox[1.5cm][c]{\colorbox[gray]{0.85}{\makebox[\dimexpr 1.5cm-2\fboxsep\relax][c]{\raisebox{0pt}[\ht\strutbox][\dp\strutbox]{\textcolor{black}{\textbf{0.950{\tiny $\pm$0.00}}}}}}} & \makebox[1.5cm][c]{\bronze{0.947{\tiny $\pm$0.01}}} & \makebox[1.5cm][c]{0.934{\tiny $\pm$0.01}} & \makebox[1.5cm][c]{\colorbox[rgb]{1,0.95,0.8}{\makebox[\dimexpr 1.5cm-2\fboxsep\relax][c]{\raisebox{0pt}[\ht\strutbox][\dp\strutbox]{\textcolor{black}{\textbf{0.980{\tiny $\pm$0.00}}}}}}} & \makebox[1.5cm][c]{0.392{\tiny $\pm$0.02}} \\
 & FOS$\downarrow$ & \makebox[1.5cm][c]{\colorbox[gray]{0.85}{\makebox[\dimexpr 1.5cm-2\fboxsep\relax][c]{\raisebox{0pt}[\ht\strutbox][\dp\strutbox]{\textcolor{black}{\textbf{0.114{\tiny $\pm$0.00}}}}}}} & \makebox[1.5cm][c]{\bronze{0.130{\tiny $\pm$0.01}}} & \makebox[1.5cm][c]{\colorbox[rgb]{1,0.95,0.8}{\makebox[\dimexpr 1.5cm-2\fboxsep\relax][c]{\raisebox{0pt}[\ht\strutbox][\dp\strutbox]{\textcolor{black}{\textbf{0.061{\tiny $\pm$0.00}}}}}}} & \makebox[1.5cm][c]{0.246{\tiny $\pm$0.01}} & \makebox[1.5cm][c]{0.392{\tiny $\pm$0.02}} \\
\midrule
\multirow[c]{4}{*}[-\dimexpr 3\fboxsep\relax]{\begin{tabular}[c]{@{}l@{}}Image $\leftrightarrow$ Human sketch \\ (Sketchy ResNet18, 125 classes)\end{tabular}} & Acc$\uparrow$ & \makebox[1.5cm][c]{\colorbox[gray]{0.85}{\makebox[\dimexpr 1.5cm-2\fboxsep\relax][c]{\raisebox{0pt}[\ht\strutbox][\dp\strutbox]{\textcolor{black}{\textbf{0.096{\tiny $\pm$0.00}}}}}}} & \makebox[1.5cm][c]{0.045{\tiny $\pm$0.00}} & \makebox[1.5cm][c]{\bronze{0.090{\tiny $\pm$0.00}}} & \makebox[1.5cm][c]{\colorbox[rgb]{1,0.95,0.8}{\makebox[\dimexpr 1.5cm-2\fboxsep\relax][c]{\raisebox{0pt}[\ht\strutbox][\dp\strutbox]{\textcolor{black}{\textbf{0.151{\tiny $\pm$0.01}}}}}}} & \makebox[1.5cm][c]{0.021{\tiny $\pm$0.00}} \\
 & Acc@10$\uparrow$ & \makebox[1.5cm][c]{\colorbox[gray]{0.85}{\makebox[\dimexpr 1.5cm-2\fboxsep\relax][c]{\raisebox{0pt}[\ht\strutbox][\dp\strutbox]{\textcolor{black}{\textbf{0.452{\tiny $\pm$0.00}}}}}}} & \makebox[1.5cm][c]{0.315{\tiny $\pm$0.01}} & \makebox[1.5cm][c]{\bronze{0.385{\tiny $\pm$0.01}}} & \makebox[1.5cm][c]{\colorbox[rgb]{1,0.95,0.8}{\makebox[\dimexpr 1.5cm-2\fboxsep\relax][c]{\raisebox{0pt}[\ht\strutbox][\dp\strutbox]{\textcolor{black}{\textbf{0.504{\tiny $\pm$0.01}}}}}}} & \makebox[1.5cm][c]{0.204{\tiny $\pm$0.01}} \\
 & AS$\uparrow$ & \makebox[1.5cm][c]{\colorbox[gray]{0.85}{\makebox[\dimexpr 1.5cm-2\fboxsep\relax][c]{\raisebox{0pt}[\ht\strutbox][\dp\strutbox]{\textcolor{black}{\textbf{0.959{\tiny $\pm$0.00}}}}}}} & \makebox[1.5cm][c]{\bronze{0.866{\tiny $\pm$0.00}}} & \makebox[1.5cm][c]{0.840{\tiny $\pm$0.02}} & \makebox[1.5cm][c]{\colorbox[rgb]{1,0.95,0.8}{\makebox[\dimexpr 1.5cm-2\fboxsep\relax][c]{\raisebox{0pt}[\ht\strutbox][\dp\strutbox]{\textcolor{black}{\textbf{0.987{\tiny $\pm$0.00}}}}}}} & \makebox[1.5cm][c]{0.735{\tiny $\pm$0.03}} \\
 & FOS$\downarrow$ & \makebox[1.5cm][c]{\colorbox[gray]{0.85}{\makebox[\dimexpr 1.5cm-2\fboxsep\relax][c]{\raisebox{0pt}[\ht\strutbox][\dp\strutbox]{\textcolor{black}{\textbf{0.146{\tiny $\pm$0.00}}}}}}} & \makebox[1.5cm][c]{0.209{\tiny $\pm$0.00}} & \makebox[1.5cm][c]{\colorbox[rgb]{1,0.95,0.8}{\makebox[\dimexpr 1.5cm-2\fboxsep\relax][c]{\raisebox{0pt}[\ht\strutbox][\dp\strutbox]{\textcolor{black}{\textbf{0.134{\tiny $\pm$0.00}}}}}}} & \makebox[1.5cm][c]{\bronze{0.173{\tiny $\pm$0.00}}} & \makebox[1.5cm][c]{0.342{\tiny $\pm$0.02}} \\
\midrule
\multirow[c]{3}{*}[-\dimexpr 2\fboxsep\relax]{\begin{tabular}[c]{@{}l@{}}Sensor 1 $\leftrightarrow$ Sensor 2 \\ (HAR, 6 classes)\end{tabular}} & Acc$\uparrow$ & \makebox[1.5cm][c]{\colorbox[rgb]{1,0.95,0.8}{\makebox[\dimexpr 1.5cm-2\fboxsep\relax][c]{\raisebox{0pt}[\ht\strutbox][\dp\strutbox]{\textcolor{black}{\textbf{0.627{\tiny $\pm$0.01}}}}}}} & \makebox[1.5cm][c]{0.215{\tiny $\pm$0.00}} & \makebox[1.5cm][c]{\bronze{0.506{\tiny $\pm$0.01}}} & \makebox[1.5cm][c]{\colorbox[gray]{0.85}{\makebox[\dimexpr 1.5cm-2\fboxsep\relax][c]{\raisebox{0pt}[\ht\strutbox][\dp\strutbox]{\textcolor{black}{\textbf{0.619{\tiny $\pm$0.01}}}}}}} & \makebox[1.5cm][c]{0.139{\tiny $\pm$0.02}} \\
 & AS$\uparrow$ & \makebox[1.5cm][c]{\bronze{0.681{\tiny $\pm$0.02}}} & \makebox[1.5cm][c]{0.081{\tiny $\pm$0.00}} & \makebox[1.5cm][c]{\colorbox[gray]{0.85}{\makebox[\dimexpr 1.5cm-2\fboxsep\relax][c]{\raisebox{0pt}[\ht\strutbox][\dp\strutbox]{\textcolor{black}{\textbf{0.847{\tiny $\pm$0.03}}}}}}} & \makebox[1.5cm][c]{\colorbox[rgb]{1,0.95,0.8}{\makebox[\dimexpr 1.5cm-2\fboxsep\relax][c]{\raisebox{0pt}[\ht\strutbox][\dp\strutbox]{\textcolor{black}{\textbf{0.891{\tiny $\pm$0.01}}}}}}} & \makebox[1.5cm][c]{0.108{\tiny $\pm$0.01}} \\
 & FOS$\downarrow$ & \makebox[1.5cm][c]{\colorbox[rgb]{1,0.95,0.8}{\makebox[\dimexpr 1.5cm-2\fboxsep\relax][c]{\raisebox{0pt}[\ht\strutbox][\dp\strutbox]{\textcolor{black}{\textbf{0.175{\tiny $\pm$0.00}}}}}}} & \makebox[1.5cm][c]{0.479{\tiny $\pm$0.00}} & \makebox[1.5cm][c]{\colorbox[gray]{0.85}{\makebox[\dimexpr 1.5cm-2\fboxsep\relax][c]{\raisebox{0pt}[\ht\strutbox][\dp\strutbox]{\textcolor{black}{\textbf{0.195{\tiny $\pm$0.01}}}}}}} & \makebox[1.5cm][c]{\bronze{0.243{\tiny $\pm$0.00}}} & \makebox[1.5cm][c]{0.435{\tiny $\pm$0.05}} \\

\bottomrule
\end{tabular}
\end{table}

%% file: tables/uci_general.tex
\begin{table}[t]
{\small
\caption{\textbf{Aggregated model performance over UCI datasets and seeds under selected distortion splits.} Each cell reports mean $\pm$ the average within-dataset standard deviation across five seeds. Results are shown for label transfer accuracy, alignment score (AS), and correspondence recovery measured by FOSCTTM. Higher is better for accuracy and AS, while lower is better for FOSCTTM. The top three results for each metric are highlighted in gold (1st), silver (2nd), and bronze (3rd). The final three rows report average scores across the displayed splits, with standard deviations also averaged across splits.}
\label{tab:uci}
}
\centering
\small
\setlength{\tabcolsep}{2pt}
\setlength{\fboxsep}{1pt}
\renewcommand{\arraystretch}{1.0}

\begin{tabular}{lr c c c c c}
\toprule
Split & Metric & \makebox[1.5cm][c]{\textbf{FoSTA}} & \makebox[1.5cm][c]{KEMAlin} & \makebox[1.5cm][c]{KEMArbf} & \makebox[1.5cm][c]{MALI} & \makebox[1.5cm][c]{Pamona} \\
\midrule
 & Accuracy$\uparrow$ & \makebox[1.5cm][c]{\colorbox[rgb]{1,0.95,0.8}{\makebox[\dimexpr 1.5cm-2\fboxsep\relax][c]{\raisebox{0pt}[\ht\strutbox][\dp\strutbox]{\textcolor{black}{\textbf{0.788{\tiny $\pm$0.02}}}}}}} & \makebox[1.5cm][c]{\bronze{0.690{\tiny $\pm$0.03}}} & \makebox[1.5cm][c]{0.671{\tiny $\pm$0.04}} & \makebox[1.5cm][c]{\colorbox[gray]{0.85}{\makebox[\dimexpr 1.5cm-2\fboxsep\relax][c]{\raisebox{0pt}[\ht\strutbox][\dp\strutbox]{\textcolor{black}{\textbf{0.691{\tiny $\pm$0.04}}}}}}} & \makebox[1.5cm][c]{0.615{\tiny $\pm$0.06}} \\
Noise & AS$\uparrow$ & \makebox[1.5cm][c]{\bronze{0.294{\tiny $\pm$0.02}}} & \makebox[1.5cm][c]{0.287{\tiny $\pm$0.04}} & \makebox[1.5cm][c]{0.290{\tiny $\pm$0.03}} & \makebox[1.5cm][c]{\colorbox[gray]{0.85}{\makebox[\dimexpr 1.5cm-2\fboxsep\relax][c]{\raisebox{0pt}[\ht\strutbox][\dp\strutbox]{\textcolor{black}{\textbf{0.883{\tiny $\pm$0.02}}}}}}} & \makebox[1.5cm][c]{\colorbox[rgb]{1,0.95,0.8}{\makebox[\dimexpr 1.5cm-2\fboxsep\relax][c]{\raisebox{0pt}[\ht\strutbox][\dp\strutbox]{\textcolor{black}{\textbf{0.936{\tiny $\pm$0.02}}}}}}} \\
 & FOSCTTM$\downarrow$ & \makebox[1.5cm][c]{\colorbox[rgb]{1,0.95,0.8}{\makebox[\dimexpr 1.5cm-2\fboxsep\relax][c]{\raisebox{0pt}[\ht\strutbox][\dp\strutbox]{\textcolor{black}{\textbf{0.251{\tiny $\pm$0.02}}}}}}} & \makebox[1.5cm][c]{0.339{\tiny $\pm$0.01}} & \makebox[1.5cm][c]{\bronze{0.335{\tiny $\pm$0.01}}} & \makebox[1.5cm][c]{\colorbox[gray]{0.85}{\makebox[\dimexpr 1.5cm-2\fboxsep\relax][c]{\raisebox{0pt}[\ht\strutbox][\dp\strutbox]{\textcolor{black}{\textbf{0.329{\tiny $\pm$0.02}}}}}}} & \makebox[1.5cm][c]{0.362{\tiny $\pm$0.03}} \\
\midrule
 & Accuracy$\uparrow$ & \makebox[1.5cm][c]{\colorbox[rgb]{1,0.95,0.8}{\makebox[\dimexpr 1.5cm-2\fboxsep\relax][c]{\raisebox{0pt}[\ht\strutbox][\dp\strutbox]{\textcolor{black}{\textbf{0.767{\tiny $\pm$0.03}}}}}}} & \makebox[1.5cm][c]{0.708{\tiny $\pm$0.05}} & \makebox[1.5cm][c]{\colorbox[gray]{0.85}{\makebox[\dimexpr 1.5cm-2\fboxsep\relax][c]{\raisebox{0pt}[\ht\strutbox][\dp\strutbox]{\textcolor{black}{\textbf{0.751{\tiny $\pm$0.03}}}}}}} & \makebox[1.5cm][c]{\bronze{0.718{\tiny $\pm$0.04}}} & \makebox[1.5cm][c]{0.703{\tiny $\pm$0.05}} \\
Alt.~Imp. & AS$\uparrow$ & \makebox[1.5cm][c]{\colorbox[gray]{0.85}{\makebox[\dimexpr 1.5cm-2\fboxsep\relax][c]{\raisebox{0pt}[\ht\strutbox][\dp\strutbox]{\textcolor{black}{\textbf{0.732{\tiny $\pm$0.04}}}}}}} & \makebox[1.5cm][c]{0.274{\tiny $\pm$0.08}} & \makebox[1.5cm][c]{0.564{\tiny $\pm$0.06}} & \makebox[1.5cm][c]{\colorbox[rgb]{1,0.95,0.8}{\makebox[\dimexpr 1.5cm-2\fboxsep\relax][c]{\raisebox{0pt}[\ht\strutbox][\dp\strutbox]{\textcolor{black}{\textbf{0.950{\tiny $\pm$0.03}}}}}}} & \makebox[1.5cm][c]{\bronze{0.567{\tiny $\pm$0.05}}} \\
 & FOSCTTM$\downarrow$ & \makebox[1.5cm][c]{\colorbox[rgb]{1,0.95,0.8}{\makebox[\dimexpr 1.5cm-2\fboxsep\relax][c]{\raisebox{0pt}[\ht\strutbox][\dp\strutbox]{\textcolor{black}{\textbf{0.313{\tiny $\pm$0.02}}}}}}} & \makebox[1.5cm][c]{0.379{\tiny $\pm$0.02}} & \makebox[1.5cm][c]{\colorbox[gray]{0.85}{\makebox[\dimexpr 1.5cm-2\fboxsep\relax][c]{\raisebox{0pt}[\ht\strutbox][\dp\strutbox]{\textcolor{black}{\textbf{0.334{\tiny $\pm$0.01}}}}}}} & \makebox[1.5cm][c]{0.368{\tiny $\pm$0.03}} & \makebox[1.5cm][c]{\bronze{0.362{\tiny $\pm$0.03}}} \\
\midrule
 & Accuracy$\uparrow$ & \makebox[1.5cm][c]{\colorbox[gray]{0.85}{\makebox[\dimexpr 1.5cm-2\fboxsep\relax][c]{\raisebox{0pt}[\ht\strutbox][\dp\strutbox]{\textcolor{black}{\textbf{0.773{\tiny $\pm$0.03}}}}}}} & \makebox[1.5cm][c]{\bronze{0.766{\tiny $\pm$0.02}}} & \makebox[1.5cm][c]{0.758{\tiny $\pm$0.03}} & \makebox[1.5cm][c]{0.761{\tiny $\pm$0.03}} & \makebox[1.5cm][c]{\colorbox[rgb]{1,0.95,0.8}{\makebox[\dimexpr 1.5cm-2\fboxsep\relax][c]{\raisebox{0pt}[\ht\strutbox][\dp\strutbox]{\textcolor{black}{\textbf{0.786{\tiny $\pm$0.03}}}}}}} \\
Distort & AS$\uparrow$ & \makebox[1.5cm][c]{\bronze{0.643{\tiny $\pm$0.04}}} & \makebox[1.5cm][c]{0.409{\tiny $\pm$0.04}} & \makebox[1.5cm][c]{0.553{\tiny $\pm$0.04}} & \makebox[1.5cm][c]{\colorbox[rgb]{1,0.95,0.8}{\makebox[\dimexpr 1.5cm-2\fboxsep\relax][c]{\raisebox{0pt}[\ht\strutbox][\dp\strutbox]{\textcolor{black}{\textbf{1.001{\tiny $\pm$0.02}}}}}}} & \makebox[1.5cm][c]{\colorbox[gray]{0.85}{\makebox[\dimexpr 1.5cm-2\fboxsep\relax][c]{\raisebox{0pt}[\ht\strutbox][\dp\strutbox]{\textcolor{black}{\textbf{0.965{\tiny $\pm$0.02}}}}}}} \\
 & FOSCTTM$\downarrow$ & \makebox[1.5cm][c]{\bronze{0.238{\tiny $\pm$0.02}}} & \makebox[1.5cm][c]{0.288{\tiny $\pm$0.02}} & \makebox[1.5cm][c]{0.303{\tiny $\pm$0.01}} & \makebox[1.5cm][c]{\colorbox[gray]{0.85}{\makebox[\dimexpr 1.5cm-2\fboxsep\relax][c]{\raisebox{0pt}[\ht\strutbox][\dp\strutbox]{\textcolor{black}{\textbf{0.211{\tiny $\pm$0.02}}}}}}} & \makebox[1.5cm][c]{\colorbox[rgb]{1,0.95,0.8}{\makebox[\dimexpr 1.5cm-2\fboxsep\relax][c]{\raisebox{0pt}[\ht\strutbox][\dp\strutbox]{\textcolor{black}{\textbf{0.156{\tiny $\pm$0.05}}}}}}} \\
\midrule
 & Accuracy$\uparrow$ & \makebox[1.5cm][c]{\colorbox[gray]{0.85}{\makebox[\dimexpr 1.5cm-2\fboxsep\relax][c]{\raisebox{0pt}[\ht\strutbox][\dp\strutbox]{\textcolor{black}{\textbf{0.720{\tiny $\pm$0.03}}}}}}} & \makebox[1.5cm][c]{\bronze{0.698{\tiny $\pm$0.03}}} & \makebox[1.5cm][c]{\colorbox[rgb]{1,0.95,0.8}{\makebox[\dimexpr 1.5cm-2\fboxsep\relax][c]{\raisebox{0pt}[\ht\strutbox][\dp\strutbox]{\textcolor{black}{\textbf{0.722{\tiny $\pm$0.03}}}}}}} & \makebox[1.5cm][c]{0.684{\tiny $\pm$0.04}} & \makebox[1.5cm][c]{0.664{\tiny $\pm$0.04}} \\
Importance & AS$\uparrow$ & \makebox[1.5cm][c]{\colorbox[gray]{0.85}{\makebox[\dimexpr 1.5cm-2\fboxsep\relax][c]{\raisebox{0pt}[\ht\strutbox][\dp\strutbox]{\textcolor{black}{\textbf{0.615{\tiny $\pm$0.04}}}}}}} & \makebox[1.5cm][c]{0.263{\tiny $\pm$0.05}} & \makebox[1.5cm][c]{0.477{\tiny $\pm$0.03}} & \makebox[1.5cm][c]{\colorbox[rgb]{1,0.95,0.8}{\makebox[\dimexpr 1.5cm-2\fboxsep\relax][c]{\raisebox{0pt}[\ht\strutbox][\dp\strutbox]{\textcolor{black}{\textbf{0.857{\tiny $\pm$0.02}}}}}}} & \makebox[1.5cm][c]{\bronze{0.533{\tiny $\pm$0.03}}} \\
 & FOSCTTM$\downarrow$ & \makebox[1.5cm][c]{\colorbox[rgb]{1,0.95,0.8}{\makebox[\dimexpr 1.5cm-2\fboxsep\relax][c]{\raisebox{0pt}[\ht\strutbox][\dp\strutbox]{\textcolor{black}{\textbf{0.329{\tiny $\pm$0.01}}}}}}} & \makebox[1.5cm][c]{0.387{\tiny $\pm$0.01}} & \makebox[1.5cm][c]{\colorbox[gray]{0.85}{\makebox[\dimexpr 1.5cm-2\fboxsep\relax][c]{\raisebox{0pt}[\ht\strutbox][\dp\strutbox]{\textcolor{black}{\textbf{0.345{\tiny $\pm$0.01}}}}}}} & \makebox[1.5cm][c]{0.378{\tiny $\pm$0.03}} & \makebox[1.5cm][c]{\bronze{0.370{\tiny $\pm$0.02}}} \\
\midrule
 & Accuracy$\uparrow$ & \makebox[1.5cm][c]{\colorbox[rgb]{1,0.95,0.8}{\makebox[\dimexpr 1.5cm-2\fboxsep\relax][c]{\raisebox{0pt}[\ht\strutbox][\dp\strutbox]{\textcolor{black}{\textbf{0.775{\tiny $\pm$0.04}}}}}}} & \makebox[1.5cm][c]{0.724{\tiny $\pm$0.05}} & \makebox[1.5cm][c]{\colorbox[gray]{0.85}{\makebox[\dimexpr 1.5cm-2\fboxsep\relax][c]{\raisebox{0pt}[\ht\strutbox][\dp\strutbox]{\textcolor{black}{\textbf{0.756{\tiny $\pm$0.05}}}}}}} & \makebox[1.5cm][c]{\bronze{0.727{\tiny $\pm$0.04}}} & \makebox[1.5cm][c]{0.694{\tiny $\pm$0.05}} \\
Random & AS$\uparrow$ & \makebox[1.5cm][c]{\colorbox[gray]{0.85}{\makebox[\dimexpr 1.5cm-2\fboxsep\relax][c]{\raisebox{0pt}[\ht\strutbox][\dp\strutbox]{\textcolor{black}{\textbf{0.707{\tiny $\pm$0.06}}}}}}} & \makebox[1.5cm][c]{0.287{\tiny $\pm$0.08}} & \makebox[1.5cm][c]{0.555{\tiny $\pm$0.08}} & \makebox[1.5cm][c]{\colorbox[rgb]{1,0.95,0.8}{\makebox[\dimexpr 1.5cm-2\fboxsep\relax][c]{\raisebox{0pt}[\ht\strutbox][\dp\strutbox]{\textcolor{black}{\textbf{0.946{\tiny $\pm$0.03}}}}}}} & \makebox[1.5cm][c]{\bronze{0.568{\tiny $\pm$0.07}}} \\
 & FOSCTTM$\downarrow$ & \makebox[1.5cm][c]{\colorbox[rgb]{1,0.95,0.8}{\makebox[\dimexpr 1.5cm-2\fboxsep\relax][c]{\raisebox{0pt}[\ht\strutbox][\dp\strutbox]{\textcolor{black}{\textbf{0.316{\tiny $\pm$0.02}}}}}}} & \makebox[1.5cm][c]{0.384{\tiny $\pm$0.02}} & \makebox[1.5cm][c]{\colorbox[gray]{0.85}{\makebox[\dimexpr 1.5cm-2\fboxsep\relax][c]{\raisebox{0pt}[\ht\strutbox][\dp\strutbox]{\textcolor{black}{\textbf{0.337{\tiny $\pm$0.02}}}}}}} & \makebox[1.5cm][c]{0.374{\tiny $\pm$0.03}} & \makebox[1.5cm][c]{\bronze{0.372{\tiny $\pm$0.03}}} \\
\midrule
 & Accuracy$\uparrow$ & \makebox[1.5cm][c]{0.807{\tiny $\pm$0.03}} & \makebox[1.5cm][c]{0.794{\tiny $\pm$0.02}} & \makebox[1.5cm][c]{\bronze{0.808{\tiny $\pm$0.02}}} & \makebox[1.5cm][c]{\colorbox[gray]{0.85}{\makebox[\dimexpr 1.5cm-2\fboxsep\relax][c]{\raisebox{0pt}[\ht\strutbox][\dp\strutbox]{\textcolor{black}{\textbf{0.819{\tiny $\pm$0.02}}}}}}} & \makebox[1.5cm][c]{\colorbox[rgb]{1,0.95,0.8}{\makebox[\dimexpr 1.5cm-2\fboxsep\relax][c]{\raisebox{0pt}[\ht\strutbox][\dp\strutbox]{\textcolor{black}{\textbf{0.832{\tiny $\pm$0.03}}}}}}} \\
Rotate & AS$\uparrow$ & \makebox[1.5cm][c]{\bronze{0.716{\tiny $\pm$0.05}}} & \makebox[1.5cm][c]{0.471{\tiny $\pm$0.07}} & \makebox[1.5cm][c]{0.600{\tiny $\pm$0.05}} & \makebox[1.5cm][c]{\colorbox[rgb]{1,0.95,0.8}{\makebox[\dimexpr 1.5cm-2\fboxsep\relax][c]{\raisebox{0pt}[\ht\strutbox][\dp\strutbox]{\textcolor{black}{\textbf{1.042{\tiny $\pm$0.02}}}}}}} & \makebox[1.5cm][c]{\colorbox[gray]{0.85}{\makebox[\dimexpr 1.5cm-2\fboxsep\relax][c]{\raisebox{0pt}[\ht\strutbox][\dp\strutbox]{\textcolor{black}{\textbf{1.018{\tiny $\pm$0.01}}}}}}} \\
 & FOSCTTM$\downarrow$ & \makebox[1.5cm][c]{\bronze{0.218{\tiny $\pm$0.02}}} & \makebox[1.5cm][c]{0.242{\tiny $\pm$0.03}} & \makebox[1.5cm][c]{0.266{\tiny $\pm$0.02}} & \makebox[1.5cm][c]{\colorbox[gray]{0.85}{\makebox[\dimexpr 1.5cm-2\fboxsep\relax][c]{\raisebox{0pt}[\ht\strutbox][\dp\strutbox]{\textcolor{black}{\textbf{0.109{\tiny $\pm$0.04}}}}}}} & \makebox[1.5cm][c]{\colorbox[rgb]{1,0.95,0.8}{\makebox[\dimexpr 1.5cm-2\fboxsep\relax][c]{\raisebox{0pt}[\ht\strutbox][\dp\strutbox]{\textcolor{black}{\textbf{0.075{\tiny $\pm$0.02}}}}}}} \\
\midrule
 & Accuracy$\uparrow$ & \makebox[1.5cm][c]{\colorbox[rgb]{1,0.95,0.8}{\makebox[\dimexpr 1.5cm-2\fboxsep\relax][c]{\raisebox{0pt}[\ht\strutbox][\dp\strutbox]{\textcolor{black}{\textbf{0.772{\tiny $\pm$0.03}}}}}}} & \makebox[1.5cm][c]{0.730{\tiny $\pm$0.03}} & \makebox[1.5cm][c]{\colorbox[gray]{0.85}{\makebox[\dimexpr 1.5cm-2\fboxsep\relax][c]{\raisebox{0pt}[\ht\strutbox][\dp\strutbox]{\textcolor{black}{\textbf{0.744{\tiny $\pm$0.03}}}}}}} & \makebox[1.5cm][c]{\bronze{0.733{\tiny $\pm$0.03}}} & \makebox[1.5cm][c]{0.716{\tiny $\pm$0.04}} \\
Average score & AS$\uparrow$ & \makebox[1.5cm][c]{\bronze{0.618{\tiny $\pm$0.04}}} & \makebox[1.5cm][c]{0.332{\tiny $\pm$0.06}} & \makebox[1.5cm][c]{0.507{\tiny $\pm$0.05}} & \makebox[1.5cm][c]{\colorbox[rgb]{1,0.95,0.8}{\makebox[\dimexpr 1.5cm-2\fboxsep\relax][c]{\raisebox{0pt}[\ht\strutbox][\dp\strutbox]{\textcolor{black}{\textbf{0.947{\tiny $\pm$0.02}}}}}}} & \makebox[1.5cm][c]{\colorbox[gray]{0.85}{\makebox[\dimexpr 1.5cm-2\fboxsep\relax][c]{\raisebox{0pt}[\ht\strutbox][\dp\strutbox]{\textcolor{black}{\textbf{0.765{\tiny $\pm$0.03}}}}}}} \\
 & FOSCTTM$\downarrow$ & \makebox[1.5cm][c]{\colorbox[rgb]{1,0.95,0.8}{\makebox[\dimexpr 1.5cm-2\fboxsep\relax][c]{\raisebox{0pt}[\ht\strutbox][\dp\strutbox]{\textcolor{black}{\textbf{0.278{\tiny $\pm$0.02}}}}}}} & \makebox[1.5cm][c]{0.337{\tiny $\pm$0.02}} & \makebox[1.5cm][c]{0.320{\tiny $\pm$0.01}} & \makebox[1.5cm][c]{\bronze{0.295{\tiny $\pm$0.03}}} & \makebox[1.5cm][c]{\colorbox[gray]{0.85}{\makebox[\dimexpr 1.5cm-2\fboxsep\relax][c]{\raisebox{0pt}[\ht\strutbox][\dp\strutbox]{\textcolor{black}{\textbf{0.283{\tiny $\pm$0.03}}}}}}} \\
\bottomrule
\end{tabular}
\end{table}

%% file: tables/uci_general_ablation.tex
\begin{table}[t]
\caption{\textbf{FoSTA ablation performance over UCI datasets and seeds under the selected distortion splits.} Each cell reports mean $\pm$ the average within-dataset standard deviation across five seeds. Results are shown for label transfer accuracy (Acc), alignment score (AS), and correspondence recovery measured by FOSCTTM (FOS). Higher is better for accuracy and AS, while lower is better for FOSCTTM. The top three results for each metric are highlighted in gold (1st), silver (2nd), and bronze (3rd). Each variant is ranked independently against non-FoSTA baselines in Table~\ref{tab:uci}. The final three rows report average scores across the displayed splits, with standard deviations also averaged across splits.}
\label{tab:uci_ablation}
\centering
\small
\setlength{\tabcolsep}{2pt}
\setlength{\fboxsep}{1pt}
\renewcommand{\arraystretch}{1.0}

\begin{adjustbox}{max width=\linewidth}
\begin{tabular}{lr c c c c c c c c c c}
\toprule
Split & Metric & \makebox[1.5cm][c]{\textbf{Default}} & \makebox[1.5cm][c]{RotF} & \makebox[1.5cm][c]{ET} & \makebox[1.5cm][c]{KeRF} & \makebox[1.5cm][c]{DPT} & \makebox[1.5cm][c]{Dense OT} & \makebox[1.5cm][c]{No Prop.} & \makebox[1.5cm][c]{Auto. $t$} & \makebox[1.5cm][c]{UMAP} & \makebox[1.5cm][c]{LE} \\
\midrule
 & Acc$\uparrow$ & \makebox[1.5cm][c]{\colorbox[rgb]{1,0.95,0.8}{\makebox[\dimexpr 1.5cm-2\fboxsep\relax][c]{\raisebox{0pt}[\ht\strutbox][\dp\strutbox]{\textcolor{black}{\textbf{0.788{\tiny $\pm$0.02}}}}}}} & \makebox[1.5cm][c]{\colorbox[rgb]{1,0.95,0.8}{\makebox[\dimexpr 1.5cm-2\fboxsep\relax][c]{\raisebox{0pt}[\ht\strutbox][\dp\strutbox]{\textcolor{black}{\textbf{0.799{\tiny $\pm$0.02}}}}}}} & \makebox[1.5cm][c]{\colorbox[rgb]{1,0.95,0.8}{\makebox[\dimexpr 1.5cm-2\fboxsep\relax][c]{\raisebox{0pt}[\ht\strutbox][\dp\strutbox]{\textcolor{black}{\textbf{0.788{\tiny $\pm$0.03}}}}}}} & \makebox[1.5cm][c]{\colorbox[rgb]{1,0.95,0.8}{\makebox[\dimexpr 1.5cm-2\fboxsep\relax][c]{\raisebox{0pt}[\ht\strutbox][\dp\strutbox]{\textcolor{black}{\textbf{0.768{\tiny $\pm$0.03}}}}}}} & \makebox[1.5cm][c]{\colorbox[rgb]{1,0.95,0.8}{\makebox[\dimexpr 1.5cm-2\fboxsep\relax][c]{\raisebox{0pt}[\ht\strutbox][\dp\strutbox]{\textcolor{black}{\textbf{0.783{\tiny $\pm$0.03}}}}}}} & \makebox[1.5cm][c]{\colorbox[rgb]{1,0.95,0.8}{\makebox[\dimexpr 1.5cm-2\fboxsep\relax][c]{\raisebox{0pt}[\ht\strutbox][\dp\strutbox]{\textcolor{black}{\textbf{0.784{\tiny $\pm$0.02}}}}}}} & \makebox[1.5cm][c]{0.629{\tiny $\pm$0.08}} & \makebox[1.5cm][c]{\colorbox[rgb]{1,0.95,0.8}{\makebox[\dimexpr 1.5cm-2\fboxsep\relax][c]{\raisebox{0pt}[\ht\strutbox][\dp\strutbox]{\textcolor{black}{\textbf{0.770{\tiny $\pm$0.03}}}}}}} & \makebox[1.5cm][c]{\colorbox[rgb]{1,0.95,0.8}{\makebox[\dimexpr 1.5cm-2\fboxsep\relax][c]{\raisebox{0pt}[\ht\strutbox][\dp\strutbox]{\textcolor{black}{\textbf{0.787{\tiny $\pm$0.03}}}}}}} & \makebox[1.5cm][c]{\colorbox[rgb]{1,0.95,0.8}{\makebox[\dimexpr 1.5cm-2\fboxsep\relax][c]{\raisebox{0pt}[\ht\strutbox][\dp\strutbox]{\textcolor{black}{\textbf{0.769{\tiny $\pm$0.02}}}}}}} \\
Noise & AS$\uparrow$ & \makebox[1.5cm][c]{\bronze{0.294{\tiny $\pm$0.02}}} & \makebox[1.5cm][c]{\bronze{0.462{\tiny $\pm$0.04}}} & \makebox[1.5cm][c]{0.278{\tiny $\pm$0.03}} & \makebox[1.5cm][c]{\bronze{0.312{\tiny $\pm$0.02}}} & \makebox[1.5cm][c]{\bronze{0.315{\tiny $\pm$0.03}}} & \makebox[1.5cm][c]{0.230{\tiny $\pm$0.02}} & \makebox[1.5cm][c]{0.000{\tiny $\pm$0.00}} & \makebox[1.5cm][c]{\bronze{0.448{\tiny $\pm$0.02}}} & \makebox[1.5cm][c]{\colorbox[rgb]{1,0.95,0.8}{\makebox[\dimexpr 1.5cm-2\fboxsep\relax][c]{\raisebox{0pt}[\ht\strutbox][\dp\strutbox]{\textcolor{black}{\textbf{0.973{\tiny $\pm$0.01}}}}}}} & \makebox[1.5cm][c]{\bronze{0.546{\tiny $\pm$0.03}}} \\
 & FOS$\downarrow$ & \makebox[1.5cm][c]{\colorbox[rgb]{1,0.95,0.8}{\makebox[\dimexpr 1.5cm-2\fboxsep\relax][c]{\raisebox{0pt}[\ht\strutbox][\dp\strutbox]{\textcolor{black}{\textbf{0.251{\tiny $\pm$0.02}}}}}}} & \makebox[1.5cm][c]{\colorbox[rgb]{1,0.95,0.8}{\makebox[\dimexpr 1.5cm-2\fboxsep\relax][c]{\raisebox{0pt}[\ht\strutbox][\dp\strutbox]{\textcolor{black}{\textbf{0.230{\tiny $\pm$0.02}}}}}}} & \makebox[1.5cm][c]{\colorbox[rgb]{1,0.95,0.8}{\makebox[\dimexpr 1.5cm-2\fboxsep\relax][c]{\raisebox{0pt}[\ht\strutbox][\dp\strutbox]{\textcolor{black}{\textbf{0.247{\tiny $\pm$0.02}}}}}}} & \makebox[1.5cm][c]{\colorbox[rgb]{1,0.95,0.8}{\makebox[\dimexpr 1.5cm-2\fboxsep\relax][c]{\raisebox{0pt}[\ht\strutbox][\dp\strutbox]{\textcolor{black}{\textbf{0.253{\tiny $\pm$0.02}}}}}}} & \makebox[1.5cm][c]{\colorbox[rgb]{1,0.95,0.8}{\makebox[\dimexpr 1.5cm-2\fboxsep\relax][c]{\raisebox{0pt}[\ht\strutbox][\dp\strutbox]{\textcolor{black}{\textbf{0.244{\tiny $\pm$0.02}}}}}}} & \makebox[1.5cm][c]{\colorbox[rgb]{1,0.95,0.8}{\makebox[\dimexpr 1.5cm-2\fboxsep\relax][c]{\raisebox{0pt}[\ht\strutbox][\dp\strutbox]{\textcolor{black}{\textbf{0.242{\tiny $\pm$0.02}}}}}}} & \makebox[1.5cm][c]{\colorbox[gray]{0.85}{\makebox[\dimexpr 1.5cm-2\fboxsep\relax][c]{\raisebox{0pt}[\ht\strutbox][\dp\strutbox]{\textcolor{black}{\textbf{0.333{\tiny $\pm$0.01}}}}}}} & \makebox[1.5cm][c]{\colorbox[rgb]{1,0.95,0.8}{\makebox[\dimexpr 1.5cm-2\fboxsep\relax][c]{\raisebox{0pt}[\ht\strutbox][\dp\strutbox]{\textcolor{black}{\textbf{0.251{\tiny $\pm$0.01}}}}}}} & \makebox[1.5cm][c]{\colorbox[rgb]{1,0.95,0.8}{\makebox[\dimexpr 1.5cm-2\fboxsep\relax][c]{\raisebox{0pt}[\ht\strutbox][\dp\strutbox]{\textcolor{black}{\textbf{0.302{\tiny $\pm$0.01}}}}}}} & \makebox[1.5cm][c]{\colorbox[rgb]{1,0.95,0.8}{\makebox[\dimexpr 1.5cm-2\fboxsep\relax][c]{\raisebox{0pt}[\ht\strutbox][\dp\strutbox]{\textcolor{black}{\textbf{0.270{\tiny $\pm$0.02}}}}}}} \\
\midrule
 & Acc$\uparrow$ & \makebox[1.5cm][c]{\colorbox[rgb]{1,0.95,0.8}{\makebox[\dimexpr 1.5cm-2\fboxsep\relax][c]{\raisebox{0pt}[\ht\strutbox][\dp\strutbox]{\textcolor{black}{\textbf{0.767{\tiny $\pm$0.03}}}}}}} & \makebox[1.5cm][c]{\colorbox[rgb]{1,0.95,0.8}{\makebox[\dimexpr 1.5cm-2\fboxsep\relax][c]{\raisebox{0pt}[\ht\strutbox][\dp\strutbox]{\textcolor{black}{\textbf{0.766{\tiny $\pm$0.03}}}}}}} & \makebox[1.5cm][c]{\colorbox[rgb]{1,0.95,0.8}{\makebox[\dimexpr 1.5cm-2\fboxsep\relax][c]{\raisebox{0pt}[\ht\strutbox][\dp\strutbox]{\textcolor{black}{\textbf{0.763{\tiny $\pm$0.02}}}}}}} & \makebox[1.5cm][c]{\colorbox[rgb]{1,0.95,0.8}{\makebox[\dimexpr 1.5cm-2\fboxsep\relax][c]{\raisebox{0pt}[\ht\strutbox][\dp\strutbox]{\textcolor{black}{\textbf{0.760{\tiny $\pm$0.03}}}}}}} & \makebox[1.5cm][c]{\colorbox[rgb]{1,0.95,0.8}{\makebox[\dimexpr 1.5cm-2\fboxsep\relax][c]{\raisebox{0pt}[\ht\strutbox][\dp\strutbox]{\textcolor{black}{\textbf{0.755{\tiny $\pm$0.03}}}}}}} & \makebox[1.5cm][c]{\colorbox[rgb]{1,0.95,0.8}{\makebox[\dimexpr 1.5cm-2\fboxsep\relax][c]{\raisebox{0pt}[\ht\strutbox][\dp\strutbox]{\textcolor{black}{\textbf{0.769{\tiny $\pm$0.03}}}}}}} & \makebox[1.5cm][c]{\colorbox[gray]{0.85}{\makebox[\dimexpr 1.5cm-2\fboxsep\relax][c]{\raisebox{0pt}[\ht\strutbox][\dp\strutbox]{\textcolor{black}{\textbf{0.735{\tiny $\pm$0.04}}}}}}} & \makebox[1.5cm][c]{\colorbox[rgb]{1,0.95,0.8}{\makebox[\dimexpr 1.5cm-2\fboxsep\relax][c]{\raisebox{0pt}[\ht\strutbox][\dp\strutbox]{\textcolor{black}{\textbf{0.769{\tiny $\pm$0.02}}}}}}} & \makebox[1.5cm][c]{\colorbox[rgb]{1,0.95,0.8}{\makebox[\dimexpr 1.5cm-2\fboxsep\relax][c]{\raisebox{0pt}[\ht\strutbox][\dp\strutbox]{\textcolor{black}{\textbf{0.769{\tiny $\pm$0.02}}}}}}} & \makebox[1.5cm][c]{\colorbox[rgb]{1,0.95,0.8}{\makebox[\dimexpr 1.5cm-2\fboxsep\relax][c]{\raisebox{0pt}[\ht\strutbox][\dp\strutbox]{\textcolor{black}{\textbf{0.760{\tiny $\pm$0.02}}}}}}} \\
Alt.~Imp. & AS$\uparrow$ & \makebox[1.5cm][c]{\colorbox[gray]{0.85}{\makebox[\dimexpr 1.5cm-2\fboxsep\relax][c]{\raisebox{0pt}[\ht\strutbox][\dp\strutbox]{\textcolor{black}{\textbf{0.732{\tiny $\pm$0.04}}}}}}} & \makebox[1.5cm][c]{\colorbox[gray]{0.85}{\makebox[\dimexpr 1.5cm-2\fboxsep\relax][c]{\raisebox{0pt}[\ht\strutbox][\dp\strutbox]{\textcolor{black}{\textbf{0.693{\tiny $\pm$0.05}}}}}}} & \makebox[1.5cm][c]{\colorbox[gray]{0.85}{\makebox[\dimexpr 1.5cm-2\fboxsep\relax][c]{\raisebox{0pt}[\ht\strutbox][\dp\strutbox]{\textcolor{black}{\textbf{0.820{\tiny $\pm$0.04}}}}}}} & \makebox[1.5cm][c]{\colorbox[gray]{0.85}{\makebox[\dimexpr 1.5cm-2\fboxsep\relax][c]{\raisebox{0pt}[\ht\strutbox][\dp\strutbox]{\textcolor{black}{\textbf{0.717{\tiny $\pm$0.05}}}}}}} & \makebox[1.5cm][c]{\colorbox[gray]{0.85}{\makebox[\dimexpr 1.5cm-2\fboxsep\relax][c]{\raisebox{0pt}[\ht\strutbox][\dp\strutbox]{\textcolor{black}{\textbf{0.746{\tiny $\pm$0.05}}}}}}} & \makebox[1.5cm][c]{\colorbox[gray]{0.85}{\makebox[\dimexpr 1.5cm-2\fboxsep\relax][c]{\raisebox{0pt}[\ht\strutbox][\dp\strutbox]{\textcolor{black}{\textbf{0.732{\tiny $\pm$0.05}}}}}}} & \makebox[1.5cm][c]{0.154{\tiny $\pm$0.04}} & \makebox[1.5cm][c]{\colorbox[gray]{0.85}{\makebox[\dimexpr 1.5cm-2\fboxsep\relax][c]{\raisebox{0pt}[\ht\strutbox][\dp\strutbox]{\textcolor{black}{\textbf{0.767{\tiny $\pm$0.05}}}}}}} & \makebox[1.5cm][c]{\colorbox[gray]{0.85}{\makebox[\dimexpr 1.5cm-2\fboxsep\relax][c]{\raisebox{0pt}[\ht\strutbox][\dp\strutbox]{\textcolor{black}{\textbf{0.945{\tiny $\pm$0.02}}}}}}} & \makebox[1.5cm][c]{\colorbox[gray]{0.85}{\makebox[\dimexpr 1.5cm-2\fboxsep\relax][c]{\raisebox{0pt}[\ht\strutbox][\dp\strutbox]{\textcolor{black}{\textbf{0.775{\tiny $\pm$0.05}}}}}}} \\
 & FOS$\downarrow$ & \makebox[1.5cm][c]{\colorbox[rgb]{1,0.95,0.8}{\makebox[\dimexpr 1.5cm-2\fboxsep\relax][c]{\raisebox{0pt}[\ht\strutbox][\dp\strutbox]{\textcolor{black}{\textbf{0.313{\tiny $\pm$0.02}}}}}}} & \makebox[1.5cm][c]{\colorbox[rgb]{1,0.95,0.8}{\makebox[\dimexpr 1.5cm-2\fboxsep\relax][c]{\raisebox{0pt}[\ht\strutbox][\dp\strutbox]{\textcolor{black}{\textbf{0.307{\tiny $\pm$0.01}}}}}}} & \makebox[1.5cm][c]{\colorbox[rgb]{1,0.95,0.8}{\makebox[\dimexpr 1.5cm-2\fboxsep\relax][c]{\raisebox{0pt}[\ht\strutbox][\dp\strutbox]{\textcolor{black}{\textbf{0.309{\tiny $\pm$0.02}}}}}}} & \makebox[1.5cm][c]{\colorbox[rgb]{1,0.95,0.8}{\makebox[\dimexpr 1.5cm-2\fboxsep\relax][c]{\raisebox{0pt}[\ht\strutbox][\dp\strutbox]{\textcolor{black}{\textbf{0.317{\tiny $\pm$0.02}}}}}}} & \makebox[1.5cm][c]{\colorbox[rgb]{1,0.95,0.8}{\makebox[\dimexpr 1.5cm-2\fboxsep\relax][c]{\raisebox{0pt}[\ht\strutbox][\dp\strutbox]{\textcolor{black}{\textbf{0.320{\tiny $\pm$0.02}}}}}}} & \makebox[1.5cm][c]{\colorbox[rgb]{1,0.95,0.8}{\makebox[\dimexpr 1.5cm-2\fboxsep\relax][c]{\raisebox{0pt}[\ht\strutbox][\dp\strutbox]{\textcolor{black}{\textbf{0.307{\tiny $\pm$0.02}}}}}}} & \makebox[1.5cm][c]{\colorbox[gray]{0.85}{\makebox[\dimexpr 1.5cm-2\fboxsep\relax][c]{\raisebox{0pt}[\ht\strutbox][\dp\strutbox]{\textcolor{black}{\textbf{0.335{\tiny $\pm$0.01}}}}}}} & \makebox[1.5cm][c]{\colorbox[rgb]{1,0.95,0.8}{\makebox[\dimexpr 1.5cm-2\fboxsep\relax][c]{\raisebox{0pt}[\ht\strutbox][\dp\strutbox]{\textcolor{black}{\textbf{0.308{\tiny $\pm$0.02}}}}}}} & \makebox[1.5cm][c]{\colorbox[rgb]{1,0.95,0.8}{\makebox[\dimexpr 1.5cm-2\fboxsep\relax][c]{\raisebox{0pt}[\ht\strutbox][\dp\strutbox]{\textcolor{black}{\textbf{0.321{\tiny $\pm$0.02}}}}}}} & \makebox[1.5cm][c]{\colorbox[rgb]{1,0.95,0.8}{\makebox[\dimexpr 1.5cm-2\fboxsep\relax][c]{\raisebox{0pt}[\ht\strutbox][\dp\strutbox]{\textcolor{black}{\textbf{0.332{\tiny $\pm$0.02}}}}}}} \\
\midrule
 & Acc$\uparrow$ & \makebox[1.5cm][c]{\colorbox[gray]{0.85}{\makebox[\dimexpr 1.5cm-2\fboxsep\relax][c]{\raisebox{0pt}[\ht\strutbox][\dp\strutbox]{\textcolor{black}{\textbf{0.773{\tiny $\pm$0.03}}}}}}} & \makebox[1.5cm][c]{\colorbox[gray]{0.85}{\makebox[\dimexpr 1.5cm-2\fboxsep\relax][c]{\raisebox{0pt}[\ht\strutbox][\dp\strutbox]{\textcolor{black}{\textbf{0.776{\tiny $\pm$0.03}}}}}}} & \makebox[1.5cm][c]{\colorbox[gray]{0.85}{\makebox[\dimexpr 1.5cm-2\fboxsep\relax][c]{\raisebox{0pt}[\ht\strutbox][\dp\strutbox]{\textcolor{black}{\textbf{0.779{\tiny $\pm$0.03}}}}}}} & \makebox[1.5cm][c]{\bronze{0.763{\tiny $\pm$0.03}}} & \makebox[1.5cm][c]{\colorbox[gray]{0.85}{\makebox[\dimexpr 1.5cm-2\fboxsep\relax][c]{\raisebox{0pt}[\ht\strutbox][\dp\strutbox]{\textcolor{black}{\textbf{0.767{\tiny $\pm$0.03}}}}}}} & \makebox[1.5cm][c]{\colorbox[gray]{0.85}{\makebox[\dimexpr 1.5cm-2\fboxsep\relax][c]{\raisebox{0pt}[\ht\strutbox][\dp\strutbox]{\textcolor{black}{\textbf{0.770{\tiny $\pm$0.03}}}}}}} & \makebox[1.5cm][c]{0.697{\tiny $\pm$0.05}} & \makebox[1.5cm][c]{\colorbox[gray]{0.85}{\makebox[\dimexpr 1.5cm-2\fboxsep\relax][c]{\raisebox{0pt}[\ht\strutbox][\dp\strutbox]{\textcolor{black}{\textbf{0.767{\tiny $\pm$0.03}}}}}}} & \makebox[1.5cm][c]{\colorbox[gray]{0.85}{\makebox[\dimexpr 1.5cm-2\fboxsep\relax][c]{\raisebox{0pt}[\ht\strutbox][\dp\strutbox]{\textcolor{black}{\textbf{0.774{\tiny $\pm$0.03}}}}}}} & \makebox[1.5cm][c]{\bronze{0.764{\tiny $\pm$0.03}}} \\
Distort & AS$\uparrow$ & \makebox[1.5cm][c]{\bronze{0.643{\tiny $\pm$0.04}}} & \makebox[1.5cm][c]{\bronze{0.629{\tiny $\pm$0.03}}} & \makebox[1.5cm][c]{\bronze{0.717{\tiny $\pm$0.04}}} & \makebox[1.5cm][c]{\bronze{0.616{\tiny $\pm$0.03}}} & \makebox[1.5cm][c]{\bronze{0.647{\tiny $\pm$0.03}}} & \makebox[1.5cm][c]{\bronze{0.657{\tiny $\pm$0.04}}} & \makebox[1.5cm][c]{0.003{\tiny $\pm$0.00}} & \makebox[1.5cm][c]{\bronze{0.746{\tiny $\pm$0.03}}} & \makebox[1.5cm][c]{\colorbox[gray]{0.85}{\makebox[\dimexpr 1.5cm-2\fboxsep\relax][c]{\raisebox{0pt}[\ht\strutbox][\dp\strutbox]{\textcolor{black}{\textbf{0.973{\tiny $\pm$0.02}}}}}}} & \makebox[1.5cm][c]{\bronze{0.783{\tiny $\pm$0.03}}} \\
 & FOS$\downarrow$ & \makebox[1.5cm][c]{\bronze{0.238{\tiny $\pm$0.02}}} & \makebox[1.5cm][c]{\bronze{0.226{\tiny $\pm$0.02}}} & \makebox[1.5cm][c]{\bronze{0.225{\tiny $\pm$0.02}}} & \makebox[1.5cm][c]{\bronze{0.235{\tiny $\pm$0.02}}} & \makebox[1.5cm][c]{\bronze{0.225{\tiny $\pm$0.02}}} & \makebox[1.5cm][c]{\bronze{0.232{\tiny $\pm$0.02}}} & \makebox[1.5cm][c]{0.304{\tiny $\pm$0.01}} & \makebox[1.5cm][c]{\bronze{0.231{\tiny $\pm$0.01}}} & \makebox[1.5cm][c]{\bronze{0.271{\tiny $\pm$0.01}}} & \makebox[1.5cm][c]{\bronze{0.258{\tiny $\pm$0.02}}} \\
\midrule
 & Acc$\uparrow$ & \makebox[1.5cm][c]{\colorbox[gray]{0.85}{\makebox[\dimexpr 1.5cm-2\fboxsep\relax][c]{\raisebox{0pt}[\ht\strutbox][\dp\strutbox]{\textcolor{black}{\textbf{0.720{\tiny $\pm$0.03}}}}}}} & \makebox[1.5cm][c]{\colorbox[rgb]{1,0.95,0.8}{\makebox[\dimexpr 1.5cm-2\fboxsep\relax][c]{\raisebox{0pt}[\ht\strutbox][\dp\strutbox]{\textcolor{black}{\textbf{0.726{\tiny $\pm$0.03}}}}}}} & \makebox[1.5cm][c]{\colorbox[rgb]{1,0.95,0.8}{\makebox[\dimexpr 1.5cm-2\fboxsep\relax][c]{\raisebox{0pt}[\ht\strutbox][\dp\strutbox]{\textcolor{black}{\textbf{0.724{\tiny $\pm$0.03}}}}}}} & \makebox[1.5cm][c]{\colorbox[rgb]{1,0.95,0.8}{\makebox[\dimexpr 1.5cm-2\fboxsep\relax][c]{\raisebox{0pt}[\ht\strutbox][\dp\strutbox]{\textcolor{black}{\textbf{0.722{\tiny $\pm$0.03}}}}}}} & \makebox[1.5cm][c]{0.679{\tiny $\pm$0.05}} & \makebox[1.5cm][c]{\colorbox[rgb]{1,0.95,0.8}{\makebox[\dimexpr 1.5cm-2\fboxsep\relax][c]{\raisebox{0pt}[\ht\strutbox][\dp\strutbox]{\textcolor{black}{\textbf{0.724{\tiny $\pm$0.04}}}}}}} & \makebox[1.5cm][c]{0.661{\tiny $\pm$0.06}} & \makebox[1.5cm][c]{\colorbox[rgb]{1,0.95,0.8}{\makebox[\dimexpr 1.5cm-2\fboxsep\relax][c]{\raisebox{0pt}[\ht\strutbox][\dp\strutbox]{\textcolor{black}{\textbf{0.723{\tiny $\pm$0.03}}}}}}} & \makebox[1.5cm][c]{\colorbox[gray]{0.85}{\makebox[\dimexpr 1.5cm-2\fboxsep\relax][c]{\raisebox{0pt}[\ht\strutbox][\dp\strutbox]{\textcolor{black}{\textbf{0.720{\tiny $\pm$0.04}}}}}}} & \makebox[1.5cm][c]{\colorbox[gray]{0.85}{\makebox[\dimexpr 1.5cm-2\fboxsep\relax][c]{\raisebox{0pt}[\ht\strutbox][\dp\strutbox]{\textcolor{black}{\textbf{0.717{\tiny $\pm$0.03}}}}}}} \\
Importance & AS$\uparrow$ & \makebox[1.5cm][c]{\colorbox[gray]{0.85}{\makebox[\dimexpr 1.5cm-2\fboxsep\relax][c]{\raisebox{0pt}[\ht\strutbox][\dp\strutbox]{\textcolor{black}{\textbf{0.615{\tiny $\pm$0.04}}}}}}} & \makebox[1.5cm][c]{\colorbox[gray]{0.85}{\makebox[\dimexpr 1.5cm-2\fboxsep\relax][c]{\raisebox{0pt}[\ht\strutbox][\dp\strutbox]{\textcolor{black}{\textbf{0.598{\tiny $\pm$0.04}}}}}}} & \makebox[1.5cm][c]{\colorbox[gray]{0.85}{\makebox[\dimexpr 1.5cm-2\fboxsep\relax][c]{\raisebox{0pt}[\ht\strutbox][\dp\strutbox]{\textcolor{black}{\textbf{0.694{\tiny $\pm$0.04}}}}}}} & \makebox[1.5cm][c]{\colorbox[gray]{0.85}{\makebox[\dimexpr 1.5cm-2\fboxsep\relax][c]{\raisebox{0pt}[\ht\strutbox][\dp\strutbox]{\textcolor{black}{\textbf{0.597{\tiny $\pm$0.04}}}}}}} & \makebox[1.5cm][c]{\colorbox[gray]{0.85}{\makebox[\dimexpr 1.5cm-2\fboxsep\relax][c]{\raisebox{0pt}[\ht\strutbox][\dp\strutbox]{\textcolor{black}{\textbf{0.641{\tiny $\pm$0.04}}}}}}} & \makebox[1.5cm][c]{\colorbox[gray]{0.85}{\makebox[\dimexpr 1.5cm-2\fboxsep\relax][c]{\raisebox{0pt}[\ht\strutbox][\dp\strutbox]{\textcolor{black}{\textbf{0.605{\tiny $\pm$0.05}}}}}}} & \makebox[1.5cm][c]{0.118{\tiny $\pm$0.03}} & \makebox[1.5cm][c]{\colorbox[gray]{0.85}{\makebox[\dimexpr 1.5cm-2\fboxsep\relax][c]{\raisebox{0pt}[\ht\strutbox][\dp\strutbox]{\textcolor{black}{\textbf{0.668{\tiny $\pm$0.04}}}}}}} & \makebox[1.5cm][c]{\colorbox[rgb]{1,0.95,0.8}{\makebox[\dimexpr 1.5cm-2\fboxsep\relax][c]{\raisebox{0pt}[\ht\strutbox][\dp\strutbox]{\textcolor{black}{\textbf{0.912{\tiny $\pm$0.02}}}}}}} & \makebox[1.5cm][c]{\colorbox[gray]{0.85}{\makebox[\dimexpr 1.5cm-2\fboxsep\relax][c]{\raisebox{0pt}[\ht\strutbox][\dp\strutbox]{\textcolor{black}{\textbf{0.702{\tiny $\pm$0.03}}}}}}} \\
 & FOS$\downarrow$ & \makebox[1.5cm][c]{\colorbox[rgb]{1,0.95,0.8}{\makebox[\dimexpr 1.5cm-2\fboxsep\relax][c]{\raisebox{0pt}[\ht\strutbox][\dp\strutbox]{\textcolor{black}{\textbf{0.329{\tiny $\pm$0.01}}}}}}} & \makebox[1.5cm][c]{\colorbox[rgb]{1,0.95,0.8}{\makebox[\dimexpr 1.5cm-2\fboxsep\relax][c]{\raisebox{0pt}[\ht\strutbox][\dp\strutbox]{\textcolor{black}{\textbf{0.321{\tiny $\pm$0.01}}}}}}} & \makebox[1.5cm][c]{\colorbox[rgb]{1,0.95,0.8}{\makebox[\dimexpr 1.5cm-2\fboxsep\relax][c]{\raisebox{0pt}[\ht\strutbox][\dp\strutbox]{\textcolor{black}{\textbf{0.327{\tiny $\pm$0.01}}}}}}} & \makebox[1.5cm][c]{\colorbox[rgb]{1,0.95,0.8}{\makebox[\dimexpr 1.5cm-2\fboxsep\relax][c]{\raisebox{0pt}[\ht\strutbox][\dp\strutbox]{\textcolor{black}{\textbf{0.330{\tiny $\pm$0.02}}}}}}} & \makebox[1.5cm][c]{\colorbox[gray]{0.85}{\makebox[\dimexpr 1.5cm-2\fboxsep\relax][c]{\raisebox{0pt}[\ht\strutbox][\dp\strutbox]{\textcolor{black}{\textbf{0.352{\tiny $\pm$0.02}}}}}}} & \makebox[1.5cm][c]{\colorbox[rgb]{1,0.95,0.8}{\makebox[\dimexpr 1.5cm-2\fboxsep\relax][c]{\raisebox{0pt}[\ht\strutbox][\dp\strutbox]{\textcolor{black}{\textbf{0.325{\tiny $\pm$0.01}}}}}}} & \makebox[1.5cm][c]{\colorbox[gray]{0.85}{\makebox[\dimexpr 1.5cm-2\fboxsep\relax][c]{\raisebox{0pt}[\ht\strutbox][\dp\strutbox]{\textcolor{black}{\textbf{0.353{\tiny $\pm$0.01}}}}}}} & \makebox[1.5cm][c]{\colorbox[rgb]{1,0.95,0.8}{\makebox[\dimexpr 1.5cm-2\fboxsep\relax][c]{\raisebox{0pt}[\ht\strutbox][\dp\strutbox]{\textcolor{black}{\textbf{0.324{\tiny $\pm$0.01}}}}}}} & \makebox[1.5cm][c]{\colorbox[rgb]{1,0.95,0.8}{\makebox[\dimexpr 1.5cm-2\fboxsep\relax][c]{\raisebox{0pt}[\ht\strutbox][\dp\strutbox]{\textcolor{black}{\textbf{0.334{\tiny $\pm$0.01}}}}}}} & \makebox[1.5cm][c]{\colorbox[rgb]{1,0.95,0.8}{\makebox[\dimexpr 1.5cm-2\fboxsep\relax][c]{\raisebox{0pt}[\ht\strutbox][\dp\strutbox]{\textcolor{black}{\textbf{0.344{\tiny $\pm$0.01}}}}}}} \\
\midrule
 & Acc$\uparrow$ & \makebox[1.5cm][c]{\colorbox[rgb]{1,0.95,0.8}{\makebox[\dimexpr 1.5cm-2\fboxsep\relax][c]{\raisebox{0pt}[\ht\strutbox][\dp\strutbox]{\textcolor{black}{\textbf{0.775{\tiny $\pm$0.04}}}}}}} & \makebox[1.5cm][c]{\colorbox[rgb]{1,0.95,0.8}{\makebox[\dimexpr 1.5cm-2\fboxsep\relax][c]{\raisebox{0pt}[\ht\strutbox][\dp\strutbox]{\textcolor{black}{\textbf{0.780{\tiny $\pm$0.04}}}}}}} & \makebox[1.5cm][c]{\colorbox[rgb]{1,0.95,0.8}{\makebox[\dimexpr 1.5cm-2\fboxsep\relax][c]{\raisebox{0pt}[\ht\strutbox][\dp\strutbox]{\textcolor{black}{\textbf{0.772{\tiny $\pm$0.05}}}}}}} & \makebox[1.5cm][c]{\colorbox[rgb]{1,0.95,0.8}{\makebox[\dimexpr 1.5cm-2\fboxsep\relax][c]{\raisebox{0pt}[\ht\strutbox][\dp\strutbox]{\textcolor{black}{\textbf{0.772{\tiny $\pm$0.04}}}}}}} & \makebox[1.5cm][c]{\colorbox[gray]{0.85}{\makebox[\dimexpr 1.5cm-2\fboxsep\relax][c]{\raisebox{0pt}[\ht\strutbox][\dp\strutbox]{\textcolor{black}{\textbf{0.739{\tiny $\pm$0.06}}}}}}} & \makebox[1.5cm][c]{\colorbox[rgb]{1,0.95,0.8}{\makebox[\dimexpr 1.5cm-2\fboxsep\relax][c]{\raisebox{0pt}[\ht\strutbox][\dp\strutbox]{\textcolor{black}{\textbf{0.776{\tiny $\pm$0.04}}}}}}} & \makebox[1.5cm][c]{\colorbox[gray]{0.85}{\makebox[\dimexpr 1.5cm-2\fboxsep\relax][c]{\raisebox{0pt}[\ht\strutbox][\dp\strutbox]{\textcolor{black}{\textbf{0.740{\tiny $\pm$0.07}}}}}}} & \makebox[1.5cm][c]{\colorbox[rgb]{1,0.95,0.8}{\makebox[\dimexpr 1.5cm-2\fboxsep\relax][c]{\raisebox{0pt}[\ht\strutbox][\dp\strutbox]{\textcolor{black}{\textbf{0.774{\tiny $\pm$0.04}}}}}}} & \makebox[1.5cm][c]{\colorbox[rgb]{1,0.95,0.8}{\makebox[\dimexpr 1.5cm-2\fboxsep\relax][c]{\raisebox{0pt}[\ht\strutbox][\dp\strutbox]{\textcolor{black}{\textbf{0.769{\tiny $\pm$0.05}}}}}}} & \makebox[1.5cm][c]{\colorbox[rgb]{1,0.95,0.8}{\makebox[\dimexpr 1.5cm-2\fboxsep\relax][c]{\raisebox{0pt}[\ht\strutbox][\dp\strutbox]{\textcolor{black}{\textbf{0.771{\tiny $\pm$0.05}}}}}}} \\
Random & AS$\uparrow$ & \makebox[1.5cm][c]{\colorbox[gray]{0.85}{\makebox[\dimexpr 1.5cm-2\fboxsep\relax][c]{\raisebox{0pt}[\ht\strutbox][\dp\strutbox]{\textcolor{black}{\textbf{0.707{\tiny $\pm$0.06}}}}}}} & \makebox[1.5cm][c]{\colorbox[gray]{0.85}{\makebox[\dimexpr 1.5cm-2\fboxsep\relax][c]{\raisebox{0pt}[\ht\strutbox][\dp\strutbox]{\textcolor{black}{\textbf{0.661{\tiny $\pm$0.07}}}}}}} & \makebox[1.5cm][c]{\colorbox[gray]{0.85}{\makebox[\dimexpr 1.5cm-2\fboxsep\relax][c]{\raisebox{0pt}[\ht\strutbox][\dp\strutbox]{\textcolor{black}{\textbf{0.787{\tiny $\pm$0.04}}}}}}} & \makebox[1.5cm][c]{\colorbox[gray]{0.85}{\makebox[\dimexpr 1.5cm-2\fboxsep\relax][c]{\raisebox{0pt}[\ht\strutbox][\dp\strutbox]{\textcolor{black}{\textbf{0.686{\tiny $\pm$0.06}}}}}}} & \makebox[1.5cm][c]{\colorbox[gray]{0.85}{\makebox[\dimexpr 1.5cm-2\fboxsep\relax][c]{\raisebox{0pt}[\ht\strutbox][\dp\strutbox]{\textcolor{black}{\textbf{0.707{\tiny $\pm$0.06}}}}}}} & \makebox[1.5cm][c]{\colorbox[gray]{0.85}{\makebox[\dimexpr 1.5cm-2\fboxsep\relax][c]{\raisebox{0pt}[\ht\strutbox][\dp\strutbox]{\textcolor{black}{\textbf{0.706{\tiny $\pm$0.06}}}}}}} & \makebox[1.5cm][c]{0.129{\tiny $\pm$0.03}} & \makebox[1.5cm][c]{\colorbox[gray]{0.85}{\makebox[\dimexpr 1.5cm-2\fboxsep\relax][c]{\raisebox{0pt}[\ht\strutbox][\dp\strutbox]{\textcolor{black}{\textbf{0.760{\tiny $\pm$0.06}}}}}}} & \makebox[1.5cm][c]{\colorbox[gray]{0.85}{\makebox[\dimexpr 1.5cm-2\fboxsep\relax][c]{\raisebox{0pt}[\ht\strutbox][\dp\strutbox]{\textcolor{black}{\textbf{0.942{\tiny $\pm$0.02}}}}}}} & \makebox[1.5cm][c]{\colorbox[gray]{0.85}{\makebox[\dimexpr 1.5cm-2\fboxsep\relax][c]{\raisebox{0pt}[\ht\strutbox][\dp\strutbox]{\textcolor{black}{\textbf{0.773{\tiny $\pm$0.05}}}}}}} \\
 & FOS$\downarrow$ & \makebox[1.5cm][c]{\colorbox[rgb]{1,0.95,0.8}{\makebox[\dimexpr 1.5cm-2\fboxsep\relax][c]{\raisebox{0pt}[\ht\strutbox][\dp\strutbox]{\textcolor{black}{\textbf{0.316{\tiny $\pm$0.02}}}}}}} & \makebox[1.5cm][c]{\colorbox[rgb]{1,0.95,0.8}{\makebox[\dimexpr 1.5cm-2\fboxsep\relax][c]{\raisebox{0pt}[\ht\strutbox][\dp\strutbox]{\textcolor{black}{\textbf{0.309{\tiny $\pm$0.02}}}}}}} & \makebox[1.5cm][c]{\colorbox[rgb]{1,0.95,0.8}{\makebox[\dimexpr 1.5cm-2\fboxsep\relax][c]{\raisebox{0pt}[\ht\strutbox][\dp\strutbox]{\textcolor{black}{\textbf{0.313{\tiny $\pm$0.02}}}}}}} & \makebox[1.5cm][c]{\colorbox[rgb]{1,0.95,0.8}{\makebox[\dimexpr 1.5cm-2\fboxsep\relax][c]{\raisebox{0pt}[\ht\strutbox][\dp\strutbox]{\textcolor{black}{\textbf{0.313{\tiny $\pm$0.02}}}}}}} & \makebox[1.5cm][c]{\colorbox[rgb]{1,0.95,0.8}{\makebox[\dimexpr 1.5cm-2\fboxsep\relax][c]{\raisebox{0pt}[\ht\strutbox][\dp\strutbox]{\textcolor{black}{\textbf{0.335{\tiny $\pm$0.04}}}}}}} & \makebox[1.5cm][c]{\colorbox[rgb]{1,0.95,0.8}{\makebox[\dimexpr 1.5cm-2\fboxsep\relax][c]{\raisebox{0pt}[\ht\strutbox][\dp\strutbox]{\textcolor{black}{\textbf{0.311{\tiny $\pm$0.02}}}}}}} & \makebox[1.5cm][c]{\colorbox[rgb]{1,0.95,0.8}{\makebox[\dimexpr 1.5cm-2\fboxsep\relax][c]{\raisebox{0pt}[\ht\strutbox][\dp\strutbox]{\textcolor{black}{\textbf{0.336{\tiny $\pm$0.02}}}}}}} & \makebox[1.5cm][c]{\colorbox[rgb]{1,0.95,0.8}{\makebox[\dimexpr 1.5cm-2\fboxsep\relax][c]{\raisebox{0pt}[\ht\strutbox][\dp\strutbox]{\textcolor{black}{\textbf{0.309{\tiny $\pm$0.02}}}}}}} & \makebox[1.5cm][c]{\colorbox[rgb]{1,0.95,0.8}{\makebox[\dimexpr 1.5cm-2\fboxsep\relax][c]{\raisebox{0pt}[\ht\strutbox][\dp\strutbox]{\textcolor{black}{\textbf{0.322{\tiny $\pm$0.02}}}}}}} & \makebox[1.5cm][c]{\colorbox[rgb]{1,0.95,0.8}{\makebox[\dimexpr 1.5cm-2\fboxsep\relax][c]{\raisebox{0pt}[\ht\strutbox][\dp\strutbox]{\textcolor{black}{\textbf{0.335{\tiny $\pm$0.02}}}}}}} \\
\midrule
 & Acc$\uparrow$ & \makebox[1.5cm][c]{0.807{\tiny $\pm$0.03}} & \makebox[1.5cm][c]{\colorbox[gray]{0.85}{\makebox[\dimexpr 1.5cm-2\fboxsep\relax][c]{\raisebox{0pt}[\ht\strutbox][\dp\strutbox]{\textcolor{black}{\textbf{0.820{\tiny $\pm$0.02}}}}}}} & \makebox[1.5cm][c]{\bronze{0.813{\tiny $\pm$0.02}}} & \makebox[1.5cm][c]{\bronze{0.808{\tiny $\pm$0.02}}} & \makebox[1.5cm][c]{0.804{\tiny $\pm$0.03}} & \makebox[1.5cm][c]{0.807{\tiny $\pm$0.02}} & \makebox[1.5cm][c]{0.738{\tiny $\pm$0.05}} & \makebox[1.5cm][c]{0.805{\tiny $\pm$0.03}} & \makebox[1.5cm][c]{\bronze{0.808{\tiny $\pm$0.02}}} & \makebox[1.5cm][c]{0.804{\tiny $\pm$0.03}} \\
Rotate & AS$\uparrow$ & \makebox[1.5cm][c]{\bronze{0.716{\tiny $\pm$0.05}}} & \makebox[1.5cm][c]{\bronze{0.656{\tiny $\pm$0.04}}} & \makebox[1.5cm][c]{\bronze{0.826{\tiny $\pm$0.04}}} & \makebox[1.5cm][c]{\bronze{0.671{\tiny $\pm$0.04}}} & \makebox[1.5cm][c]{\bronze{0.707{\tiny $\pm$0.04}}} & \makebox[1.5cm][c]{\bronze{0.731{\tiny $\pm$0.04}}} & \makebox[1.5cm][c]{0.010{\tiny $\pm$0.01}} & \makebox[1.5cm][c]{\bronze{0.798{\tiny $\pm$0.05}}} & \makebox[1.5cm][c]{\bronze{0.962{\tiny $\pm$0.02}}} & \makebox[1.5cm][c]{\bronze{0.822{\tiny $\pm$0.04}}} \\
 & FOS$\downarrow$ & \makebox[1.5cm][c]{\bronze{0.218{\tiny $\pm$0.02}}} & \makebox[1.5cm][c]{\bronze{0.189{\tiny $\pm$0.01}}} & \makebox[1.5cm][c]{\bronze{0.196{\tiny $\pm$0.02}}} & \makebox[1.5cm][c]{\bronze{0.221{\tiny $\pm$0.02}}} & \makebox[1.5cm][c]{\bronze{0.213{\tiny $\pm$0.03}}} & \makebox[1.5cm][c]{\bronze{0.208{\tiny $\pm$0.02}}} & \makebox[1.5cm][c]{0.289{\tiny $\pm$0.02}} & \makebox[1.5cm][c]{\bronze{0.207{\tiny $\pm$0.01}}} & \makebox[1.5cm][c]{0.246{\tiny $\pm$0.02}} & \makebox[1.5cm][c]{\bronze{0.240{\tiny $\pm$0.02}}} \\
\midrule
 & Acc$\uparrow$ & \makebox[1.5cm][c]{\colorbox[rgb]{1,0.95,0.8}{\makebox[\dimexpr 1.5cm-2\fboxsep\relax][c]{\raisebox{0pt}[\ht\strutbox][\dp\strutbox]{\textcolor{black}{\textbf{0.772{\tiny $\pm$0.03}}}}}}} & \makebox[1.5cm][c]{\colorbox[rgb]{1,0.95,0.8}{\makebox[\dimexpr 1.5cm-2\fboxsep\relax][c]{\raisebox{0pt}[\ht\strutbox][\dp\strutbox]{\textcolor{black}{\textbf{0.778{\tiny $\pm$0.03}}}}}}} & \makebox[1.5cm][c]{\colorbox[rgb]{1,0.95,0.8}{\makebox[\dimexpr 1.5cm-2\fboxsep\relax][c]{\raisebox{0pt}[\ht\strutbox][\dp\strutbox]{\textcolor{black}{\textbf{0.773{\tiny $\pm$0.03}}}}}}} & \makebox[1.5cm][c]{\colorbox[rgb]{1,0.95,0.8}{\makebox[\dimexpr 1.5cm-2\fboxsep\relax][c]{\raisebox{0pt}[\ht\strutbox][\dp\strutbox]{\textcolor{black}{\textbf{0.766{\tiny $\pm$0.03}}}}}}} & \makebox[1.5cm][c]{\colorbox[rgb]{1,0.95,0.8}{\makebox[\dimexpr 1.5cm-2\fboxsep\relax][c]{\raisebox{0pt}[\ht\strutbox][\dp\strutbox]{\textcolor{black}{\textbf{0.755{\tiny $\pm$0.04}}}}}}} & \makebox[1.5cm][c]{\colorbox[rgb]{1,0.95,0.8}{\makebox[\dimexpr 1.5cm-2\fboxsep\relax][c]{\raisebox{0pt}[\ht\strutbox][\dp\strutbox]{\textcolor{black}{\textbf{0.772{\tiny $\pm$0.03}}}}}}} & \makebox[1.5cm][c]{0.700{\tiny $\pm$0.06}} & \makebox[1.5cm][c]{\colorbox[rgb]{1,0.95,0.8}{\makebox[\dimexpr 1.5cm-2\fboxsep\relax][c]{\raisebox{0pt}[\ht\strutbox][\dp\strutbox]{\textcolor{black}{\textbf{0.768{\tiny $\pm$0.03}}}}}}} & \makebox[1.5cm][c]{\colorbox[rgb]{1,0.95,0.8}{\makebox[\dimexpr 1.5cm-2\fboxsep\relax][c]{\raisebox{0pt}[\ht\strutbox][\dp\strutbox]{\textcolor{black}{\textbf{0.771{\tiny $\pm$0.03}}}}}}} & \makebox[1.5cm][c]{\colorbox[rgb]{1,0.95,0.8}{\makebox[\dimexpr 1.5cm-2\fboxsep\relax][c]{\raisebox{0pt}[\ht\strutbox][\dp\strutbox]{\textcolor{black}{\textbf{0.764{\tiny $\pm$0.03}}}}}}} \\
Average score & AS$\uparrow$ & \makebox[1.5cm][c]{\bronze{0.618{\tiny $\pm$0.04}}} & \makebox[1.5cm][c]{\bronze{0.616{\tiny $\pm$0.04}}} & \makebox[1.5cm][c]{\bronze{0.687{\tiny $\pm$0.04}}} & \makebox[1.5cm][c]{\bronze{0.600{\tiny $\pm$0.04}}} & \makebox[1.5cm][c]{\bronze{0.627{\tiny $\pm$0.04}}} & \makebox[1.5cm][c]{\bronze{0.610{\tiny $\pm$0.04}}} & \makebox[1.5cm][c]{0.069{\tiny $\pm$0.02}} & \makebox[1.5cm][c]{\bronze{0.698{\tiny $\pm$0.04}}} & \makebox[1.5cm][c]{\colorbox[rgb]{1,0.95,0.8}{\makebox[\dimexpr 1.5cm-2\fboxsep\relax][c]{\raisebox{0pt}[\ht\strutbox][\dp\strutbox]{\textcolor{black}{\textbf{0.951{\tiny $\pm$0.02}}}}}}} & \makebox[1.5cm][c]{\bronze{0.733{\tiny $\pm$0.04}}} \\
 & FOS$\downarrow$ & \makebox[1.5cm][c]{\colorbox[rgb]{1,0.95,0.8}{\makebox[\dimexpr 1.5cm-2\fboxsep\relax][c]{\raisebox{0pt}[\ht\strutbox][\dp\strutbox]{\textcolor{black}{\textbf{0.278{\tiny $\pm$0.02}}}}}}} & \makebox[1.5cm][c]{\colorbox[rgb]{1,0.95,0.8}{\makebox[\dimexpr 1.5cm-2\fboxsep\relax][c]{\raisebox{0pt}[\ht\strutbox][\dp\strutbox]{\textcolor{black}{\textbf{0.264{\tiny $\pm$0.01}}}}}}} & \makebox[1.5cm][c]{\colorbox[rgb]{1,0.95,0.8}{\makebox[\dimexpr 1.5cm-2\fboxsep\relax][c]{\raisebox{0pt}[\ht\strutbox][\dp\strutbox]{\textcolor{black}{\textbf{0.270{\tiny $\pm$0.02}}}}}}} & \makebox[1.5cm][c]{\colorbox[rgb]{1,0.95,0.8}{\makebox[\dimexpr 1.5cm-2\fboxsep\relax][c]{\raisebox{0pt}[\ht\strutbox][\dp\strutbox]{\textcolor{black}{\textbf{0.278{\tiny $\pm$0.02}}}}}}} & \makebox[1.5cm][c]{\colorbox[rgb]{1,0.95,0.8}{\makebox[\dimexpr 1.5cm-2\fboxsep\relax][c]{\raisebox{0pt}[\ht\strutbox][\dp\strutbox]{\textcolor{black}{\textbf{0.282{\tiny $\pm$0.02}}}}}}} & \makebox[1.5cm][c]{\colorbox[rgb]{1,0.95,0.8}{\makebox[\dimexpr 1.5cm-2\fboxsep\relax][c]{\raisebox{0pt}[\ht\strutbox][\dp\strutbox]{\textcolor{black}{\textbf{0.271{\tiny $\pm$0.02}}}}}}} & \makebox[1.5cm][c]{0.325{\tiny $\pm$0.01}} & \makebox[1.5cm][c]{\colorbox[rgb]{1,0.95,0.8}{\makebox[\dimexpr 1.5cm-2\fboxsep\relax][c]{\raisebox{0pt}[\ht\strutbox][\dp\strutbox]{\textcolor{black}{\textbf{0.272{\tiny $\pm$0.01}}}}}}} & \makebox[1.5cm][c]{\bronze{0.299{\tiny $\pm$0.02}}} & \makebox[1.5cm][c]{\bronze{0.296{\tiny $\pm$0.02}}} \\
\bottomrule
\end{tabular}
\end{adjustbox}
\end{table}

%% file: tables/singlecell.tex
\begin{table}[!t]
\caption{\textbf{Biological conservation and batch correction on the lung integration benchmark.} Results are mean $\pm$ average within-pair standard deviation across seeds. Standard deviations are computed separately within each selected folder and batch pair, then averaged across pairs; higher is better for every metric. Methods are ordered identically in both sections by average Bio/Batch rank, with alphabetical tie-breaking. Gold, silver, and bronze mark the top three distinct scores per metric; ties share a highlight.}
\label{tab:singlecell_full}
\centering
\small
\providecommand{\bronze}[1]{\textcolor[rgb]{0.70,0.43,0.20}{\textbf{#1}}}
\setlength{\tabcolsep}{2pt}
\setlength{\fboxsep}{1pt}
\renewcommand{\arraystretch}{1.0}
\begin{tabular}{lccccc@{\hspace{7pt}}c}
\toprule
\multicolumn{7}{c}{\textbf{Biological conservation}} \\
\cmidrule(lr){1-7}
Method & \makebox[1.5cm][c]{Isolated $\uparrow$} & \makebox[1.5cm][c]{NMI $\uparrow$} & \makebox[1.5cm][c]{ARI $\uparrow$} & \makebox[1.5cm][c]{ASW label $\uparrow$} & \makebox[1.5cm][c]{cLISI $\uparrow$} & \makebox[1.5cm][c]{Bio score $\uparrow$} \\
\midrule
\textbf{FoSTA} & \makebox[1.5cm][c]{0.490{\tiny $\pm$0.02}} & \makebox[1.5cm][c]{\colorbox[gray]{0.85}{\makebox[\dimexpr 1.5cm-2\fboxsep\relax][c]{\raisebox{0pt}[\ht\strutbox][\dp\strutbox]{\textcolor{black}{\textbf{0.628{\tiny $\pm$0.01}}}}}}} & \makebox[1.5cm][c]{\colorbox[rgb]{1,0.95,0.8}{\makebox[\dimexpr 1.5cm-2\fboxsep\relax][c]{\raisebox{0pt}[\ht\strutbox][\dp\strutbox]{\textcolor{black}{\textbf{0.525{\tiny $\pm$0.03}}}}}}} & \makebox[1.5cm][c]{0.452{\tiny $\pm$0.01}} & \makebox[1.5cm][c]{0.989{\tiny $\pm$0.00}} & \makebox[1.5cm][c]{\colorbox[gray]{0.85}{\makebox[\dimexpr 1.5cm-2\fboxsep\relax][c]{\raisebox{0pt}[\ht\strutbox][\dp\strutbox]{\textcolor{black}{\textbf{0.617{\tiny $\pm$0.01}}}}}}} \\
scANVI & \makebox[1.5cm][c]{\colorbox[rgb]{1,0.95,0.8}{\makebox[\dimexpr 1.5cm-2\fboxsep\relax][c]{\raisebox{0pt}[\ht\strutbox][\dp\strutbox]{\textcolor{black}{\textbf{0.560{\tiny $\pm$0.02}}}}}}} & \makebox[1.5cm][c]{\colorbox[rgb]{1,0.95,0.8}{\makebox[\dimexpr 1.5cm-2\fboxsep\relax][c]{\raisebox{0pt}[\ht\strutbox][\dp\strutbox]{\textcolor{black}{\textbf{0.735{\tiny $\pm$0.01}}}}}}} & \makebox[1.5cm][c]{\colorbox[gray]{0.85}{\makebox[\dimexpr 1.5cm-2\fboxsep\relax][c]{\raisebox{0pt}[\ht\strutbox][\dp\strutbox]{\textcolor{black}{\textbf{0.498{\tiny $\pm$0.04}}}}}}} & \makebox[1.5cm][c]{\colorbox[rgb]{1,0.95,0.8}{\makebox[\dimexpr 1.5cm-2\fboxsep\relax][c]{\raisebox{0pt}[\ht\strutbox][\dp\strutbox]{\textcolor{black}{\textbf{0.495{\tiny $\pm$0.01}}}}}}} & \makebox[1.5cm][c]{\colorbox[rgb]{1,0.95,0.8}{\makebox[\dimexpr 1.5cm-2\fboxsep\relax][c]{\raisebox{0pt}[\ht\strutbox][\dp\strutbox]{\textcolor{black}{\textbf{1.000{\tiny $\pm$0.00}}}}}}} & \makebox[1.5cm][c]{\colorbox[rgb]{1,0.95,0.8}{\makebox[\dimexpr 1.5cm-2\fboxsep\relax][c]{\raisebox{0pt}[\ht\strutbox][\dp\strutbox]{\textcolor{black}{\textbf{0.658{\tiny $\pm$0.01}}}}}}} \\
scVI & \makebox[1.5cm][c]{\colorbox[gray]{0.85}{\makebox[\dimexpr 1.5cm-2\fboxsep\relax][c]{\raisebox{0pt}[\ht\strutbox][\dp\strutbox]{\textcolor{black}{\textbf{0.535{\tiny $\pm$0.01}}}}}}} & \makebox[1.5cm][c]{\bronze{0.627{\tiny $\pm$0.01}}} & \makebox[1.5cm][c]{\bronze{0.387{\tiny $\pm$0.03}}} & \makebox[1.5cm][c]{0.433{\tiny $\pm$0.01}} & \makebox[1.5cm][c]{\colorbox[gray]{0.85}{\makebox[\dimexpr 1.5cm-2\fboxsep\relax][c]{\raisebox{0pt}[\ht\strutbox][\dp\strutbox]{\textcolor{black}{\textbf{0.994{\tiny $\pm$0.00}}}}}}} & \makebox[1.5cm][c]{\bronze{0.595{\tiny $\pm$0.01}}} \\
MALI & \makebox[1.5cm][c]{0.495{\tiny $\pm$0.02}} & \makebox[1.5cm][c]{0.513{\tiny $\pm$0.02}} & \makebox[1.5cm][c]{0.312{\tiny $\pm$0.03}} & \makebox[1.5cm][c]{0.377{\tiny $\pm$0.01}} & \makebox[1.5cm][c]{0.961{\tiny $\pm$0.00}} & \makebox[1.5cm][c]{0.532{\tiny $\pm$0.01}} \\
KEMArbf & \makebox[1.5cm][c]{0.464{\tiny $\pm$0.02}} & \makebox[1.5cm][c]{0.297{\tiny $\pm$0.06}} & \makebox[1.5cm][c]{0.170{\tiny $\pm$0.07}} & \makebox[1.5cm][c]{0.430{\tiny $\pm$0.01}} & \makebox[1.5cm][c]{0.912{\tiny $\pm$0.02}} & \makebox[1.5cm][c]{0.455{\tiny $\pm$0.03}} \\
Scanorama & \makebox[1.5cm][c]{0.473{\tiny $\pm$0.00}} & \makebox[1.5cm][c]{0.595{\tiny $\pm$0.00}} & \makebox[1.5cm][c]{0.382{\tiny $\pm$0.00}} & \makebox[1.5cm][c]{\colorbox[gray]{0.85}{\makebox[\dimexpr 1.5cm-2\fboxsep\relax][c]{\raisebox{0pt}[\ht\strutbox][\dp\strutbox]{\textcolor{black}{\textbf{0.460{\tiny $\pm$0.00}}}}}}} & \makebox[1.5cm][c]{0.983{\tiny $\pm$0.00}} & \makebox[1.5cm][c]{0.579{\tiny $\pm$0.00}} \\
LIGER & \makebox[1.5cm][c]{0.467{\tiny $\pm$0.00}} & \makebox[1.5cm][c]{0.470{\tiny $\pm$0.01}} & \makebox[1.5cm][c]{0.291{\tiny $\pm$0.01}} & \makebox[1.5cm][c]{0.386{\tiny $\pm$0.00}} & \makebox[1.5cm][c]{0.949{\tiny $\pm$0.00}} & \makebox[1.5cm][c]{0.513{\tiny $\pm$0.00}} \\
Pamona & \makebox[1.5cm][c]{\bronze{0.523{\tiny $\pm$0.02}}} & \makebox[1.5cm][c]{0.548{\tiny $\pm$0.02}} & \makebox[1.5cm][c]{0.286{\tiny $\pm$0.02}} & \makebox[1.5cm][c]{0.393{\tiny $\pm$0.01}} & \makebox[1.5cm][c]{\bronze{0.990{\tiny $\pm$0.00}}} & \makebox[1.5cm][c]{0.548{\tiny $\pm$0.01}} \\
Unintegrated & \makebox[1.5cm][c]{0.492{\tiny $\pm$0.00}} & \makebox[1.5cm][c]{0.579{\tiny $\pm$0.00}} & \makebox[1.5cm][c]{0.365{\tiny $\pm$0.00}} & \makebox[1.5cm][c]{\bronze{0.454{\tiny $\pm$0.00}}} & \makebox[1.5cm][c]{0.984{\tiny $\pm$0.00}} & \makebox[1.5cm][c]{0.575{\tiny $\pm$0.00}} \\
KEMAlin & \makebox[1.5cm][c]{0.482{\tiny $\pm$0.02}} & \makebox[1.5cm][c]{0.311{\tiny $\pm$0.04}} & \makebox[1.5cm][c]{0.180{\tiny $\pm$0.05}} & \makebox[1.5cm][c]{0.421{\tiny $\pm$0.01}} & \makebox[1.5cm][c]{0.884{\tiny $\pm$0.01}} & \makebox[1.5cm][c]{0.456{\tiny $\pm$0.02}} \\
\addlinespace[0.8ex]
\midrule
\multicolumn{7}{c}{\textbf{Batch correction}} \\
\cmidrule(lr){1-7}
Method & \makebox[1.5cm][c]{BRAS $\uparrow$} & \makebox[1.5cm][c]{iLISI $\uparrow$} & \makebox[1.5cm][c]{kBET $\uparrow$} & \makebox[1.5cm][c]{Graph $\uparrow$} & \makebox[1.5cm][c]{PCR $\uparrow$} & \makebox[1.5cm][c]{Batch score $\uparrow$} \\
\midrule
\textbf{FoSTA} & \makebox[1.5cm][c]{0.292{\tiny $\pm$0.02}} & \makebox[1.5cm][c]{0.484{\tiny $\pm$0.03}} & \makebox[1.5cm][c]{0.465{\tiny $\pm$0.05}} & \makebox[1.5cm][c]{0.716{\tiny $\pm$0.03}} & \makebox[1.5cm][c]{\colorbox[rgb]{1,0.95,0.8}{\makebox[\dimexpr 1.5cm-2\fboxsep\relax][c]{\raisebox{0pt}[\ht\strutbox][\dp\strutbox]{\textcolor{black}{\textbf{0.966{\tiny $\pm$0.00}}}}}}} & \makebox[1.5cm][c]{\bronze{0.585{\tiny $\pm$0.02}}} \\
scANVI & \makebox[1.5cm][c]{\colorbox[rgb]{1,0.95,0.8}{\makebox[\dimexpr 1.5cm-2\fboxsep\relax][c]{\raisebox{0pt}[\ht\strutbox][\dp\strutbox]{\textcolor{black}{\textbf{0.349{\tiny $\pm$0.02}}}}}}} & \makebox[1.5cm][c]{0.390{\tiny $\pm$0.03}} & \makebox[1.5cm][c]{\colorbox[rgb]{1,0.95,0.8}{\makebox[\dimexpr 1.5cm-2\fboxsep\relax][c]{\raisebox{0pt}[\ht\strutbox][\dp\strutbox]{\textcolor{black}{\textbf{0.699{\tiny $\pm$0.06}}}}}}} & \makebox[1.5cm][c]{\colorbox[rgb]{1,0.95,0.8}{\makebox[\dimexpr 1.5cm-2\fboxsep\relax][c]{\raisebox{0pt}[\ht\strutbox][\dp\strutbox]{\textcolor{black}{\textbf{0.953{\tiny $\pm$0.02}}}}}}} & \makebox[1.5cm][c]{0.515{\tiny $\pm$0.17}} & \makebox[1.5cm][c]{0.581{\tiny $\pm$0.04}} \\
scVI & \makebox[1.5cm][c]{0.316{\tiny $\pm$0.02}} & \makebox[1.5cm][c]{0.444{\tiny $\pm$0.03}} & \makebox[1.5cm][c]{0.528{\tiny $\pm$0.05}} & \makebox[1.5cm][c]{\colorbox[gray]{0.85}{\makebox[\dimexpr 1.5cm-2\fboxsep\relax][c]{\raisebox{0pt}[\ht\strutbox][\dp\strutbox]{\textcolor{black}{\textbf{0.843{\tiny $\pm$0.02}}}}}}} & \makebox[1.5cm][c]{0.750{\tiny $\pm$0.09}} & \makebox[1.5cm][c]{0.576{\tiny $\pm$0.03}} \\
MALI & \makebox[1.5cm][c]{0.269{\tiny $\pm$0.02}} & \makebox[1.5cm][c]{\colorbox[rgb]{1,0.95,0.8}{\makebox[\dimexpr 1.5cm-2\fboxsep\relax][c]{\raisebox{0pt}[\ht\strutbox][\dp\strutbox]{\textcolor{black}{\textbf{0.695{\tiny $\pm$0.02}}}}}}} & \makebox[1.5cm][c]{0.446{\tiny $\pm$0.05}} & \makebox[1.5cm][c]{0.710{\tiny $\pm$0.03}} & \makebox[1.5cm][c]{\colorbox[gray]{0.85}{\makebox[\dimexpr 1.5cm-2\fboxsep\relax][c]{\raisebox{0pt}[\ht\strutbox][\dp\strutbox]{\textcolor{black}{\textbf{0.930{\tiny $\pm$0.03}}}}}}} & \makebox[1.5cm][c]{\colorbox[gray]{0.85}{\makebox[\dimexpr 1.5cm-2\fboxsep\relax][c]{\raisebox{0pt}[\ht\strutbox][\dp\strutbox]{\textcolor{black}{\textbf{0.610{\tiny $\pm$0.02}}}}}}} \\
KEMArbf & \makebox[1.5cm][c]{\colorbox[gray]{0.85}{\makebox[\dimexpr 1.5cm-2\fboxsep\relax][c]{\raisebox{0pt}[\ht\strutbox][\dp\strutbox]{\textcolor{black}{\textbf{0.345{\tiny $\pm$0.03}}}}}}} & \makebox[1.5cm][c]{\bronze{0.569{\tiny $\pm$0.05}}} & \makebox[1.5cm][c]{\colorbox[gray]{0.85}{\makebox[\dimexpr 1.5cm-2\fboxsep\relax][c]{\raisebox{0pt}[\ht\strutbox][\dp\strutbox]{\textcolor{black}{\textbf{0.628{\tiny $\pm$0.07}}}}}}} & \makebox[1.5cm][c]{0.656{\tiny $\pm$0.04}} & \makebox[1.5cm][c]{\bronze{0.881{\tiny $\pm$0.04}}} & \makebox[1.5cm][c]{\colorbox[rgb]{1,0.95,0.8}{\makebox[\dimexpr 1.5cm-2\fboxsep\relax][c]{\raisebox{0pt}[\ht\strutbox][\dp\strutbox]{\textcolor{black}{\textbf{0.616{\tiny $\pm$0.02}}}}}}} \\
Scanorama & \makebox[1.5cm][c]{\bronze{0.345{\tiny $\pm$0.00}}} & \makebox[1.5cm][c]{0.391{\tiny $\pm$0.00}} & \makebox[1.5cm][c]{\bronze{0.585{\tiny $\pm$0.00}}} & \makebox[1.5cm][c]{0.748{\tiny $\pm$0.00}} & \makebox[1.5cm][c]{0.283{\tiny $\pm$0.00}} & \makebox[1.5cm][c]{0.470{\tiny $\pm$0.00}} \\
LIGER & \makebox[1.5cm][c]{0.301{\tiny $\pm$0.00}} & \makebox[1.5cm][c]{\colorbox[gray]{0.85}{\makebox[\dimexpr 1.5cm-2\fboxsep\relax][c]{\raisebox{0pt}[\ht\strutbox][\dp\strutbox]{\textcolor{black}{\textbf{0.656{\tiny $\pm$0.02}}}}}}} & \makebox[1.5cm][c]{0.482{\tiny $\pm$0.03}} & \makebox[1.5cm][c]{0.570{\tiny $\pm$0.01}} & \makebox[1.5cm][c]{0.721{\tiny $\pm$0.01}} & \makebox[1.5cm][c]{0.546{\tiny $\pm$0.01}} \\
Pamona & \makebox[1.5cm][c]{0.307{\tiny $\pm$0.02}} & \makebox[1.5cm][c]{0.283{\tiny $\pm$0.02}} & \makebox[1.5cm][c]{0.404{\tiny $\pm$0.06}} & \makebox[1.5cm][c]{\bronze{0.767{\tiny $\pm$0.02}}} & \makebox[1.5cm][c]{0.785{\tiny $\pm$0.10}} & \makebox[1.5cm][c]{0.509{\tiny $\pm$0.03}} \\
Unintegrated & \makebox[1.5cm][c]{0.334{\tiny $\pm$0.00}} & \makebox[1.5cm][c]{0.335{\tiny $\pm$0.00}} & \makebox[1.5cm][c]{0.516{\tiny $\pm$0.00}} & \makebox[1.5cm][c]{0.729{\tiny $\pm$0.00}} & \makebox[1.5cm][c]{0.304{\tiny $\pm$0.00}} & \makebox[1.5cm][c]{0.444{\tiny $\pm$0.00}} \\
KEMAlin & \makebox[1.5cm][c]{0.322{\tiny $\pm$0.02}} & \makebox[1.5cm][c]{0.468{\tiny $\pm$0.05}} & \makebox[1.5cm][c]{0.437{\tiny $\pm$0.07}} & \makebox[1.5cm][c]{0.603{\tiny $\pm$0.04}} & \makebox[1.5cm][c]{0.719{\tiny $\pm$0.07}} & \makebox[1.5cm][c]{0.510{\tiny $\pm$0.03}} \\
\bottomrule
\end{tabular}
\end{table}